\documentclass[]{worldbench}
\definecolor{w_blue}{RGB}{52,204,204}
\definecolor{w_yellow}{RGB}{255,192,0}

\usepackage{soul}
\newcommand{\hlred}[1]{{\sethlcolor{pi3det_red!12}\hl{#1}}}
\newcommand{\hlblue}[1]{{\sethlcolor{pi3det_blue!12}\hl{#1}}}

\usepackage{pifont}
\newcommand{\cmark}{\ding{51}}
\newcommand{\xmark}{\ding{55}}

\definecolor{pi3det_red}{RGB}{192,0,0}
\definecolor{pi3det_blue}{RGB}{0,112,192}
\definecolor{pi3det_green}{RGB}{0,176,80}
\definecolor{cell}{RGB}{230,248,243}

\usepackage{amsfonts}
\usepackage{amsmath}
\usepackage{amssymb}

\usepackage{colortbl}

\usepackage{makecell}
\usepackage{multicol}
\usepackage{multirow}
\usepackage{pifont}

\usepackage{wrapfig}

\usepackage{titletoc}

\title{Perspective-Invariant 3D Object Detection}

\author[]{Ao~Liang~\raisebox{0.2em}{\includegraphics[width=0.019\linewidth]{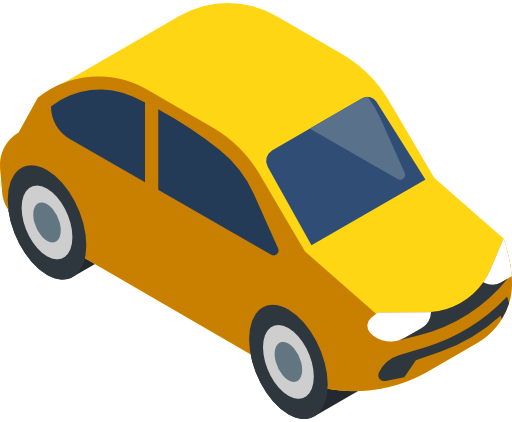}}}
\author[]{Lingdong~Kong~\raisebox{0.2em}{\includegraphics[width=0.019\linewidth]{figures/icons/car1.png}}~\raisebox{0.2em}{\includegraphics[width=0.019\linewidth]{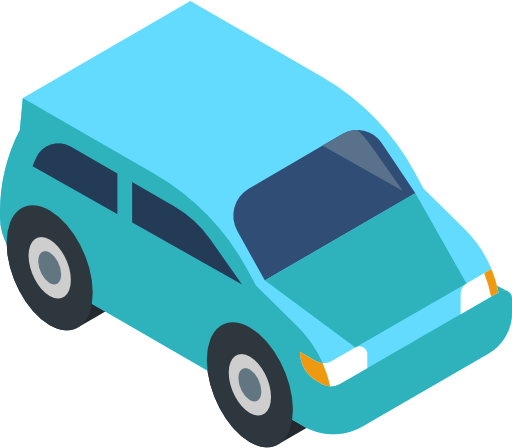}}}
\author[]{Dongyue~Lu~\raisebox{0.2em}{\includegraphics[width=0.019\linewidth]{figures/icons/car1.png}}}
\author[]{Youquan~Liu}
\author[]{Jian~Fang}
\author[]{Huaici~Zhao~\raisebox{0.15em}{\includegraphics[width=0.017\linewidth]{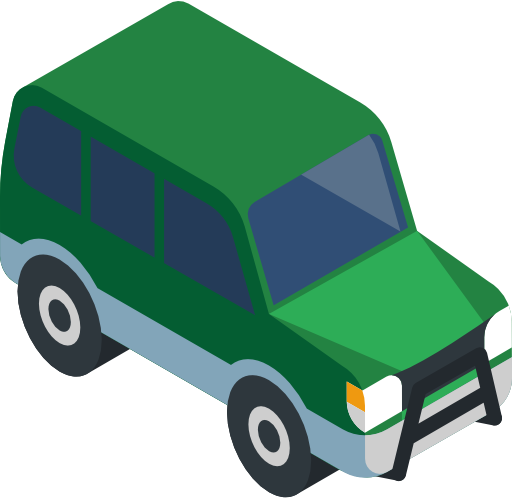}}}
\author[]{Wei~Tsang~Ooi~\raisebox{0.15em}{\includegraphics[width=0.017\linewidth]{figures/icons/car4.png}}}

\affiliation[]{
\raisebox{-0.1em}{\includegraphics[width=0.029\linewidth]{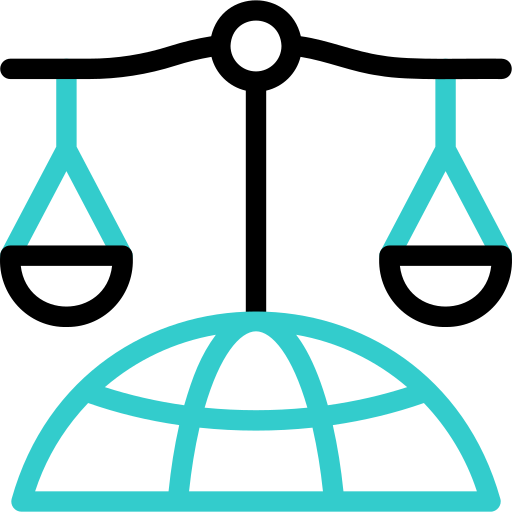}}~WorldBench Team
\\[1.2ex]
~\raisebox{-0.2em}{\includegraphics[width=0.032\linewidth]{figures/icons/car1.png}}~{\small \textbf{Equal Contributions}}
\quad
\raisebox{-0.2em}{\includegraphics[width=0.031\linewidth]{figures/icons/car2.png}}~{\small \textbf{Project Lead}}
\quad
\raisebox{-0.2em}{\includegraphics[width=0.028\linewidth]{figures/icons/car4.png}}~{\small \textbf{Corresponding Author}}
}

\abstract{
With the rise of robotics, LiDAR-based 3D object detection has garnered significant attention in both academia and industry. However, existing datasets and methods predominantly focus on vehicle-mounted platforms, leaving other autonomous platforms underexplored. To bridge this gap, we introduce \textbf{Pi3DET}, the first benchmark featuring LiDAR data and 3D bounding box annotations collected from multiple platforms: vehicle, quadruped, and drone, thereby facilitating research in 3D object detection for non-vehicle platforms as well as cross-platform 3D detection. Based on Pi3DET, we propose a novel cross-platform adaptation framework that transfers knowledge from the well-studied vehicle platform to other platforms. This framework achieves perspective-invariant 3D detection through robust alignment at both geometric and feature levels. Additionally, we establish a benchmark to evaluate the resilience and robustness of current 3D detectors in cross-platform scenarios, providing valuable insights for developing adaptive 3D perception systems. Extensive experiments validate the effectiveness of our approach on challenging cross-platform tasks, demonstrating substantial gains over existing adaptation methods. We hope this work paves the way for generalizable and unified 3D perception systems across diverse and complex environments. Our Pi3DET dataset, cross-platform benchmark suite, and annotation toolkit have been made publicly available.
}

\metadata[
\raisebox{-0.2em}{\includegraphics[width=0.025\linewidth]{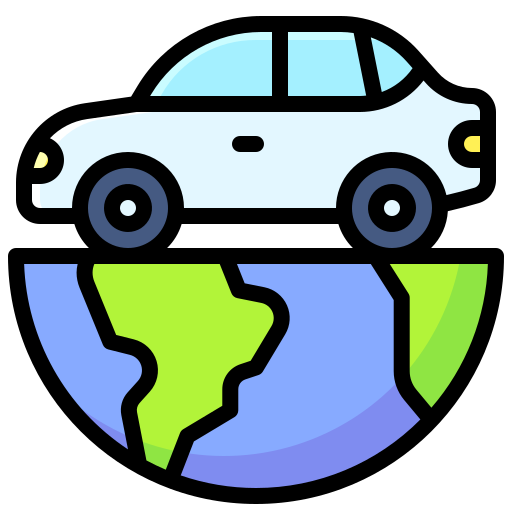}}~~Project Page]{\href{https://pi3det.github.io/}{\texttt{https://pi3det.github.io}}
\\[-1.5ex]}

\metadata[
\raisebox{-0.2em}{\includegraphics[width=0.025\linewidth]{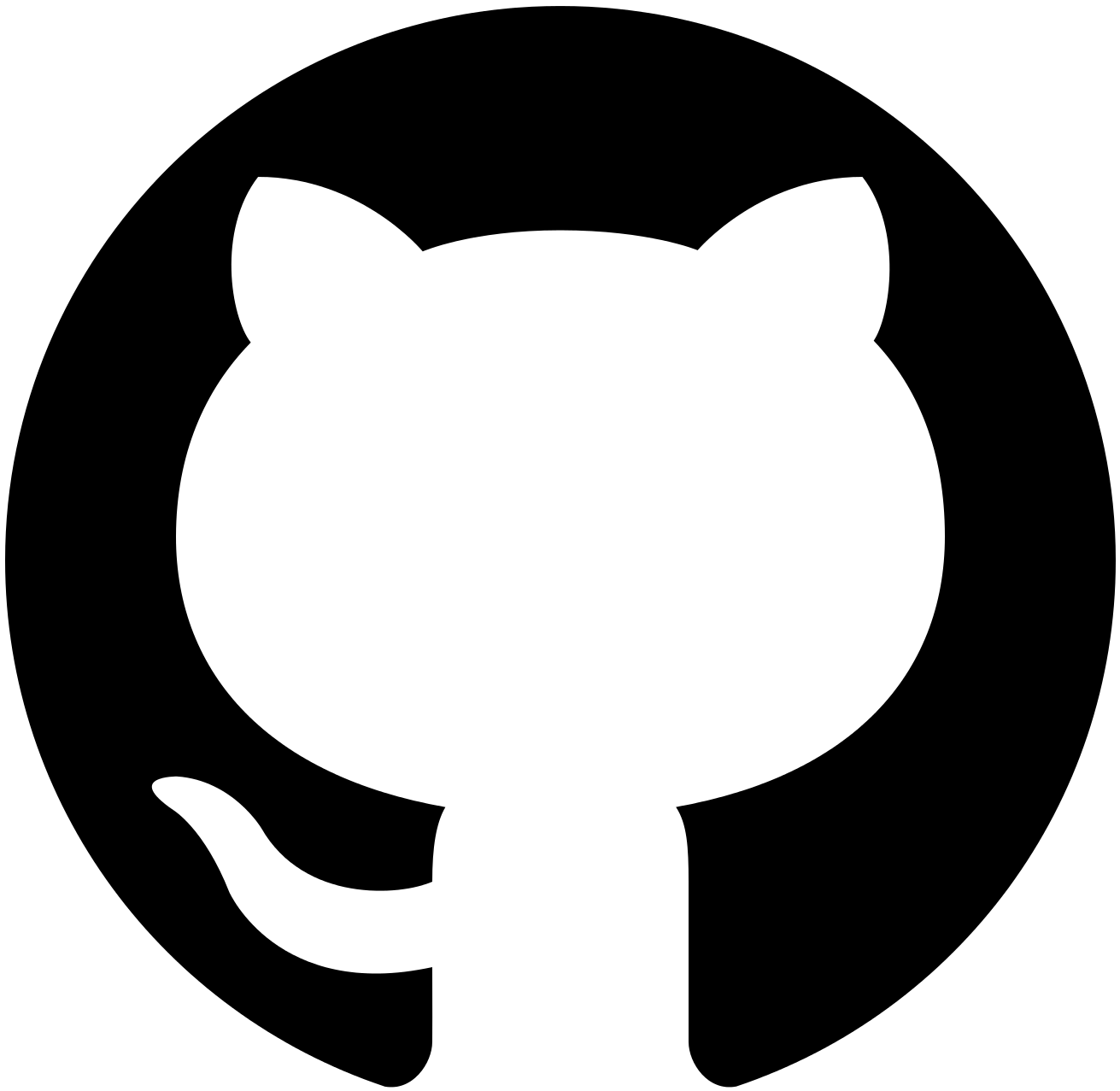}}~~GitHub Repo]{\href{https://github.com/worldbench/pi3det}{\texttt{https://github.com/worldbench/pi3det}}
\\[-1.7ex]}

\metadata[
\raisebox{-0.3em}{\includegraphics[width=0.026\linewidth]{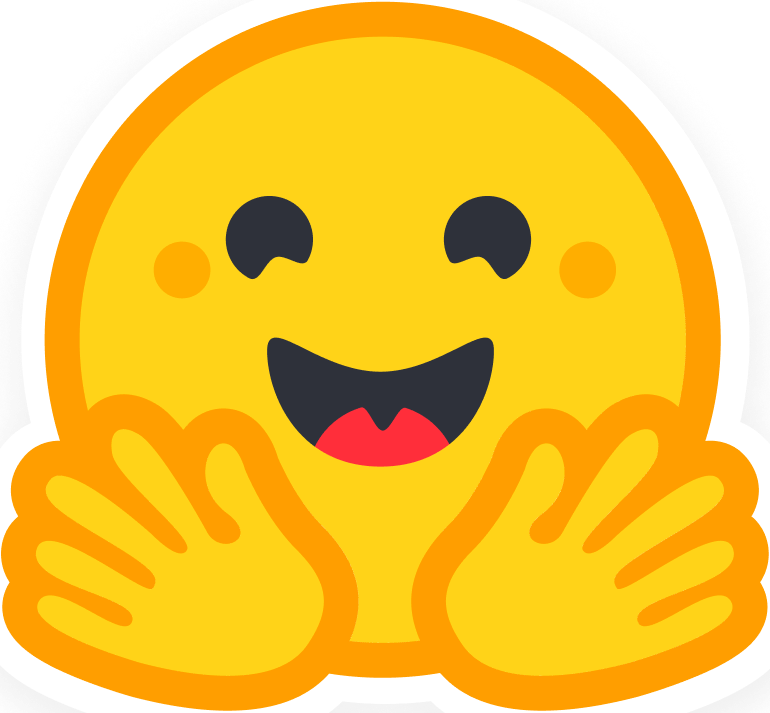}}~~HuggingFace Dataset]{\href{https://huggingface.co/datasets/Pi3DET/data}{\texttt{https://huggingface.co/datasets/Pi3DET/data}}}

\begin{document}

\maketitle

\begin{figure}[h]
    \centering
    \vspace{0.1cm}
    \includegraphics[width=\linewidth]{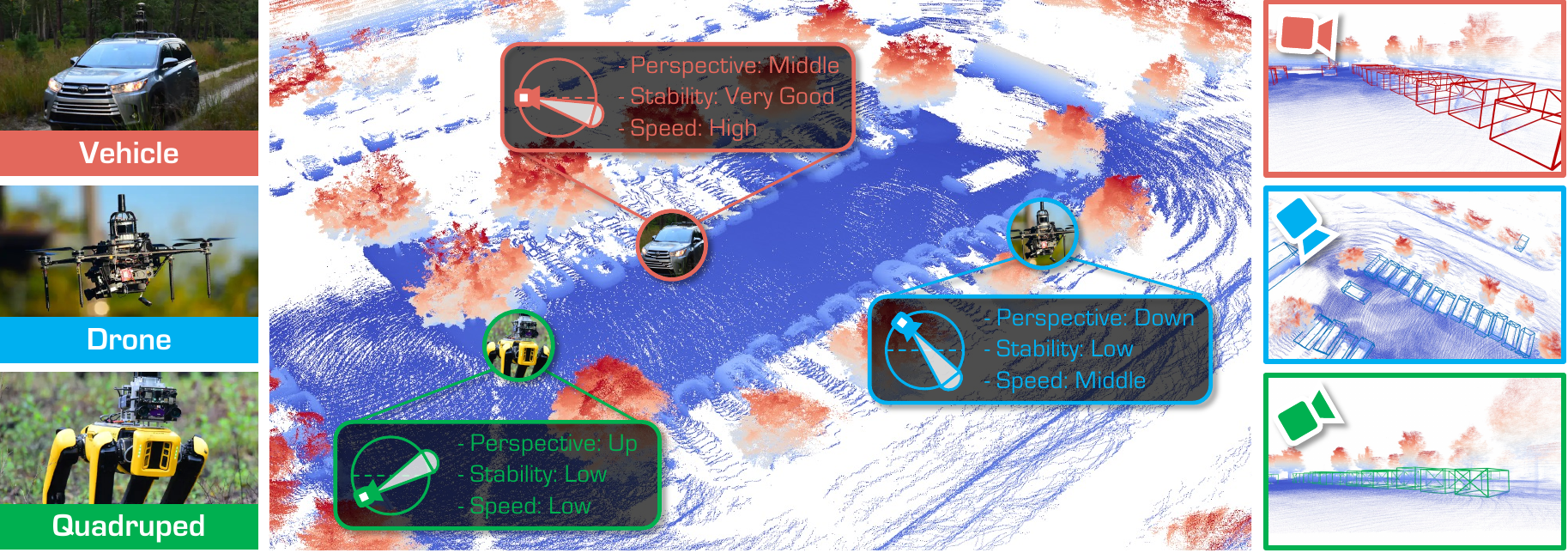}
    \vspace{-0.6cm}
    \caption{Motivation of \underline{\textbf{P}}erspective \underline{\textbf{i}}nvariant \underline{\textbf{3}}D object \underline{\textbf{DET}}ection (\textbf{Pi3DET}). We focus the practical yet challenging task of 3D object detection from heterogeneous robot platforms: \includegraphics[width=0.022\linewidth]{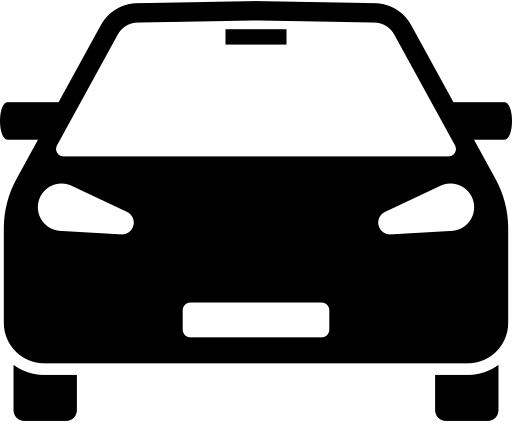} \textbf{Vehicle}, \includegraphics[width=0.023\linewidth]{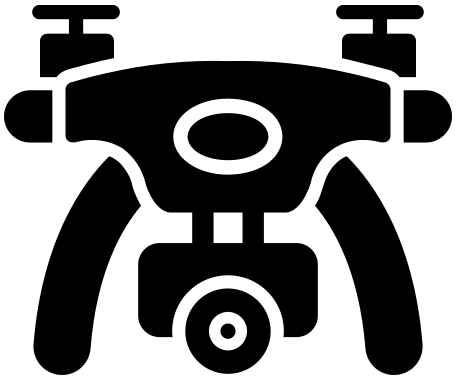} \textbf{Drone}, and \includegraphics[width=0.023\linewidth]{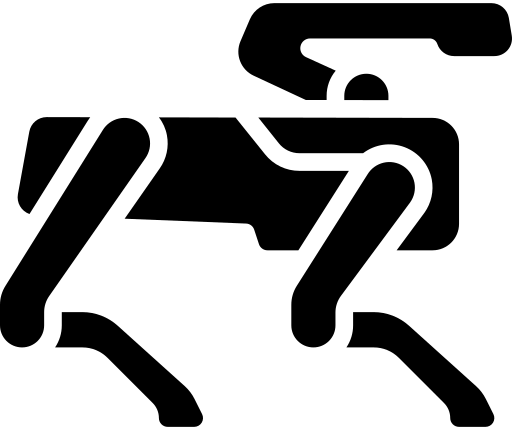} \textbf{Quadruped}. To achieve strong generalization, we contribute: \textbf{1)} The first \textbf{dataset} for multi-platform 3D detection, comprising more than $\mathbf{51}$K LiDAR frames with over $\mathbf{250k}$ meticulously annotated 3D bounding boxes; \textbf{2)} An adaptation \textbf{framework}, effectively transfers capabilities from vehicles to other platforms by integrating geometric and feature-level representations; \textbf{3)} A comprehensive \textbf{benchmark} study of state-of-the-art 3D detectors on cross-platform scenarios.}
    \label{fig:taxonomy}
    \vspace{-0.3cm}
\end{figure}

\section{Introduction}
\label{sec:intro}

LiDAR-based 3D object detection provides detailed spatial and geometric information about objects of interest, attracting significant research attention \cite{abbasi2022lidar, kong2025multi, zheng2021efficient, li2025your}. Despite this trend, existing datasets \cite{geiger2012we, caesar2020nuscenes, sun2020scalability, mao2021one} and methods \cite{shi2020pv, shi2023pv, yin2021center, zhang2022not, li2023pillarnext, kong2023lasermix, kong2023rangeformer} predominantly target autonomous vehicles, leaving other platforms underexplored.

With rapid advancements in robotics, autonomous systems such as quadrupeds and drones are becoming increasingly vital for diverse real-world applications \cite{chaney2023m3ed, kong2025eventfly, tian2024ucdnet, hong2024dsnet, liu2023uniseg, bian2025dynamiccity, puy2023waffle, xu2025frnet, ando2023rangevit}. Equipping these emerging platforms with accurate 3D perception capabilities comparable to those of autonomous vehicles is therefore highly significant \cite{li2025your,kong2023robodepth,xie2025robobev,kong2025calib3d,bijelic2020seeing,geyer2020a2d2}. Currently, research into non-vehicle platforms remains sparse \cite{liu2021efficient,qin2020lins,tian2024ucdnet,mmdet3d,lazarow2025cubify}, revealing a critical gap in cross-platform 3D object detection studies.

A major barrier impeding progress in multi-platform detection is the lack of annotated multi-platform LiDAR datasets. Current benchmarks almost exclusively focus on vehicles \cite{caesar2020nuscenes, geiger2012we, sun2020scalability, zhang2024sparselif, song2024robustness}. Although some drone datasets exist \cite{tian2024ucdnet, chaney2023m3ed}, they often lack comprehensive 3D annotations and sufficient platform diversity. Chaney \emph{et al.} introduce M3ED \cite{chaney2023m3ed}, a dataset compiled from multiple platforms. However, the lack of annotated 3D bounding boxes currently limits its direct applicability for 3D detection tasks. 

Training platform-specific models independently is both resource-intensive and impractical for real-world deployment, especially in resource-constrained scenarios. Cross-platform adaptation, transferring knowledge from well-studied vehicle datasets to other platforms like drones and quadrupeds, emerges as a promising alternative. Existing domain adaptation techniques \cite{zhou2023survey}, however, primarily tackle cross-dataset shifts and neglect intrinsic geometric discrepancies caused by differences in platform dynamics and sensor viewpoints.

To address these limitations, we introduce \textbf{Pi3DET}, the \textbf{first} publicly available multi-platform 3D detection dataset. Our dataset consists of $\mathbf{51{,}545}$ LiDAR frames with over $\mathbf{250{,}000}$ meticulously annotated 3D bounding boxes spanning \includegraphics[width=0.022\linewidth]{figures/icons/vehicle.png} \textbf{Vehicle}, \includegraphics[width=0.023\linewidth]{figures/icons/drone.png} \textbf{Drone}, and \includegraphics[width=0.023\linewidth]{figures/icons/quadruped.png} \textbf{Quadruped}. Our dataset is constructed using an automated labeling pipeline, supplemented by extensive manual refinement totaling approximately $\mathbf{500}$ hours. As detailed in Table~\ref{tab:dataset-compare}, Pi3DET contains $\mathbf{25}$ sequences covering diverse environments under varying day and night conditions (examples in Appendix~\ref{sec_supp:dataset_examples}). Analyses of Pi3DET highlight \textbf{three crucial discrepancies across platforms}: differences in ego-motion characteristics, variations in point-cloud distributions, and distinct bounding box properties, underscoring the necessity for specialized adaptation methods and techniques.

Motivated by these insights, we propose \textbf{Pi3DET-Net}, a novel cross-platform adaptation framework. Our approach consists of two stages. In the \textit{Pre-Adaptation (PA)} stage, we learn global transformations and extract geometric cues from the source platform. In the \textit{Knowledge Adaptation (KA)} stage, we propagate the acquired knowledge and align features between the source and target platforms to improve cross-platform generalization. In particular, our method effectively bridges the platform gap among heterogeneous robotic systems at both the \textbf{geometric} and \textbf{feature} levels:

\noindent\textbf{\textcolor{pi3det_red}{{\footnotesize$\blacksquare$}} Geometry-Level.} We develop \textit{Random Platform Jitter (RPJ)} to augment source data with simulated ego-motion disturbances, enhancing robustness to platform-specific motion variations. Moreover, \textit{Virtual Platform Pose (VPP)} projects target platform point clouds into a source-like coordinate frame, mitigating viewpoint discrepancies.

\noindent\textbf{\textcolor{pi3det_blue}{{\footnotesize$\blacksquare$}} Feature-Level.} Our \textit{Geometry-Aware Transformation Descriptor (GTD)} encodes platform-specific geometric properties (\emph{e.g.}, sensor elevation distributions), guiding effective feature alignment. The proposed \textit{KL Probabilistic Feature Alignment (PFA)} leverages variational inference to minimize domain-specific distribution gaps, thereby facilitating accurate platform-specific pose adaptation.

\begin{table*}
    \centering
    \caption{\textbf{Summary of LiDAR-based 3D object detection datasets}. We compare \textbf{key aspects} from $^1$robot platforms, $^2$scale, $^3$sensor setups, $^4$temporal (Temp.), $^5$multi-conditions, \emph{etc}. To our knowledge, \textbf{Pi3DET} stands out as the first work to feature multi-platform 3D detection from {\includegraphics[width=0.02\linewidth]{figures/icons/vehicle.png}} \textbf{Vehicle}, {\includegraphics[width=0.021\linewidth]{figures/icons/drone.png}} \textbf{Drone}, and {\includegraphics[width=0.021\linewidth]{figures/icons/quadruped.png}} \textbf{Quadruped}, with fine-grained 3D bounding box annotations, conditions, and practical use cases.}
    \vspace{-0.1cm}
    \resizebox{\linewidth}{!}{
    \begin{tabular}{r|r|ccc|c|c|c|c|cc|c} 
    \toprule
    \multirow{2}{*}{\textbf{Dataset}} & \multirow{2}{*}{\textbf{Venue}} & \multicolumn{3}{c|}{\textbf{Platform}} & \multirow{2}{*}{\textbf{\makecell{\# of\\Frames}}} & \multirow{2}{*}{\textbf{\makecell{LiDAR\\Setup}}} & \multirow{2}{*}{\textbf{Temp.}} & \multirow{2}{*}{\textbf{\makecell{Freq.\\(Hz)}}} & \multicolumn{2}{c|}{\textbf{Condition}} & \multirow{2}{*}{\textbf{Other Sensors Supported}}
    \\
    & & \raisebox{-0.2\height}{\includegraphics[width=0.023\linewidth]{figures/icons/vehicle.png}} & \raisebox{-0.2\height}{\includegraphics[width=0.024\linewidth]{figures/icons/drone.png}}  & \raisebox{-0.2\height}{\includegraphics[width=0.024\linewidth]{figures/icons/quadruped.png}} & & & & & ~\raisebox{-0.2\height}{\includegraphics[width=0.0222\linewidth]{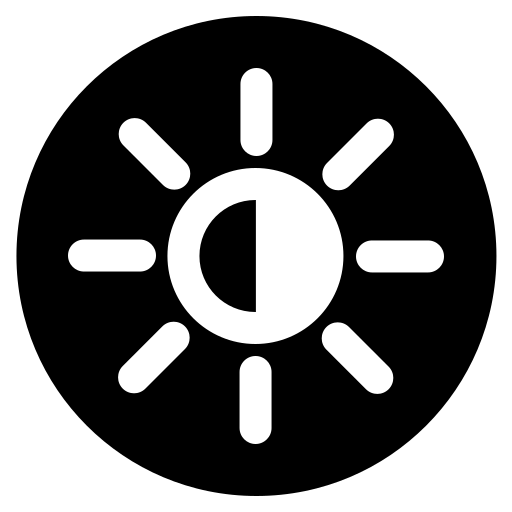}} & \raisebox{-0.2\height}{\includegraphics[width=0.023\linewidth]{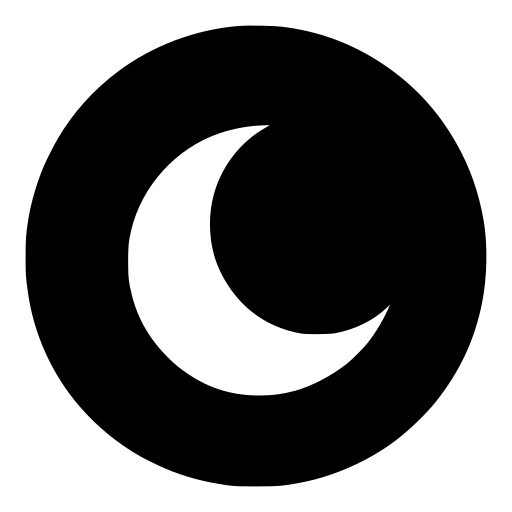}}
    \\\midrule\midrule
    KITTI \cite{geiger2012we} & {\small CVPR'12} & \textcolor{pi3det_blue}{\cmark} & \textcolor{pi3det_red}{\xmark} & \textcolor{pi3det_red}{\xmark} & $14,999$ & $1\times64$ & \textcolor{pi3det_red}{No} & - & ~\textcolor{pi3det_blue}{\cmark} & \textcolor{pi3det_red}{\xmark} & {\raisebox{-0.2\height}{\includegraphics[width=0.022\linewidth]{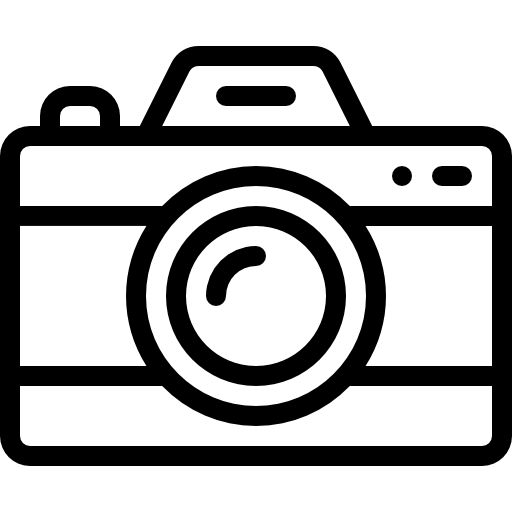}} RGB,~ \raisebox{-0.2\height}{\includegraphics[width=0.021\linewidth]{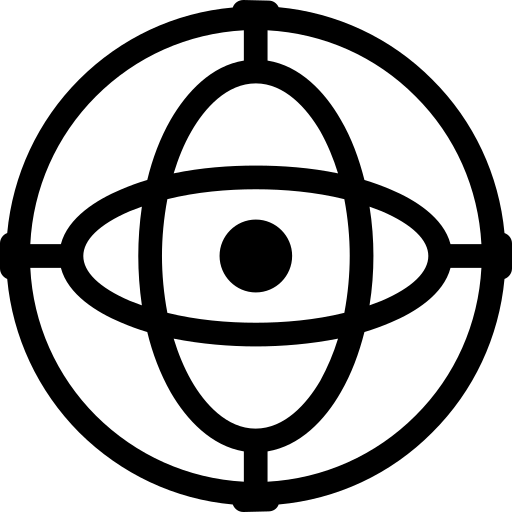}} IMU,~ \raisebox{-0.2\height}{\includegraphics[width=0.02\linewidth]{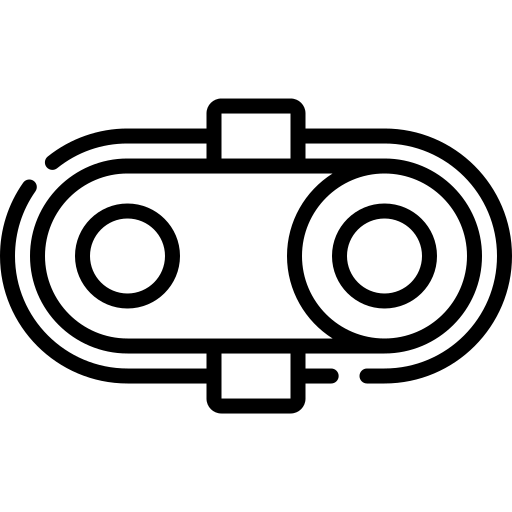}} Stereo}         
    \\
    ApolloScape \cite{8753527} & {\small TPAMI'18} & \textcolor{pi3det_blue}{\cmark} & \textcolor{pi3det_red}{\xmark} & \textcolor{pi3det_red}{\xmark} & $143,906$ & $1\times64$ & \textcolor{pi3det_green}{Yes} & $2$ & ~\textcolor{pi3det_blue}{\cmark} & \textcolor{pi3det_blue}{\cmark} & {\raisebox{-0.2\height}{\includegraphics[width=0.022\linewidth]{figures/icons/camera.png}} RGB,~ \raisebox{-0.2\height}{\includegraphics[width=0.021\linewidth]{figures/icons/imu.png}} IMU,~ \raisebox{-0.2\height}{\includegraphics[width=0.022\linewidth]{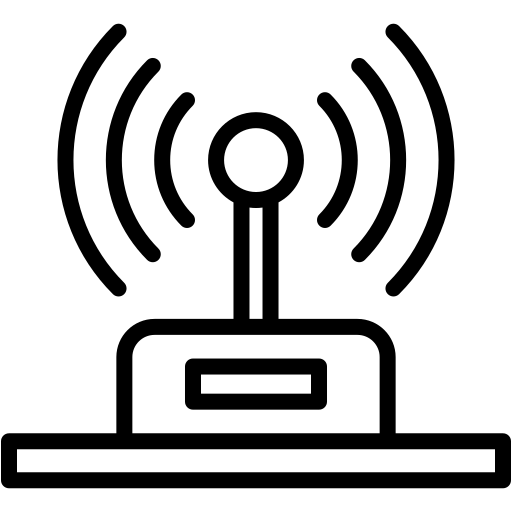}} Radar}
    \\
    Waymo Open \cite{sun2020scalability} & {\small CVPR'19} & \textcolor{pi3det_blue}{\cmark} & \textcolor{pi3det_red}{\xmark} & \textcolor{pi3det_red}{\xmark} & $198,000$ & $1\times64$, $4 \times 16$ & \textcolor{pi3det_green}{Yes} & $10$ & ~\textcolor{pi3det_blue}{\cmark} & \textcolor{pi3det_blue}{\cmark} & {\raisebox{-0.2\height}{\includegraphics[width=0.022\linewidth]{figures/icons/camera.png}} RGB,~ \raisebox{-0.2\height}{\includegraphics[width=0.021\linewidth]{figures/icons/imu.png}} IMU,~ \raisebox{-0.2\height}{\includegraphics[width=0.022\linewidth]{figures/icons/radar.png}} Radar}
    \\
    nuScenes \cite{caesar2020nuscenes} & {\small CVPR'20} & \textcolor{pi3det_blue}{\cmark} & \textcolor{pi3det_red}{\xmark} & \textcolor{pi3det_red}{\xmark} & $35,149$ & $1\times32$ & \textcolor{pi3det_green}{Yes} & $2$ & ~\textcolor{pi3det_blue}{\cmark} & \textcolor{pi3det_blue}{\cmark} & {\raisebox{-0.2\height}{\includegraphics[width=0.022\linewidth]{figures/icons/camera.png}} RGB,~ \raisebox{-0.2\height}{\includegraphics[width=0.021\linewidth]{figures/icons/imu.png}} IMU,~ \raisebox{-0.2\height}{\includegraphics[width=0.022\linewidth]{figures/icons/radar.png}} Radar}
    \\
    % STF \cite{bijelic2020seeing} & {\small CVPR'20} & \textcolor{pi3det_blue}{\cmark} & \textcolor{pi3det_red}{\xmark} & \textcolor{pi3det_red}{\xmark} & $13,500$ & $1\times64$ & \textcolor{pi3det_red}{No} & - & ~\textcolor{pi3det_blue}{\cmark} & \textcolor{pi3det_blue}{\cmark} & {-}
    % \\
    A2D2 \cite{geyer2020a2d2} & {\small arXiv'20} & \textcolor{pi3det_blue}{\cmark} & \textcolor{pi3det_red}{\xmark} & \textcolor{pi3det_red}{\xmark} & $41,277$ & $5\times16$ & \textcolor{pi3det_red}{No} & - & ~\textcolor{pi3det_blue}{\cmark} & \textcolor{pi3det_red}{\xmark} & {\raisebox{-0.2\height}{\includegraphics[width=0.022\linewidth]{figures/icons/camera.png}} RGB,~ \raisebox{-0.2\height}{\includegraphics[width=0.021\linewidth]{figures/icons/imu.png}} IMU}
    \\
    ONCE \cite{mao2021one} & {\small arXiv'21} & \textcolor{pi3det_blue}{\cmark} & \textcolor{pi3det_red}{\xmark} & \textcolor{pi3det_red}{\xmark} & $\sim1$M & $1\times40$ & \textcolor{pi3det_red}{No} & $2$ & ~\textcolor{pi3det_blue}{\cmark} & \textcolor{pi3det_blue}{\cmark} & {\raisebox{-0.2\height}{\includegraphics[width=0.022\linewidth]{figures/icons/camera.png}} RGB,~ \raisebox{-0.2\height}{\includegraphics[width=0.021\linewidth]{figures/icons/imu.png}} IMU}
    \\
    Argoverse 2 \cite{wilson2023argoverse} & {\small NeurIPS'21} & \textcolor{pi3det_blue}{\cmark} & \textcolor{pi3det_red}{\xmark} & \textcolor{pi3det_red}{\xmark} & $\sim6$M & $2\times32$ & \textcolor{pi3det_green}{Yes} & $10$ & ~\textcolor{pi3det_blue}{\cmark} & \textcolor{pi3det_red}{\xmark} & {\raisebox{-0.2\height}{\includegraphics[width=0.022\linewidth]{figures/icons/camera.png}} RGB,~ \raisebox{-0.2\height}{\includegraphics[width=0.021\linewidth]{figures/icons/imu.png}} IMU}
    \\
    aiMotive \cite{matuszka2022aimotive} & {\small ICLRW'23} & \textcolor{pi3det_blue}{\cmark} & \textcolor{pi3det_red}{\xmark} & \textcolor{pi3det_red}{\xmark} & $26,583$ & $1\times64$ & \textcolor{pi3det_green}{Yes} & $10$ & ~\textcolor{pi3det_blue}{\cmark} & \textcolor{pi3det_blue}{\cmark} & {\raisebox{-0.2\height}{\includegraphics[width=0.022\linewidth]{figures/icons/camera.png}} RGB,~ \raisebox{-0.2\height}{\includegraphics[width=0.021\linewidth]{figures/icons/imu.png}} IMU}
    \\
    Zenseact Open \cite{alibeigi2023zenseact} & {\small ICCV'23} & \textcolor{pi3det_blue}{\cmark} & \textcolor{pi3det_red}{\xmark} & \textcolor{pi3det_red}{\xmark} & $\sim100$K & $1\times128$, $4\times16$ & \textcolor{pi3det_green}{Yes} & $1$ & ~\textcolor{pi3det_blue}{\cmark} & \textcolor{pi3det_blue}{\cmark} & {\raisebox{-0.2\height}{\includegraphics[width=0.022\linewidth]{figures/icons/camera.png}} RGB,~ \raisebox{-0.2\height}{\includegraphics[width=0.021\linewidth]{figures/icons/imu.png}} IMU}
    \\
    MAN TruckScenes \cite{fent2025man} & {\small NeurIPS'24} & \textcolor{pi3det_blue}{\cmark} & \textcolor{pi3det_red}{\xmark} & \textcolor{pi3det_red}{\xmark} & $\sim30$K & $6\times64$ & \textcolor{pi3det_green}{Yes} & $2$ & ~\textcolor{pi3det_blue}{\cmark} & \textcolor{pi3det_blue}{\cmark} & {\raisebox{-0.2\height}{\includegraphics[width=0.022\linewidth]{figures/icons/camera.png}} RGB,~ \raisebox{-0.2\height}{\includegraphics[width=0.021\linewidth]{figures/icons/imu.png}} IMU,~ \raisebox{-0.2\height}{\includegraphics[width=0.022\linewidth]{figures/icons/radar.png}} Radar}
    \\
    AeroCollab3D \cite{tian2024ucdnet} & {\small TGRS'24} & \textcolor{pi3det_red}{\xmark} & \textcolor{pi3det_blue}{\cmark} & \textcolor{pi3det_red}{\xmark} & $3,200$ & N/A & \textcolor{pi3det_red}{No} & - & ~\textcolor{pi3det_blue}{\cmark} & \textcolor{pi3det_red}{\xmark} & {\raisebox{-0.2\height}{\includegraphics[width=0.022\linewidth]{figures/icons/camera.png}} RGB,~ \raisebox{-0.2\height}{\includegraphics[width=0.021\linewidth]{figures/icons/imu.png}} IMU}          
    \\\midrule
    \cellcolor{pi3det_blue!9}\textbf{Pi3DET} {\small(M3ED)} & \cellcolor{pi3det_blue!9}\textbf{Ours} & \cellcolor{pi3det_blue!9}\textcolor{pi3det_blue}{\cmark} & \cellcolor{pi3det_blue!9}\textcolor{pi3det_blue}{\cmark} & \cellcolor{pi3det_blue!9}\textcolor{pi3det_blue}{\cmark} & \cellcolor{pi3det_blue!9}$\mathbf{51,545}$ & \cellcolor{pi3det_blue!9}$\mathbf{1\times64}$ & \cellcolor{pi3det_blue!9}\textcolor{pi3det_green}{\textbf{Yes}} & \cellcolor{pi3det_blue!9}$\mathbf{10}$ & ~\cellcolor{pi3det_blue!9}\textcolor{pi3det_blue}{\cmark} & \cellcolor{pi3det_blue!9}\textcolor{pi3det_blue}{\cmark} & \cellcolor{pi3det_blue!9}{\raisebox{-0.2\height}{\includegraphics[width=0.022\linewidth]{figures/icons/camera.png}} \textbf{RGB}, \raisebox{-0.2\height}{\includegraphics[width=0.021\linewidth]{figures/icons/imu.png}} \textbf{IMU}, \raisebox{-0.2\height}{\includegraphics[width=0.02\linewidth]{figures/icons/stereo.png}} \textbf{Stereo}, \raisebox{-0.2\height}{\includegraphics[width=0.021\linewidth]{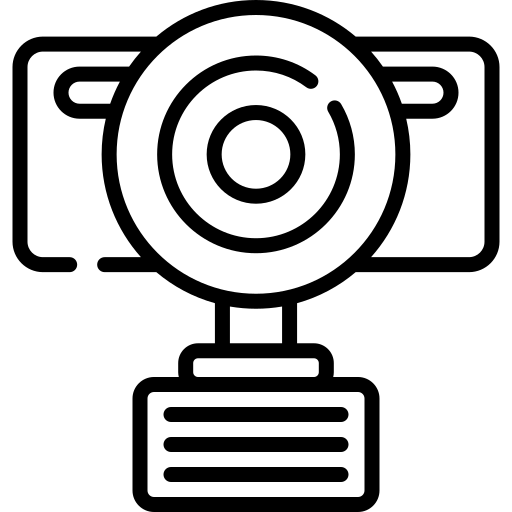}} \textbf{Event}}
    \\
    \bottomrule
    \end{tabular}}
    \label{tab:dataset-compare}
\end{table*}

Extensive experiments on KITTI \cite{geiger2012we}, nuScenes \cite{caesar2020nuscenes}, and our \textbf{Pi3DET} validate our effectiveness. Specifically, Pi3DET-Net achieves mAP gains of $+11.84\%$ and $+12.03\%$ in Vehicle $\rightarrow$ Drone and Vehicle $\rightarrow$ Quadruped adaptations, respectively. Additionally, cross-dataset experiments show an average improvement of $+25.27\%$ mAP over source-only methods in the nuScenes $\rightarrow$ KITTI scenario. We further establish a \textbf{comprehensive benchmark} on Pi3DET with $18$ state-of-the-art detectors, identifying insights to enhance resilience against platform variations. When combined with these detectors, our method consistently boosts performance, underscoring its architecture-agnostic nature and wide applicability.

In summary, the contributions of this work are:
\begin{itemize} 
    \item We introduce \textbf{Pi3DET}, a diverse and large-scale multi-platform 3D object detection dataset, serving as a solid foundation for cross-platform 3D detection research.
    
    \item We propose a novel cross-platform 3D object detection framework, \textbf{Pi3DET-Net}, to effectively transfer 3D detection capabilities from vehicles to other platforms by integrating geometric and feature-level representations. 
    
    \item We establish an extensive benchmark, providing crucial insights for future development of generalizable 3D detection systems across heterogeneous robot platforms. To our knowledge, this is the first work in this line of research.
\end{itemize}
\section{Related Work}
\label{sec:related_work}

\textbf{Datasets \& Benchmarks for 3D Detection.} 
LiDAR-based 3D detection aims to estimate an object’s 3D position and geometric dimensions \cite{mao20233d, qian20223d, wang2023multi}. Typical detectors are classified by their approach to process point cloud data: grid-based (using voxels \cite{deng2021voxel,li2022unifying, mao2021voxel, lee2024re}, range grids \cite{fan2021rangedet, zhang2022ri, tian2022fully} and BEV grids \cite{wang2024club, song2024graphbev, li2024bevnext}, pillars \cite{shi2022pillarnet, wang2020pillar, li2023pillarnext}, or cylindrical partitions \cite{chen2020every, rapoport2021s, zhu2021cylindrical}), point-based (directly learning features from raw points \cite{qi2017pointnet++, zhang2022not, yang20203dssd, yang2022dbq}), or hybrid point-grid \cite{song2022jpv, shishaoshuai2020pv, shi2023pv, li2022voxel}, which often delivers state-of-the-art results but at higher computational cost. Datasets such as KITTI \cite{geiger2012we}, nuScenes \cite{caesar2020nuscenes}, Waymo Open \cite{sun2020scalability}, and others \cite{8753527, mao2021one, wilson2023argoverse, bijelic2020seeing, xie2025drivebench} have driven progress in accuracy \cite{shishaoshuai2020pv, li2024bevnext}, robustness \cite{song2024robustness,dong2023benchmarking,kong2023robo3d,hao2024is,kong2024robodrive}, and efficiency \cite{ye2020hvnet, yang2018pixor}. Yet, most research targets vehicle-mounted sensors, leaving quadrupeds and drones underexplored despite similar LiDAR payloads. To address this gap, we present \textbf{Pi3DET}, the \textbf{first} publicly available dataset incorporating heterogeneous data from multi-platform setups for 3D object detection.

\noindent\textbf{Cross-Dataset 3D Detection.}
Prior work transfers knowledge often in cross-dataset settings. ST3D \cite{yang2021st3d} and ST3D++ \cite{yang2022st3d++} introduced a three-stage approach (pretraining, pseudo-labeling, and self-training) to improve generalization on target data. Further work refines pseudo-label accuracy \cite{zhang2021srdan, zhang2024detect, chang2024cmda, zhang2024pseudo, tsai2024ms3d++} and self-training guidance \cite{zhu2019adapting, yuan2024reg}, or leverages unified training sets \cite{deng2024cmd, zhang2023uni3d} and knowledge distillation \cite{yang2022towards, hu2023density, yuzhe2025vexkd}. However, most ignore the more challenging cross-platform scenario. While Wozniak \emph{et al.} \cite{wozniak2023applying} highlight its importance, they lack a suitable dataset for vehicle-to-other-platform experiments. In contrast, we analyze platform-level shifts and propose the first method tailored for cross-platform transfers. Building on \textbf{Pi3DET}, we validate its effectiveness on genuine multi-platform data.

\noindent\textbf{Auto-Labeling 3D Object Detection.} 
Accurate point cloud annotations are crucial for 3D detection, yet labeling a single point cloud can take over $100$ seconds \cite{zhou2024openannotate3d}. To reduce this burden, researchers have explored semi-automated \cite{wu2024efficient,liu2023seal} and fully-automated \cite{zhang2023towards} approaches, including active learning \cite{feng2019deep,ghita2024activeanno3d,yuan2023bi3d,samet2023seeding}, weak supervision \cite{meng2020weakly,zhang2024general,li2024rapid-seg,michele2024train}, and pseudo-label refinement \cite{tsai2024ms3d++,fan2023once,yang2023labelformer, chen2020every,chen2023towards,boulch2023also,li2023lim3d}. Recent works integrate vision–language models \cite{zhou2024openannotate3d,zhou2024openannotate2,zhang2024general,liu2025m3net,xu2025limoe,xu2024superflow} for greater efficiency. However, these methods primarily target vehicle-mounted platforms. In contrast, we design \textbf{Pi3DET-Net} to address multi-platform auto-labeling, including quadruped and drones, to advance 3D object detection in broader operational scenarios.

\begin{figure}[t]
    \begin{center}
    \includegraphics[width=\textwidth]{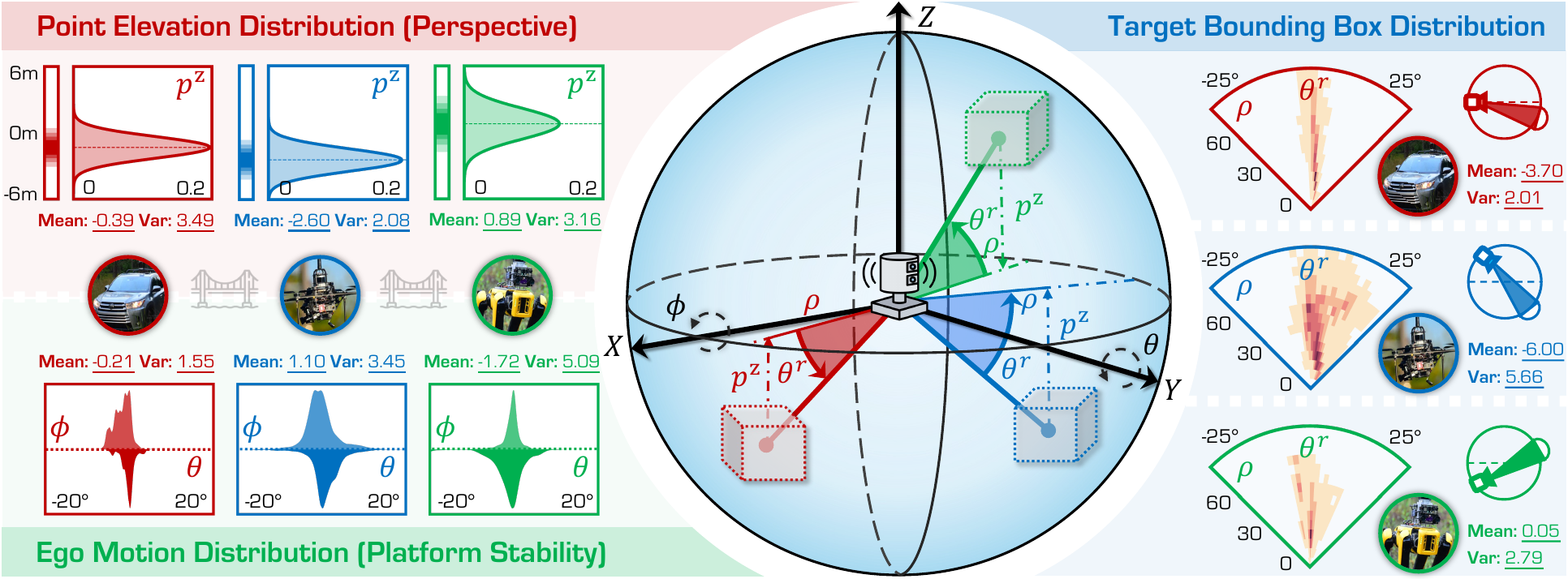}
    \end{center}
    \vspace{-0.1cm}
    \caption{Analysis of perspective differences across three robot platforms. We present the statistics of point elevation distribution (\textbf{upper-left}), ego motion distribution (\textbf{bottom-left}), and target bounding box distribution (\textbf{right}), along with means and variances for each platform's data. We use different colors to denote different platforms for simplicity, \emph{i.e.}, {\includegraphics[width=0.02\linewidth]{figures/icons/vehicle.png}} \textcolor{pi3det_red}{\textbf{Vehicle}}, {\includegraphics[width=0.021\linewidth]{figures/icons/drone.png}} \textcolor{pi3det_blue}{\textbf{Drone}}, and {\includegraphics[width=0.021\linewidth]{figures/icons/quadruped.png}} \textcolor{pi3det_green}{\textbf{Quadruped}}. Best viewed in colors.}
    \label{fig:distribution_comparsion}
\end{figure}
\section{Pi3DET: Dataset \& Benchmark}
\label{sec:dataset}

\subsection{Motivation}
While existing LiDAR-based 3D detection datasets predominantly focus on vehicle data, their utility diminishes for other platforms (\emph{e.g.}, drones and quadrupeds) due to diverging operational perspectives. To address this limitation, we introduce \textbf{Pi3DET} (\underline{\textbf{P}}erspective \underline{\textbf{i}}nvariant \underline{\textbf{3}}D object \underline{\textbf{DET}}ection), the first multi-platform dataset for LiDAR-based 3D object detection. Built upon M3ED \cite{chaney2023m3ed}, Pi3DET provides annotated LiDAR sequences across  \includegraphics[width=0.023\linewidth]{figures/icons/vehicle.png} \textbf{Vehicle}, \includegraphics[width=0.023\linewidth]{figures/icons/drone.png} \textbf{Drone}, and \includegraphics[width=0.023\linewidth]{figures/icons/quadruped.png} \textbf{Quadruped}, specifically designed to advance research in multi-platform 3D object detection.

\subsection{Dataset Statistics}
Our \textbf{Pi3DET} benchmark spans $25$ sequences collected from vehicle, quadruped, and drone platforms, annotated at $10$ Hz. Compared to other datasets in Table~\ref{tab:dataset-compare}, Pi3DET provides $\mathbf{51{,}545}$ \textbf{frames} and more than $\mathbf{250{,}000}$ \textbf{box annotations} across two object categories (\textit{Vehicle} and \textit{Pedestrian}), covering day/night conditions in urban, suburban, and rural areas. We combine an automated labeling pipeline with extensive manual refinement, requiring about $\mathbf{500}$ \textbf{hours} of human effort. For additional details on the annotation process, dataset statistics, and examples, please refer to Appendix~\ref{sec_supp:dataset_statistics}.

\subsection{Perspective Discrepancies Analysis}
\label{sec:Discrepancies Analysis}

To quantify cross-platform gaps, we first formalize the problem setup and analyze \textbf{geometric discrepancies} across three platforms. We define a point cloud as $\mathcal{P}^{\beta} = \{\mathbf{p}_{i} \}_{i=1}^{N^{\beta}}$, and a single point\footnote{For simplicity, we use $\mathbf{p}$ to represent a point from a point cloud, rather than explicitly referencing each individual sample from the point set. The same applies to the 3D bounding boxes.} from the set as $\mathbf{p} = (p^{x}, p^{y}, p^{z}) \in \mathbb{R}^3$, $\beta$ denotes the platform, including vehicles, drones, and quadrupeds, and $N^{\beta}$ is the number of point clouds for platform $\beta$. The 3D bounding boxes are denoted by $\mathcal{B}^{\beta} = \{\mathbf{b}_{j} \}_{j=1}^{M^{\beta}}$.  We denote one bounding box from this set as $\mathbf{b} = (c^{x}, c^{y}, c^{z}, l, w, h, \varphi) \in \mathbb{R}^7$. Here, \(\mathbf{c}=(c^{x}, c^{y}, c^{z})\) represents the bounding box center, \((l, w, h)\) the dimensions, \(\varphi\) the heading angle, and $M^{\beta}$ is the number of bounding box. Additionally, the ego pose is given by a transformation \(\mathbf{T} \in \text{SE}(3)\), decomposed into a rotation matrix \(\mathbf{R} \in \text{SO}(3)\) (parameterized by Euler angle \(\phi\), \(\theta\), and \(\psi\) for roll, pitch, yaw) and a translation vector $\mathbf{t} = [t^x,t^y,t^z]$. We further define the distance between the target bounding box and the ego platform in bird’s-eye view (BEV) as $\rho$, and denote the relative pitch from the bounding box to the ego platform in the ego coordinate system as $\theta^{r}$. As shown in Figure~\ref{fig:distribution_comparsion}, we identify \textbf{three} critical cross-platform discrepancies.

\noindent\textbf{Ego Motion Distributions.} Vehicle-mounted LiDAR sensors exhibit stable motion with minimal roll/pitch variance ($\phi, \theta<5^{\circ}$). In contrast, drones and quadrupeds suffer significant ego jitter due to dynamic locomotion and aerodynamics, inducing roll/pitch fluctuations up to $20^{\circ}$, shown in the bottom-left part in Figure~\ref{fig:distribution_comparsion}. This instability introduces high-frequency perturbations in point cloud geometry.

\noindent \textbf{Point Elevation Distributions.}
Beyond the roll and pitch jitter caused by ego motion, the overall distribution of the elevation $p^z$ of the input point cloud varies significantly among the platforms due to their different intrinsic heights. As shown in the upper-left in Figure~\ref{fig:distribution_comparsion}, for vehicles, most points lie slightly below their own height ($p^z<t^z$). In contrast, on quadrupeds, the points cluster above the height of the platform ($p^z\ > t^z$), while for drones, the points are distributed substantially lower than the drone’s altitude ($p^z<<t^z$).

\noindent \textbf{Target Bounding Box Distributions.} 
Variations in platform height influence the relative orientation of the detected object. The right part of Figure~\ref{fig:distribution_comparsion} shows the relationship between targets' relative pitch angles $\theta^{r}$ and BEV distances $\rho$. Comparatively, drones observe objects with larger downward pitch angles and large variances, indicating that targets are positioned lower relative to the ego platform with a more uneven distribution. In contrast, quadrupeds exhibit larger upward pitch angles, suggesting that objects are relatively higher in their view. Vehicles, benefiting from stable motion, display the smallest variance in pitch angle distribution.

These discrepancies make single-platform models ineffective for cross-platform deployment. Training separate models for each platform is resource-intensive and impractical for real-world scalability. Instead, we aim to propose a unified cross-platform adaptation framework that trains on large-scale readily available source platform data ($\textcolor{pi3det_red}{\mathcal{S}}$, \emph{e.g.}, vehicle) and generalizes to target platform data ($\textcolor{pi3det_blue}{\mathcal{T}}$) without target labels, addressing geometric shifts through perspective-invariant learning.
\section{Methodology}
\label{sec:method}

\begin{figure}[t]
    \centering
    \includegraphics[width=\textwidth]{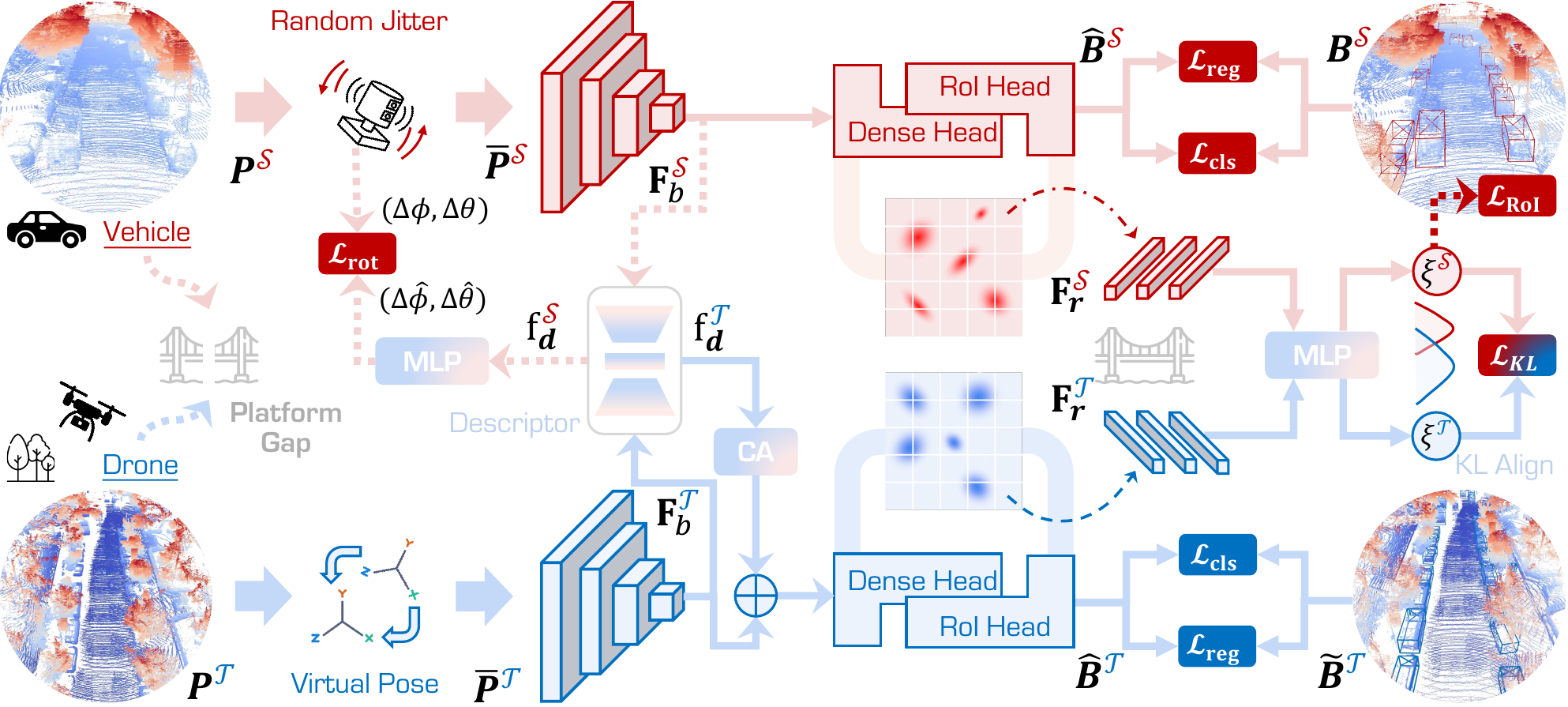}
    \vspace{-0.5cm}
    \caption{
    \textbf{Framework Overview.} The proposed \textbf{Pi3DET-Net} consists of two main stages: \textit{Pre-Adaption (PA)} and \textit{Knowledge-Adaption (KA)}, aiming at bridging the gap across heterogeneous robot platforms through alignment at both geometric (Section~\ref{sec:geometry alignment}) and feature levels (Section~\ref{sec:feature alignment}). On the geometric side, PA employs \textit{Random Platform Jitter} to enhance robustness against ego-motion variations, while KA uses \textit{Virtual Platform Pose} to simulate source-like viewpoints to achieve bidirectional geometric alignment across platforms. On the feature side, Pi3DET-Net further incorporates \textit{KL Probabilistic Feature Alignment} to align target features with the source space, along with a \textit{Geometry-Aware Transformation Descriptor} to correct global transformations across platforms.
    }
    \label{fig:method}
\end{figure}

As illustrated in Figure~\ref{fig:method}, we propose a two-stage \textbf{Pi3DET-Net} consisting of \textit{Pre-Adaption (PA)} and \textit{Knowledge-Adaption (KA)} for cross-platform adaptation. For geometric alignment (Section~\ref{sec:geometry alignment}), Random Platform Jitter facilitates robustness against ego-motion variations, while Virtual Platform Pose aligns viewpoints. For feature alignment (Section~\ref{sec:feature alignment}), {KL Probabilistic Feature Alignment} aligns target features with the source space, and a {Geometry-Aware Transformation Descriptor} corrects global transformations across platforms. The training pipeline is illustrated in Section~\ref{sec:training}.

\subsection{Cross-Platform Geometry Alignment}
\label{sec:geometry alignment}

As outlined in Section~\ref{sec:Discrepancies Analysis}, platform-induced point cloud discrepancies arise from varying ego motions, point elevations, and target bounding box distributions. To mitigate these, we propose two complementary strategies. First, we apply {Random Platform Jitter} during {PA} on the source platform, enhancing robustness to pose jitter. Second, we use a {Virtual Platform Pose} in {KA} on the target platform to achieve effective scene alignment. Together, these approaches enable smoother geometric adaptation from source to target.

\noindent\textbf{Random Platform Jitter (RPJ).} 
To emulate the roll and pitch jitters observed on quadruped and drone platforms, we introduce Random Platform Jitter during PA on the source platform. Specifically, we sample two angles \(\Delta \phi\) and \(\Delta \theta\) from a uniform distribution for roll and pitch, and define a composite rotation \(\mathbf{R}(\Delta \phi,\Delta \theta)\). For point \(\mathbf{p}\in \mathcal{P}^{\textcolor{pi3det_red}{S}}\), bounding-box \(\mathbf{b}\in \mathcal{B}^{\textcolor{pi3det_red}{S}}\) and its center $\mathbf{c}$, we have:
\begin{equation}
    \bar{\mathbf{p}} = \mathbf{R}(\Delta \phi, \Delta \theta)\,\mathbf{p}~, \quad \bar{\mathbf{c}} = \mathbf{R}(\Delta \phi, \Delta \theta)\,\mathbf{c}~.
\end{equation}
Here, the box dimensions are unchanged, and the heading angle is preserved. The transformed point cloud $\mathcal{\bar{P}}^{\textcolor{pi3det_red}{S}}$ is then input into the backbone for feature extraction. Exposing the model to these rotated point cloud inputs tends to enhance the robustness to roll-pitch variations on target platforms.

\noindent\textbf{Virtual Platform Pose (VPP).} We establish a virtual pose on the target platform during KA to mimic the source viewpoint and reduce the platform geometry gap. Since input point cloud and bounding box distributions diverge, we define a virtual pose \(\bar{\mathbf{T}}\) from the actual ego pose \(\mathbf{T}\). We set roll and pitch to zero (\(\bar{\phi}=0, \bar{\theta}=0\)), keep the actual yaw (\(\bar{\psi}=\psi\)), and preserve planar coordinates \((\bar{t}^x = t^x, \bar{t}^y = t^y)\), fixing the height at \(\bar{t}^z=t^z_{\mathrm{vehicle}}\). Given a point cloud \(\mathbf{p}\in \mathcal{P}^{\textcolor{pi3det_blue}{\mathcal{T}}}\) from target platform, along with the bounding box  \(\mathbf{b}\in \mathcal{B}^{\textcolor{pi3det_blue}{\mathcal{T}}}\) and its center $\mathbf{c}$, we express them in homogeneous coordinates $\mathbf{P}, \mathbf{C}$, and then transform them to the following:
\begin{equation}
{\bar{\mathbf{P}}} = \bar{\mathbf{T}}\,{\mathbf{T}}^{-1}\,{\mathbf{P}}~, \quad {\bar{\mathbf{C}}} = \bar{\mathbf{T}}\,{\mathbf{T}}^{-1}\,{\mathbf{C}}~.
\end{equation}
Here, dimensions remain unchanged, while the heading \(\varphi\) is offset by \(\Delta(\bar{\psi}, \psi)\). The resulting point cloud $\mathcal{\bar{P}}^{\textcolor{pi3det_blue}{\mathcal{T}}}$ is used for feature extraction. Transforming both point clouds and bounding boxes to this virtual coordinate frame mitigates platform gaps and improves cross-platform adaptations.

\subsection{Cross-Platform Feature Alignment}
\label{sec:feature alignment}

To address domain shifts across platforms, we leverage both probabilistic modeling and global geometric cues to align cross-platform features. As illustrated in Figure~\ref{fig:method}, our feature alignment consists of two key components: \textbf{1)} a transformation descriptor that learns global geometric invariance; and \textbf{2)} a probabilistic feature alignment guided by KL divergence.

\noindent\textbf{Geometry-Aware Transformation Descriptor (GTD).}
As discussed in Section~\ref{sec:Discrepancies Analysis}, differing ego-motion distributions cause global shifts in source and target point clouds. We address these by learning a geometry-aware descriptor on the source platform, then applying it to correct transformations on the target. During PA, we apply global max-pooling to the backbone’s feature $\mathbf{F}_{b}^{\textcolor{pi3det_red}{\mathcal{S}}}$ to obtain a compact vector, which is encoded by a hierarchical convolutional module into a large-scale geometric descriptor $\mathbf{f}_{d}^{\textcolor{pi3det_red}{\mathcal{S}}}$. A small regression $\mathrm{MLP}$ then predicts the artificially introduced random jitter angles $(\Delta\hat\theta, \Delta\hat\phi)$ from this descriptor, optimizing the following rotation loss:
\begin{equation}
\mathcal{L}_{\mathrm{rot}}
=
\|\Delta\hat{\phi} - \Delta \phi\|^2
+
\|\Delta\hat{\theta} - \Delta \theta\|^2~.
\end{equation}
Notably, minimizing \(\mathcal{L}_{\mathrm{rot}}\) equips the network with platform-agnostic transformation cues. This descriptor, learned on the source platform, corrects global offsets on the target platform during KA, ensuring robust cross-platform performance.

\noindent{\textbf{KL Probabilistic Feature Alignment (PFA).}}
We aim to reduce cross-platform discrepancies by matching the Region-of-Interest (RoI) feature distributions of source and target platforms during KA. 

Specifically, we approximate each platform’s RoI features before the detection head with a probabilistic method, ensuring robust distribution alignment. For source-platform RoI feature \(\mathbf{F}_r^{\textcolor{pi3det_red}{\mathcal{S}}}\), a probabilistic encoder $p(\mathbf{\xi}^{\textcolor{pi3det_red}{\mathcal{S}}}|\mathbf{F}_r^{\textcolor{pi3det_red}{\mathcal{S}}}) = \mathcal{N}\bigl(\boldsymbol{\mu}(\mathbf{F}_r^{\textcolor{pi3det_red}{\mathcal{S}}}), \boldsymbol{\sigma}^2(\mathbf{F}_r^{\textcolor{pi3det_red}{\mathcal{S}}})\bigr)$ maps this feature into a Gaussian distribution, which predicts $\boldsymbol{\mu}(\mathbf{F}_r^{\textcolor{pi3det_red}{\mathcal{S}}})$ and $\boldsymbol{\sigma}^2(\mathbf{F}_r^{\textcolor{pi3det_red}{\mathcal{S}}})$ with $\mathrm{MLP}$s. Using the reparameterization trick~\cite{kingma2022autoencodingvariationalbayes}, latent samples $\mathbf{\xi}^{\textcolor{pi3det_red}{\mathcal{S}}} = \boldsymbol{\mu}(\mathbf{F}_r^{\textcolor{pi3det_red}{\mathcal{S}}})+\boldsymbol{\sigma}(\mathbf{F}_r^{\textcolor{pi3det_red}{\mathcal{S}}})\odot\boldsymbol{\epsilon}$ are generated ($\boldsymbol{\epsilon} \,\sim\, \mathcal{N}(\mathbf{0}, \mathbf{I})$). Analogous encoding applies to the target-platform RoI feature \(\mathbf{F}_r^{\textcolor{pi3det_blue}{\mathcal{T}}}\), producing latent samples \(\mathbf{\xi}^{\textcolor{pi3det_blue}{\mathcal{T}}}\) accordingly.

Since the true distribution of latent features is unknown, we can only estimate it from latent samples on both platforms. By comparing these samples via the KL term, we have:
\begin{equation}
    \mathcal{L}_\mathrm{KL} 
    \;=\; 
    D_\mathrm{KL}\Bigl[p(\mathbf{\xi}^{\textcolor{pi3det_red}{\mathcal{S}}} \mid \mathbf{F}_r^{\textcolor{pi3det_red}{\mathcal{S}}})
    \,\Big\|\,
    p(\mathbf{\xi}^{\textcolor{pi3det_blue}{\mathcal{T}}} \mid \mathbf{F}_r^{\textcolor{pi3det_blue}{\mathcal{T}}})
    \Bigr]~.
\end{equation}

The model pushes the target platform’s features toward the source manifold. Crucially, this nonadversarial approach provides a stable alignment in the absence of direct target supervision. As investigated by \cite{nguyen2021kl}, the KL objective not only prevents out-of-distribution samples but also offers a mode-seeking alignment, ultimately improving target performance. For the source platform, we also train a classification head \(q(\mathbf{g}|\mathbf{\xi})\) to discriminate foreground from background:
\begin{equation}
\mathcal{L}_\mathrm{RoI} 
\;=\; 
\mathbb{E}_{\mathbf{\xi}^{\textcolor{pi3det_red}{\mathcal{S}}}\sim p(\mathbf{\xi}^{\textcolor{pi3det_red}{\mathcal{S}}}\mid F_r^{\textcolor{pi3det_red}{\mathcal{S}}})}
\!\bigl[-\log q(\mathbf{g}^{\textcolor{pi3det_red}{\mathcal{S}}} \mid \mathbf{\xi}^{\textcolor{pi3det_red}{\mathcal{S}}})\bigr]~,
\end{equation}
where $\mathbf{g}^{\textcolor{pi3det_red}{\mathcal{S}}}$ is the classification task ground truth. This loss ensures the latent representation \(\mathbf{\xi}^{\textcolor{pi3det_red}{\mathcal{S}}}\) captures semantic features in the source platform for effective alignment through $\mathcal{L}_\mathrm{KL}$.

\subsection{Objective \& Optimization}
\label{sec:training}
The overall framework aims to learn global transformations and semantic cues during \textit{Pre-Adaptation}, then propagate and align target data during \textit{Knowledge-Adaptation}.

\noindent\textbf{Pre-Adaptation (PA).} In the source platform, our goal is to extract and internalize the necessary knowledge while enhancing geometric robustness through Random Platform Jitter, addressing platform-specific discrepancies through the rotation loss $\mathcal{L}_\mathrm{rot}$. and learning RoI-based semantic features via $\mathcal{L}_\mathrm{RoI}$. We also apply a standard detection loss composed of a classification loss and a bounding-box regression loss:
\begin{equation}
        \mathcal{L}_\mathrm{det} = \mathcal{L}_\mathrm{cls}(\hat{\mathcal{B}}^{\textcolor{pi3det_red}{\mathcal{S}}}, \mathcal{B}^{\textcolor{pi3det_red}{\mathcal{S}}}) + \mathcal{L}_\mathrm{reg}(\hat{\mathcal{B}}^{\textcolor{pi3det_red}{\mathcal{S}}}, \mathcal{B}^{\textcolor{pi3det_red}{\mathcal{S}}})~,
\end{equation}
where $\hat{\mathcal{B}}^{\textcolor{pi3det_red}{\mathcal{S}}}$ denotes the predicted bounding box. The overall pre-adaptation objective is:
$\mathcal{L}_\mathrm{PA} = \mathcal{L}_\mathrm{det} + \lambda_\mathrm{rot}\mathcal{L}_\mathrm{rot} + \lambda_\mathrm{RoI}\mathcal{L}_\mathrm{RoI}$,
where $\lambda_\mathrm{rot}$ and $\lambda_\mathrm{RoI}$ are weights used to balance the losses. This step trains a robust 3D detector while imparting global geometric awareness for adaptation.

\noindent\textbf{Knowledge-Adaptation (KA).} After PA, we first use the source-platform knowledge to generate pseudo-annotations $\tilde{\mathcal{B}}^{\textcolor{pi3det_blue}{\mathcal{T}}}$ on target data, then train jointly on both platforms:
\begin{itemize}
    \item \textbf{Source Platform:} To preserve source performance, we disable \(\mathcal{L}_\mathrm{rot}\) and optimize only detection and RoI classification, \emph{i.e.},
    $
        \mathcal{L}^{\textcolor{pi3det_red}{\mathcal{S}}}_\mathrm{KA} = \mathcal{L}^{\textcolor{pi3det_red}{\mathcal{S}}}_\mathrm{det}+\lambda_\mathrm{RoI}\mathcal{L}^{\textcolor{pi3det_red}{\mathcal{S}}}_\mathrm{RoI}~.
    $
    \item \textbf{Target Platform:} We encode the learned global descriptor $\mathbf{f}_{d}^{\textcolor{pi3det_blue}{\mathcal{T}}}$ with channel attention (\emph{i.e.}, CA in Figure~\ref{fig:method}) and add it to the backbone features as a residual offset, enforce a detection loss, and align RoI features via KL. This process can be formulated as: 
    $
        \mathcal{L}^{\textcolor{pi3det_blue}{\mathcal{T}}}_\mathrm{KA} = \mathcal{L}^{\textcolor{pi3det_blue}{\mathcal{T}}}_\mathrm{det}+\lambda_\mathrm{KL}\mathcal{L}_\mathrm{KL}~,
    $ where $\lambda_\mathrm{KL}$ is used to balance the KL loss.
\end{itemize}
The combined objective is $\mathcal{L}_\mathrm{KA} = \mathcal{L}^{\textcolor{pi3det_blue}{\mathcal{T}}}_\mathrm{KA}+ \mathcal{L}^{\textcolor{pi3det_red}{\mathcal{S}}}_\mathrm{KA}$. By decoupling geometry learning (during PA) from feature correction (during KA), the geometry-aware transformation descriptor remains focused on platform-induced differences. Meanwhile, RoI feature alignment pulls target features toward the source distribution, narrowing the cross-platform gap and enabling accurate 3D detection on target platforms.

\begin{table*}
    \centering
    \caption{
        \textbf{Comparisons of 3D detection methods for vehicle$\rightarrow$drone/quadruped tasks.} We report the average precision (AP) in ``BEV / 3D'' at the IoU thresholds of $0.7$ and $0.5$, respectively. Symbol $\ddagger$ denotes algorithms \textit{w.o.} ROS \cite{yang2021st3d}. All scores are given in percentage (\%). ``-'' denotes the code is not available. The \textcolor{pi3det_red}{\hlred{Best}} and \textcolor{pi3det_blue}{\hlblue{Second Best}} scores under each metric are highlighted in \textcolor{pi3det_red}{\hlred{Red}} and \textcolor{pi3det_blue}{\hlblue{Blue}}, respectively.
    }
    \vspace{-0.1cm}
    \resizebox{\linewidth}{!}{
    \begin{tabular}{c|l|cc|cc|cc|cc|cc} 
    \toprule
    \multirow{3.5}{*}{\textbf{\#}} & \multirow{3.5}{*}{\textbf{Method}} & \multicolumn{4}{c|}{\raisebox{-0.2\height}{\includegraphics[width=0.024\linewidth]{figures/icons/vehicle.png}} \textbf{Vehicle} ~~$\rightarrow$~~ \raisebox{-0.2\height}{\includegraphics[width=0.025\linewidth]{figures/icons/quadruped.png}} \textbf{Quadruped}} & \multicolumn{4}{c|}{\raisebox{-0.2\height}{\includegraphics[width=0.024\linewidth]{figures/icons/vehicle.png}} \textbf{Vehicle} ~~$\rightarrow$~~ \raisebox{-0.2\height}{\includegraphics[width=0.025\linewidth]{figures/icons/drone.png}} \textbf{Drone}} & \multicolumn{2}{c}{\multirow{2.5}{*}{\textbf{Average}}}
    \\\cmidrule{3-10}
    & & \multicolumn{2}{c|}{\textbf{PV-RCNN} \cite{shishaoshuai2020pv}} & \multicolumn{2}{c|}{\textbf{Voxel RCNN} \cite{deng2021voxel}} & \multicolumn{2}{c|}{\textbf{PV-RCNN} \cite{shishaoshuai2020pv}} & \multicolumn{2}{c|}{\textbf{Voxel RCNN} \cite{deng2021voxel}} & 
    \\
    & & \textbf{AP@0.7} & \textbf{AP@0.5} & \textbf{AP@0.7} & \textbf{AP@0.5} & \textbf{AP@0.7} & \textbf{AP@0.5} & \textbf{AP@0.7} & \textbf{AP@0.5} & \textbf{AP@0.7} & \textbf{AP@0.5}
    \\\midrule\midrule
    \multirow{10}{*}{\rotatebox{90}{\textbf{nuScenes \cite{caesar2020nuscenes}}}} & \cellcolor{gray!5}\textcolor{gray}{Source Platform} & \cellcolor{gray!5}\textcolor{gray}{$43.40$ / $33.55$} & \cellcolor{gray!5}\textcolor{gray}{$44.86$ / $42.84$} & \cellcolor{gray!5}\textcolor{gray}{$43.25$ / $33.74$} & \cellcolor{gray!5}\textcolor{gray}{$45.62$ / $43.32$} & \cellcolor{gray!5}\textcolor{gray}{$50.91$ / $35.26$} & \cellcolor{gray!5}\textcolor{gray}{$57.73$ / $50.24$} & \cellcolor{gray!5}\textcolor{gray}{$50.15$ / $29.41$} & \cellcolor{gray!5}\textcolor{gray}{$57.10$ / $49.10$} & \cellcolor{gray!5}\textcolor{gray}{$46.93$ / $32.99$} & \cellcolor{gray!5}\textcolor{gray}{$51.33$ / $46.34$}
    \\\cmidrule{2-12}
    & ST3D \cite{yang2021st3d} & $55.40$ / $42.02$ & $59.59$ / $54.75$ & $44.54$ / $35.96$ & $45.81$ / $44.38$ & $65.05$ / $40.01$ & $68.93$ / $64.09$ & $54.62$ / $33.79$ & $58.45$ / $52.89$ & $54.90$ / $37.95$ & $58.20$ / $54.03$
    \\
    & ST3D$^{\ddagger}$ \cite{yang2021st3d} & $55.68$ / \textcolor{pi3det_blue}{\hlblue{$44.50$}} & $59.32$ / $55.32$ & $45.01$ / $37.13$ & $46.73$ / $45.45$ & $65.40$ / $43.63$ & \textcolor{pi3det_blue}{\hlblue{$69.24$}} / $64.88$ & $55.23$ / $36.51$ & $59.30$ / $54.23$ & $55.33$ / $40.44$ & $58.65$ / $54.97$
    \\
    & ST3D++ \cite{yang2022st3d++} & $55.76$ / $43.51$ & $59.93$ / $55.28$ & $45.56$ / $36.97$ & $47.28$ / $45.84$ & $60.91$ / $40.09$ & $68.96$ / $59.96$ & $57.02$ / $37.52$ & $61.30$ / $55.43$ & $54.81$ / $39.52$ & $59.37$ / $54.13$
    \\
    & ST3D++$^{\ddagger}$ \cite{yang2022st3d++} & $54.96$ / $40.81$ & $60.47$ / $54.65$ & $45.69$ / $36.76$ & $48.30$ / $46.05$ & \textcolor{pi3det_red}{\hlred{$65.50$}} / $43.46$ & $68.99$ / $64.62$ & $55.92$ / $39.46$ & $59.93$ / $55.19$ & $55.52$ / $40.12$ & $59.42$ /
    $55.13$
    \\
    & REDB \cite{chen2023revisiting} & $52.43$ / $41.34$ & $57.12$ / $54.18$ & - / - & - / - & $65.31$ / $39.19$ & $68.74$ / $64.13$ & - / - & - / - & - / - & - / -
    \\
    & MS3D++ \cite{tsai2024ms3d++} & \textcolor{pi3det_blue}{\hlblue{$56.24$}} / $43.20$ & \textcolor{pi3det_blue}{\hlblue{$60.88$}} / \textcolor{pi3det_blue}{\hlblue{$56.13$}} & \textcolor{pi3det_blue}{\hlblue{$51.50$}} / \textcolor{pi3det_blue}{\hlblue{$40.14$}} & \textcolor{pi3det_blue}{\hlblue{$56.03$}} / \textcolor{pi3det_blue}{\hlblue{$53.86$}} & \textcolor{pi3det_blue}{\hlblue{$66.99$}} / \textcolor{pi3det_blue}{\hlblue{$43.76$}} & \textcolor{pi3det_red}{\hlred{$69.87$}} / \textcolor{pi3det_blue}{\hlblue{$65.85$}} & \textcolor{pi3det_blue}{\hlblue{$62.68$}} / \textcolor{pi3det_blue}{\hlblue{$38.26$}} & \textcolor{pi3det_blue}{\hlblue{$68.34$}} / \textcolor{pi3det_blue}{\hlblue{$61.09$}} & \textcolor{pi3det_blue}{\hlblue{$59.35$}} / \textcolor{pi3det_blue}{\hlblue{$41.34$}} & \textcolor{pi3det_blue}{\hlblue{$63.78$}} / \textcolor{pi3det_blue}{\hlblue{$59.23$}}
    \\
    & \textbf{Pi3DET-Net} & \textcolor{pi3det_red}{\hlred{$56.80$}} / \textcolor{pi3det_red}{\hlred{$46.36$}} & \textcolor{pi3det_red}{\hlred{$61.54$}} / \textcolor{pi3det_red}{\hlred{$57.20$}} & \textcolor{pi3det_red}{\hlred{$54.85$}} / \textcolor{pi3det_red}{\hlred{$42.38$}} & \textcolor{pi3det_red}{\hlred{$57.41$}} / \textcolor{pi3det_red}{\hlred{$55.54$}} & $65.43$ / \textcolor{pi3det_red}{\hlred{$45.94$}} & \textcolor{pi3det_blue}{\hlblue{$69.24$}} / \textcolor{pi3det_red}{\hlred{$65.87$}} & \textcolor{pi3det_red}{\hlred{$65.63$}} / \textcolor{pi3det_red}{\hlred{$44.62$}} & \textcolor{pi3det_red}{\hlred{$72.05$}} / \textcolor{pi3det_red}{\hlred{$63.83$}} & \textcolor{pi3det_red}{\hlred{$60.68$}} / \textcolor{pi3det_red}{\hlred{$44.83$}} & \textcolor{pi3det_red}{\hlred{$65.06$}} / \textcolor{pi3det_red}{\hlred{$60.61$}}
    \\\cmidrule{2-12}
    & Target Platform & $54.15$ / $40.24$ & $58.63$ / $54.96$ & $54.90$ / $39.74$ & $56.46$ / $55.19$ & $67.67$ / $46.11$ & $70.04$ / $66.14$ & $68.52$ / $46.53$ & $70.67$ / $61.42$ & $61.31$ / $43.16$ & $63.95$ / $59.43$
    \\
    \midrule
    \multirow{10}{*}{\rotatebox{90}{\textbf{Pi3DET (Vehicle)}}} & \cellcolor{gray!5}\textcolor{gray}{Source Platform} & \cellcolor{gray!5}\textcolor{gray}{$38.61$ / $26.84$} & \cellcolor{gray!5}\textcolor{gray}{$40.64$ / $39.22$} & \cellcolor{gray!5}\textcolor{gray}{$43.95$ / $31.24$} & \cellcolor{gray!5}\textcolor{gray}{$48.22$ / $44.17$} & \cellcolor{gray!5}\textcolor{gray}{$57.29$ / $36.62$} & \cellcolor{gray!5}\textcolor{gray}{$58.92$ / $56.19$} & \cellcolor{gray!5}\textcolor{gray}{$52.85$ / $37.96$} & \cellcolor{gray!5}\textcolor{gray}{$61.10$ / $52.47$} & $48.17$ / $33.16$ & $52.22$ / $48.01$
    \\\cmidrule{2-12}
    & ST3D \cite{yang2021st3d} & $49.29$ / $38.69$ & $51.02$ / $49.71$ & $47.70$ / $37.91$ & $48.07$ / $47.59$ & $60.17$ / $33.01$ & $62.84$ / $54.51$ & $53.79$ / $40.18$ & \textcolor{pi3det_blue}{\hlblue{$65.29$}} / $53.40$ & $52.74$ / $37.45$ & $56.81$ / $51.30$
    \\
    & ST3D$^{\ddagger}$ \cite{yang2021st3d} & $47.89$ / $38.07$ & $49.50$ / $48.23$ & $47.01$ / $41.85$ & $54.01$ / $53.46$ & $60.67$ / $33.27$ & $62.98$ / $54.61$ & $53.85$ / $40.02$ & $62.70$ / $53.08$ & $52.35$ / $38.30$ & $57.30$ / $52.34$
    \\
    & ST3D++ \cite{yang2022st3d++} & $46.05$ / $37.22$ & $49.33$ / $47.84$ & $48.52$ / $37.84$ & \textcolor{pi3det_blue}{\hlblue{$55.82$}} / $48.53$ & $60.04$ / $33.98$ & $62.71$ / $54.13$ & $53.71$ / $39.94$ & $62.43$ / $53.20$ & $52.08$ / $37.24$&$57.57$ / $50.92$
    \\
    & ST3D++$^{\ddagger}$ \cite{yang2022st3d++} & $45.14$ / $35.70$ & $46.94$ / $45.37$ & $47.52$ / $37.13$ & $54.37$ / $47.63$ & $64.15$ / $34.20$ & $63.81$ / $55.44$ & $53.64$ / $40.27$ & $62.43$ / $53.10$ &$52.61$ / $36.83$&$56.89$ / $50.38$
    \\
    & REDB \cite{chen2023revisiting} & $46.74$ / $38.47$ & $50.29$ / $49.54$ & - / - & - / - & $61.57$ / $34.05$ & $63.22$ / $54.07$ & - / - & - / -  &  - / - & - / -  
    \\
    & MS3D++ \cite{tsai2024ms3d++} & \textcolor{pi3det_blue}{\hlblue{$53.66$}} / \textcolor{pi3det_blue}{\hlblue{$40.66$}} & \textcolor{pi3det_blue}{\hlblue{$55.21$}} / \textcolor{pi3det_blue}{\hlblue{$53.78$}} & \textcolor{pi3det_blue}{\hlblue{$53.65$}} / \textcolor{pi3det_blue}{\hlblue{$41.93$}} & $54.69$ / \textcolor{pi3det_blue}{\hlblue{$54.00$}} & \textcolor{pi3det_blue}{\hlblue{$66.05$}} / \textcolor{pi3det_blue}{\hlblue{$41.17$}} & \textcolor{pi3det_blue}{\hlblue{$67.80$}} / \textcolor{pi3det_blue}{\hlblue{$63.26$}} & \textcolor{pi3det_blue}{\hlblue{$53.85$}} / \textcolor{pi3det_blue}{\hlblue{$40.91$}} & $62.87$ / \textcolor{pi3det_blue}{\hlblue{$53.44$}} & \textcolor{pi3det_blue}{\hlblue{$56.80$}} / \textcolor{pi3det_blue}{\hlblue{$41.17$}}&\textcolor{pi3det_blue}{\hlblue{$60.14$}} / \textcolor{pi3det_blue}{\hlblue{$56.12$}}
    \\
    & \textbf{Pi3DET-Net} & \textcolor{pi3det_red}{\hlred{$56.19$}} / \textcolor{pi3det_red}{\hlred{$44.28$}} & \textcolor{pi3det_red}{\hlred{$60.35$}} / \textcolor{pi3det_red}{\hlred{$56.20$}} & \textcolor{pi3det_red}{\hlred{$55.54$}} / \textcolor{pi3det_red}{\hlred{$45.18$}} & \textcolor{pi3det_red}{\hlred{$59.48$}} / \textcolor{pi3det_red}{\hlred{$58.90$}} & \textcolor{pi3det_red}{\hlred{$66.26$}} / \textcolor{pi3det_red}{\hlred{$44.47$}} & \textcolor{pi3det_red}{\hlred{$68.25$}} / \textcolor{pi3det_red}{\hlred{$63.36$}} & \textcolor{pi3det_red}{\hlred{$67.87$}} / \textcolor{pi3det_red}{\hlred{$46.83$}} & \textcolor{pi3det_red}{\hlred{$69.95$}} / \textcolor{pi3det_red}{\hlred{$66.26$}} &\textcolor{pi3det_red}{\hlred{$61.47$}} / \textcolor{pi3det_red}{\hlred{$45.19$}}& \textcolor{pi3det_red}{\hlred{$64.51$}} / \textcolor{pi3det_red}{\hlred{$61.18$}}
    \\\cmidrule{2-12}
    & Target Platform & $54.15$ / $40.24$ & $58.63$ / $54.96$ & $54.90$ / $39.74$ & $56.46$ / $55.19$ & $67.67$ / $46.11$ & $70.04$ / $66.14$ & $68.52$ / $46.53$ & $70.67$ / $61.42$ & $61.31$ / $43.16$&$63.95$ / $59.43$
    \\\midrule\midrule
    - & Combined All & ${58.21}$ / ${46.27}$ & ${62.18}$ / ${59.67}$ & ${60.96}$ / ${48.15}$ & ${63.04}$ / ${61.04}$ & ${68.44}$ / ${48.19}$ & ${71.11}$ / ${68.24}$ & ${68.90}$ / ${48.88}$ & ${72.55}$ / ${69.18}$ & $64.13$ / $47.87$ & $67.22$ / $64.53$
    \\
    \bottomrule
\end{tabular}}
\label{tab:benchmark}
\vspace{-0.29cm}
\end{table*}

\section{Experiments}
\label{sec:experiments}

\subsection{Experimental Settings}
\noindent\textbf{Datasets.} 
We evaluate cross-platform and cross-dataset 3D detection using three benchmarks: nuScenes \cite{caesar2020nuscenes}, KITTI \cite{geiger2012we}, and our Pi3DET. nuScenes \cite{caesar2020nuscenes} provides $35{,}149$ frames from day and night urban scenes, KITTI \cite{geiger2012we} provides $14{,}999$ daytime frames, and Pi3DET comprises $51{,}545$ frames spanning urban, suburban, and rural environments. For additional dataset details, please refer to Appendix~\ref{sec_supp:dataset_statistics}.

\noindent\textbf{Benchmark Setup.} 
We design six cross-platform adaptation benchmarks and two cross-dataset adaptation benchmarks to cover a wide range of scenarios and to demonstrate the generalizability of our method. Due to space limits, please refer to Appendix~\ref{sec:supp_b.1} for the complete benchmark settings.

\noindent\textbf{Baselines.}
We use PV-RCNN~\cite{shi2020pv} and Voxel-RCNN~\cite{deng2021voxel} as our detection backbones. Our comparisons include several related cross-domain detection methods ST3D \cite{yang2021st3d}, ST3D++ \cite{yang2021st3d}, and MS3D++ \cite{tsai2024ms3d++}, as well as three baseline training strategies: training on \textit{``source data only''}, training on \textit{``target data only''}, and training on \textit{``both source and target data''}. For more details, please refer to Appendix~\ref{sec:supp_b.5}.

\noindent\textbf{Implementation Details.}
Our experiments follow the setting of ST3D++ \cite{yang2022st3d++}, and are implemented using OpenPCDet \cite{openpcdet2020}, with experiments run on two NVIDIA Titan RTX GPUs. We follow the KITTI evaluation protocol by reporting average precision (AP) in both bird’s-eye view (BEV) and 3D over $40$ recall positions. The hyperparameters are set as \(\lambda_\mathrm{rot}=0.1\), \(\lambda_\mathrm{RoI}=0.2\), and \(\lambda_\mathrm{KL}=10^{-4}\). For more details, please refer to Appendix~\ref{subsec: training_configurations}.

\subsection{Comparative Study}
We analyze the performance of Pi3DET-Net across various cross-platform and cross-dataset adaptation tasks.

\noindent\textbf{Adaptation with Vehicle as Source.}
Table~\ref{tab:benchmark} presents the cross-platform adaptation results for vehicle $\rightarrow$ quadruped/drone tasks. In these experiments, source data are taken from nuScenes \cite{caesar2020nuscenes} and Pi3DET, while all target data come from Pi3DET. Overall, Pi3DET-Net consistently outperforms the baselines. For instance, on the vehicle $\rightarrow$ quadruped task using nuScenes as source, our method with PV-RCNN achieves a $12.81\%$ gain in \(\text{AP}_{\text{3D}}@0.7\) compared to the source-only baseline, validating the effectiveness of our approach. Notably, our method even outperforms target-only training, likely due to the smaller target dataset size.

\begin{wraptable}{r}{0.59\textwidth}
\centering
    \centering
    \caption{
    \textbf{Study on cross-platform 3D detection between drone and quadruped platforms}. We report the average precision (AP) in ``BEV / 3D'' at the IoU thresholds of $0.7$ and $0.5$, respectively.
    }
\vspace{-0.1cm}
\resizebox{\linewidth}{!}{
\begin{tabular}{c|l|cc|cc} 
    \toprule
    \multirow{2}{*}{\textbf{\#}} & \multirow{2}{*}{\textbf{Method}} & \multicolumn{2}{c|}{\textbf{PV-RCNN} \cite{shishaoshuai2020pv}} & \multicolumn{2}{c}{\textbf{Voxel RCNN} \cite{deng2021voxel}}
    \\
    & & \textbf{AP@0.7} & \textbf{AP@0.5} & \textbf{AP@0.7} & \textbf{AP@0.5}
    \\ 
    \midrule\midrule
    \multirow{8}{*}{\rotatebox{90}{\textbf{Quad $\rightarrow$ Drone}}} & \textcolor{gray}{Source Platform} & \textcolor{gray}{$27.43$ / $11.08$} & \textcolor{gray}{$36.97$ / $27.92$} & \textcolor{gray}{$33.22$ / $20.20$} & \textcolor{gray}{$41.17$ / $33.29$}
    \\\cmidrule{2-6}
    & ST3D$^{\ddagger}$ \cite{yang2021st3d} & $33.85$ / $18.45$ & $44.35$ / $35.83$ & $35.21$ / $22.87$ & $36.05$ / $35.52$
    \\
    & ST3D++$^{\ddagger}$ \cite{yang2022st3d++} & $32.92$ / $17.76$ & $40.91$ / $32.97$ & $43.30$ / $28.86$ & $44.69$ / $43.24$
    \\
    & REDB \cite{hu2023density} & $37.24$ / $20.89$ & $44.43$ / $37.29$ & $44.27$ / $30.55$ & $46.69$ / $44.29$
    \\
    & MS3D++ \cite{tsai2024ms3d++} & \textcolor{pi3det_blue}{\hlblue{$39.74$}} / \textcolor{pi3det_blue}{\hlblue{$22.31$}} & \textcolor{pi3det_blue}{\hlblue{$47.59$}} / \textcolor{pi3det_blue}{\hlblue{$41.61$}} & \textcolor{pi3det_blue}{\hlblue{$45.84$}} / \textcolor{pi3det_blue}{\hlblue{$32.21$}} & \textcolor{pi3det_blue}{\hlblue{$48.27$}} / \textcolor{pi3det_blue}{\hlblue{$45.87$}}
    \\
    & \textbf{Pi3DET-Net} & \textcolor{pi3det_red}{\hlred{$43.11$}} / \textcolor{pi3det_red}{\hlred{$25.16$}} & \textcolor{pi3det_red}{\hlred{$52.87$}} / \textcolor{pi3det_red}{\hlred{$47.55$}} & \textcolor{pi3det_red}{\hlred{$49.27$}} / \textcolor{pi3det_red}{\hlred{$36.24$}} & \textcolor{pi3det_red}{\hlred{$54.58$}} / \textcolor{pi3det_red}{\hlred{$49.63$}}
    \\\cmidrule{2-6}
    & Target Platform & $67.67$ / $46.11$ & $70.04$ / $66.14$ & $68.52$ / $46.53$ & $70.67$ / $61.42$  
    \\ 
    \midrule 
    \multirow{8}{*}{\rotatebox{90}{\textbf{Drone $\rightarrow$ Quad}}} & \textcolor{gray}{Source Platform} &  \textcolor{gray}{$27.23$ / $20.36$} &  \textcolor{gray}{$30.27$ / $28.92$} & \textcolor{gray}{$32.18$ / $23.35$} & \textcolor{gray}{$33.94$ / $32.70$}
    \\\cmidrule{2-6}
    & ST3D$^{\ddagger}$ \cite{yang2021st3d} & $46.06$ / $35.14$ & $51.17$ / $49.53$ & $49.04$ / $36.94$ & $55.73$ / $49.73$
    \\
    & ST3D++$^{\ddagger}$ \cite{yang2022st3d++} & \textcolor{pi3det_blue}{\hlblue{$49.09$}} / \textcolor{pi3det_blue}{\hlblue{$37.57$}} & \textcolor{pi3det_blue}{\hlblue{$55.30$}} / \textcolor{pi3det_blue}{\hlblue{$50.90$}} & $48.74$ / \textcolor{pi3det_blue}{\hlblue{$38.22$}} & $55.19$ / $48.94$
    \\
    & REDB \cite{hu2023density} & $47.29$ / $35.67$ & $53.21$ / $49.76$ & \textcolor{pi3det_blue}{\hlblue{$49.36$}} / $38.11$ & $55.96$ / \textcolor{pi3det_blue}{\hlblue{$50.21$}}
    \\
    & MS3D++ \cite{tsai2024ms3d++} & $48.24$ / $34.12$ & $52.43$ / $48.66$ & $49.76$ / $37.55$ & \textcolor{pi3det_blue}{\hlblue{$56.17$}} / $49.97$
    \\
    & \textbf{Pi3DET-Net} & \textcolor{pi3det_red}{\hlred{$51.24$}} / \textcolor{pi3det_red}{\hlred{$38.94$}} & \textcolor{pi3det_red}{\hlred{$57.31$}} / \textcolor{pi3det_red}{\hlred{$52.90$}} & \textcolor{pi3det_red}{\hlred{$52.64$}} / \textcolor{pi3det_red}{\hlred{$38.88$}} & \textcolor{pi3det_red}{\hlred{$57.57$}} / \textcolor{pi3det_red}{\hlred{$51.83$}}
    \\\cmidrule{2-6}
    & Target Platform & $54.15$ / $40.24$ & $58.63$ / $54.96$ & $54.90$ / $39.74$ & $56.46$ / $55.19$  
    \\
    \bottomrule
\end{tabular}}
\label{tab:cross_quadruped_drone}
\vspace{-0.1cm}
\end{wraptable}
\noindent\textbf{Adaptation with Drone and Quadruped as Source.}
Table~\ref{tab:cross_quadruped_drone} presents cross-platform detection results between the quadruped and drone platforms. Under our approach, both PV-RCNN and Voxel-RCNN achieve the best performance across all evaluated metrics. For instance, in the drone $\rightarrow$ quadruped task, our method with PV-RCNN improves $\text{AP}_{\text{3D}}@0.7$ by $18.58\%$ relative to the source-only baseline, nearly matching the target-only performance.

\noindent\textbf{Cross-Dataset Adaptation.}  
To demonstrate the broad applicability of Pi3DET-Net, we evaluate on the cross-dataset task from nuScenes to KITTI. Following \cite{yang2022st3d++}, we adopt SECOND-IoU~\cite{yan2018second} as the backbone. Table~\ref{tab:nusc_kitti} presents the results, which show that Pi3DET-Net achieves state-of-the-art performance on both {Car} and {Cyclist}. For {Car} targets, our \(\text{AP}_{\text{3D}}@0.7\) is only $3.25\%$ lower than that of the target-only baseline. Additionally, we design a separate cross-dataset adaptation task from nuScenes to Pi3DET on the vehicle platform, detailed analysis is provided in Appendix~\ref{sec:supp_c.3}.

\subsection{Ablation Study}
In this section, we use Voxel-RCNN \cite{deng2021voxel} as the backbone detector to validate the effectiveness of individual components in Pi3DET-Net for cross-platform tasks.

\begin{wraptable}{r}{0.59\textwidth}
\centering
\vspace{-0.4cm}
\centering
\caption{
    \textbf{Cross-dataset 3D detection benchmark}. Experiments are conducted on the nuScenes \cite{caesar2020nuscenes} $\rightarrow$ KITTI \cite{geiger2012we} task. We report AP in ``BEV / 3D'' at the IoU thresholds of $0.7$, $0.5$, and $0.5$ for Car, Pedestrian, and Cyclist classes, respectively. The scores are for moderate cases. Symbol $\dagger$ denotes method \textit{w.o.} RPJ, since no pitch or roll jitter occurs when both the source and target platforms are vehicles. \(w.\text{temp}\) indicates the use of temporal information, and \(w.\text{SN}\) denotes the incorporation of statistic normalization \cite{wang2020train}.
}
\vspace{-0.1cm}
\resizebox{\linewidth}{!}{
\begin{tabular}{l|c|c|c|c} 
    \toprule
    \multirow{2}{*}{\textbf{Method}} & \textbf{Car} & \textbf{Pedestrian} & \textbf{Cyclist} & \multirow{2}{*}{\textbf{Average}}
    \\
    & \textbf{AP@0.7} & \textbf{AP@0.5} & \textbf{AP@0.5} & 
    \\ 
    \midrule\midrule
    \textcolor{gray}{Source Dataset} & \textcolor{gray}{$51.80$ / $17.90$} & \textcolor{gray}{$39.95$ / $34.57$} & \textcolor{gray}{$17.70$ / $11.08$} & \textcolor{gray}{$36.48$ / $21.18$}
    \\\midrule
    SN \cite{wang2020train} & $40.30$ / $21.23$ & $38.91$ / $34.36$ & $11.11$ / $5.67$~~ & $30.17$ / $20.42$
    \\
    ST3D \cite{yang2021st3d} & $75.90$ / $54.10$ & $44.00$ / $42.60$ & $29.58$ / $21.21$ & $49.83$ / $39.30$
    \\
    ST3D \cite{yang2021st3d} \(w.\text{SN}\) & $79.02$ / $62.55$ & $43.12$ / $40.54$ & $16.60$ / $11.33$ & $46.25$ / $38.14$
    \\
    ST3D \cite{yang2021st3d} \(w.\text{temp}\) & $81.06$ / $66.98$ & $34.65$ / $31.76$ & $27.32$ / $20.52$ & $47.68$ / $39.75$
    \\
    ST3D++ \cite{yang2022st3d++} & $80.50$ / $62.40$ & \textcolor{pi3det_blue}{\hlblue{$47.20$}} / \textcolor{pi3det_blue}{\hlblue{$43.96$}} & \textcolor{pi3det_blue}{\hlblue{$30.87$}} / $23.93$ & \textcolor{pi3det_blue}{\hlblue{$52.86$}} / \textcolor{pi3det_blue}{\hlblue{$43.43$}}
    \\
    ST3D++ \cite{yang2022st3d++} \(w.\text{SN}\) & $78.87$ / $65.56$ & \textcolor{pi3det_red}{\hlred{$47.94$}} / \textcolor{pi3det_red}{\hlred{$45.57$}} & $13.57$ / $12.64$ & $46.79$ / $41.26$
    \\
    ST3D++ \cite{yang2022st3d++} \(w.\text{temp}\) & $80.91$ / $68.23$ & $30.48$ / $27.86$ & $29.88$ / \textcolor{pi3det_blue}{\hlblue{$25.57$}} & $47.09$ / $40.55$
    \\
    REDB \cite{chen2023revisiting} & $74.23$ / $51.31$ & $25.95$ / $18.38$ & $13.82$ / $8.64$~~ & $38.00$ / $26.11$
    \\
    DTS \cite{hu2023density} & $81.40$ / $66.60$ & - / - & - / - & - / -
    \\
    CMDA \cite{chang2024cmda} & \textcolor{pi3det_blue}{\hlblue{$82.13$}} / \textcolor{pi3det_blue}{\hlblue{$68.95$}} & - / - & - / - & - / -
    \\
    PLR \cite{zhang2024pseudo} & $73.65$ / $66.84$ & $42.69$ / $35.47$ & $17.38$ / $15.95$ & $44.57$ / $39.42$
    \\
    \textbf{Pi3DET-Net$^{\dagger}$} & \textcolor{pi3det_red}{\hlred{$82.86$}} / \textcolor{pi3det_red}{\hlred{$70.20$}} & $46.23$ / $43.44$ & \textcolor{pi3det_red}{\hlred{$31.14$}} / \textcolor{pi3det_red}{\hlred{$25.72$}} & \textcolor{pi3det_red}{\hlred{$57.51$}} / \textcolor{pi3det_red}{\hlred{$46.45$}}
    \\\midrule
    Target Dataset & $83.29$ / $73.45$ & $46.64$ / $41.33$ & $62.92$ / $60.32$ & $62.92$ / $60.32$ 
    \\
    \bottomrule
\end{tabular}}
\label{tab:nusc_kitti}
\vspace{-0.3cm}
\end{wraptable}
\noindent\textbf{Random Platform Jitter.} 
As shown in Table~\ref{tab:ablation_study}, adding RPJ leads to performance improvements across all metrics. For instance, in the vehicle \(\rightarrow\) drone task, the addition of RPJ boosts \(\text{AP}_{\text{BEV}}@0.7\) by $7.35\%$ relative to the source-only baseline. These results confirm that simulating ego-motion noise through RPJ effectively augments the source data, thereby enhancing the model’s robustness to the jitters observed on non-vehicle platforms.

\noindent\textbf{Virtual Platform Pose.}  
We also evaluate the impact of Virtual Platform Pose (VPP) in Table~\ref{tab:ablation_study}. The results clearly show that VPP enhances Pi3DET-Net's performance, achieving a $7\%$ improvement in \(\text{AP}_{\text{3D}}@0.5\) relative to the source-only baseline in the Vehicle \(\rightarrow\) Drone task. Notably, when RPJ and VP are combined, they yield greater improvements, see an enhancement of $9.67\%$ in \(\text{AP}_{\text{BEV}}@0.7\). These findings underscore the importance of both geometric alignment strategies in improving cross-platform detection performance.

\noindent\textbf{KL Probabilistic Feature Alignment.}  
PFA is designed to narrow the cross-platform gap during the Knowledge-Adaption stage. As shown in Table~\ref{tab:ablation_study}, incorporating PFA leads to significant performance gains on cross-platform tasks. By approximating the RoI features with probabilistic encoders and aligning their distributions using a KL divergence loss, PFA ensures that the target features are gradually pulled toward the source feature manifold. This alignment is crucial for reducing domain discrepancies and improving the overall detection accuracy on the target platform.

\begin{wraptable}{r}{0.65\textwidth}
\centering
\vspace{-0.4cm}
\centering
\caption{
    \textbf{Ablation study of components in Pi3DET-Net}. Experiments are conducted on the vehicle $\rightarrow$ drone/quadruped tasks, respectively. We report the average precision (AP) in ``BEV / 3D'' at the IoU thresholds of $0.7$ and $0.5$, respectively. All scores are given in \%.}
\vspace{-0.1cm}
\resizebox{\linewidth}{!}{
\begin{tabular}{cccc|cc|cc} 
    \toprule
    \multirow{2}{*}{\textbf{RPJ}} & \multirow{2}{*}{\textbf{VPP}} & \multirow{2}{*}{\textbf{PFA}} & \multirow{2}{*}{\textbf{GTD}} & \multicolumn{2}{c|}{\textbf{Vehicle $\rightarrow$ Drone}} & \multicolumn{2}{c}{\textbf{Vehicle $\rightarrow$ Quadruped}}  
    \\
    & & & & \textbf{AP@0.7} & \textbf{AP@0.5} & \textbf{AP@0.7} & \textbf{AP@0.5}
    \\ 
    \midrule\midrule
    \textcolor{pi3det_red}{\xmark} & \textcolor{pi3det_red}{\xmark} & \textcolor{pi3det_red}{\xmark} & \textcolor{pi3det_red}{\xmark} & $52.85$ / $37.96$ & $61.10$ / $52.47$ & $43.95$ / $31.24$ & $48.22$ / $44.17$     
    \\\midrule
    \textcolor{pi3det_blue}{\cmark} & \textcolor{pi3det_red}{\xmark} & \textcolor{pi3det_red}{\xmark} & \textcolor{pi3det_red}{\xmark} & $60.20$ / $39.93$ & $64.76$ / $59.52$ & $45.36$ / $33.01$ & $49.26$ / $47.03$     
    \\
    \textcolor{pi3det_red}{\xmark} & \textcolor{pi3det_blue}{\cmark} & \textcolor{pi3det_red}{\xmark}& \textcolor{pi3det_red}{\xmark} & $59.83$ / $39.26$ & $63.55$ / $59.47$ & $44.43$ / $32.23$ & $51.59$ / $49.47$     
    \\
    \textcolor{pi3det_blue}{\cmark} & \textcolor{pi3det_blue}{\cmark} & \textcolor{pi3det_red}{\xmark} & \textcolor{pi3det_red}{\xmark} & $64.52$ / $41.50$ & $66.84$ / $60.68$ & $48.45$ / $36.10$ & $53.83$ / $51.52$     
    \\ 
    \midrule
    \textcolor{pi3det_blue}{\cmark} & \textcolor{pi3det_blue}{\cmark} & \textcolor{pi3det_blue}{\cmark} & \textcolor{pi3det_red}{\xmark} & \textcolor{pi3det_blue}{\hlblue{$67.87$}} / \textcolor{pi3det_blue}{\hlblue{$46.83$}} & \textcolor{pi3det_red}{\hlred{$69.95$}} / \textcolor{pi3det_blue}{\hlblue{$66.26$}} & \textcolor{pi3det_red}{\hlred{$55.72$}} / \textcolor{pi3det_blue}{\hlblue{$44.77$}} & \textcolor{pi3det_blue}{\hlblue{$59.48$}} / \textcolor{pi3det_blue}{\hlblue{$58.90$}} 
    \\
    \textcolor{pi3det_blue}{\cmark} & \textcolor{pi3det_blue}{\cmark} & \textcolor{pi3det_blue}{\cmark} & \textcolor{pi3det_blue}{\cmark} & \textcolor{pi3det_red}{\hlred{$68.48$}} / \textcolor{pi3det_red}{\hlred{$47.75$}} & \textcolor{pi3det_blue}{\hlblue{$69.87$}} / \textcolor{pi3det_red}{\hlred{$67.82$}} & \textcolor{pi3det_blue}{\hlblue{$55.54$}} / \textcolor{pi3det_red}{\hlred{$45.18$}} & \textcolor{pi3det_red}{\hlred{$62.02$}} / \textcolor{pi3det_red}{\hlred{$60.29$}}      
    \\
    \bottomrule
\end{tabular}}
\label{tab:ablation_study}
\vspace{-0.3cm}
\end{wraptable}
\noindent\textbf{Geometry-Aware Transformation Descriptor.}  
GTD is designed to capture global transformation cues on the source platform during the PA stage and correct global offsets on the target platform during the KA stage. As demonstrated in Table~\ref{tab:ablation_study}, incorporating GTD leads to significant performance gains. By learning geometric intrinsic that reflect sensor-specific characteristics such as sensor height and pitch distribution, GTD helps the network to predict and correct spatial misalignments between platforms.

In Appendix~\ref{sec:supp_c.3}, we provide a detailed analysis of the impact of varying the jitter angles introduced by RPJ across different platforms,  where we investigate how different levels of simulated ego-motion affect detection performance.

\subsection{Multi-Platform 3D Detection Benchmark}

\begin{table}
\centering
\caption{
    \textbf{Cross-platform 3D object detection benchmark}. Experiments are conducted on the {\includegraphics[width=0.02\linewidth]{figures/icons/vehicle.png}} \textbf{Vehicle}, {\includegraphics[width=0.021\linewidth]{figures/icons/drone.png}} \textbf{Drone}, and {\includegraphics[width=0.021\linewidth]{figures/icons/quadruped.png}} \textbf{Quadruped} platforms. We report the average precision (AP) in ``BEV / 3D'' at the IoU thresholds of $0.7$. All scores are given in percentage (\%). ``-C" and ``-A" denote detectors with the Anchor-based or Center-based detection head.
}
\vspace{-0.1cm}
\resizebox{0.8\linewidth}{!}{
\begin{tabular}{c|l|c|c|c|c} 
    \toprule
    \multirow{2}{*}{\textbf{\#}} & \multirow{2}{*}{\textbf{Method}} & \textbf{Vehicle} & \textbf{Quadruped} & \textbf{Drone} & \multirow{2}{*}{\textbf{Average}}
    \\
    && \textbf{AP@0.7} & \textbf{AP@0.7} & \textbf{AP@0.7} & 
    \\ 
    \midrule\midrule

    \multirow{7}{*}{\rotatebox{90}{\textbf{Grid}}} & PointPillar \cite{lang2019pointpillars} & $51.85$ / $44.34$& $36.24$ / $14.51$& $49.53$ / $27.02$& $45.87$ / $28.62$
    \\
    % & SECOND-IOU \cite{yan2018second} & $50.99$ / $38.99$& $38.01$ / $18.11$& $56.25$ / $34.11$& $48.42$ / $30.40$
    % \\
    & CenterPoint \cite{yin2021center} & $51.90$ / $42.12$& $37.74$ / $14.68$& $53.14$ / $29.29$& $47.59$ / $28.70$
    \\
    % & PillarNet \cite{shi2022pillarnet} & $50.18$ / $38.02$& $34.14$ / $12.06$& $47.59$ / $24.00$& $43.97$ / $24.69$
    % \\
    & Part A$^*$ \cite{shi2020points} & \textcolor{pi3det_red}{\hlred{$54.88$}} / \textcolor{pi3det_red}{\hlred{$48.23$}}& \textcolor{pi3det_red}{\hlred{$45.47$}} / \textcolor{pi3det_blue}{\hlblue{$20.10$}} & \textcolor{pi3det_blue}{\hlblue{$56.72$}} / \textcolor{pi3det_blue}{\hlblue{$34.44$}} & \textcolor{pi3det_blue}{\hlblue{$52.36$}} / \textcolor{pi3det_blue}{\hlblue{$34.26$}}
    \\
    & Transfusion-L \cite{bai2022transfusion} & $49.27$ / $38.21$& $36.29$ / $14.43$& $51.27$ / $24.63$& $45.61$ / $25.76$
    \\
    & HEDNet \cite{zhang2023hednet} & $46.73$ / $37.60$& $34.30$ / $14.51$& $49.31$ / $20.89$& $43.45$ / $24.33$
    \\
    & SAFNet \cite{kong2024safnet} & $42.60$ / $34.88$& $33.47$ / $13.65$& $49.93$ / $24.70$& $42.00$ / $24.41$
    \\
    &\textbf{Part A$^*$+ Ours}& \textcolor{pi3det_blue}{\hlblue{$53.81$}} / \textcolor{pi3det_blue}{\hlblue{$47.56$}} & \textcolor{pi3det_blue}{\hlblue{$44.31$}} / \textcolor{pi3det_red}{\hlred{$23.73$}} & \textcolor{pi3det_red}{\hlred{$59.53$}} / \textcolor{pi3det_red}{\hlred{$38.31$}} & \textcolor{pi3det_red}{\hlred{$52.55$}} / \textcolor{pi3det_red}{\hlred{$36.53$}}
    \\
    \midrule
    \multirow{5}{*}{\rotatebox{90}{\textbf{Point}}} & PointRCNN \cite{shi2019pointrcnn} & \textcolor{pi3det_blue}{\hlblue{$49.38$}}  / \textcolor{pi3det_blue}{\hlblue{$43.03$}} & $41.35$ / $23.69$& $52.59$ / \textcolor{pi3det_blue}{\hlblue{$38.67$}} & $47.77$ / \textcolor{pi3det_blue}{\hlblue{$35.13$}}
    \\
    & 3DSSD \cite{yang20203dssd} & $46.58$ / $39.88$& $42.47$ / $23.89$& $51.54$ / $37.78$& $46.86$ / $33.85$
    \\
    & IA-SSD \cite{zhang2022not} & $44.00$ / $34.91$& \textcolor{pi3det_red}{\hlred{$48.11$}} / \textcolor{pi3det_blue}{\hlblue{$24.89$}} & \textcolor{pi3det_red}{\hlred{$59.69$}} / $35.79$&  \textcolor{pi3det_red}{\hlred{$50.60$}} / $31.86$
    \\
    & DBQ-SSD \cite{yang2022dbq} & $41.28$ / $33.19$ &  \textcolor{pi3det_blue}{\hlblue{$44.27$}} / $21.85$& $54.65$ / $32.08$& $46.73$ / $29.04$
    \\
    &\textbf{PointRCNN + Ours}& \textcolor{pi3det_red}{\hlred{$51.19$}} / \textcolor{pi3det_red}{\hlred{$48.09$}} & $42.18$ / \textcolor{pi3det_red}{\hlred{$26.07$}} & \textcolor{pi3det_blue}{\hlblue{$57.54$}} / \textcolor{pi3det_red}{\hlred{$41.70$}} & \textcolor{pi3det_blue}{\hlblue{$50.30$}} / \textcolor{pi3det_red}{\hlred{$38.62$}}
    \\
    \midrule
    \multirow{6}{*}{\rotatebox{90}{\textbf{Grid-Point}}} & PV-RCNN \cite{shi2020pv} & $63.32$ / $56.58$& $45.22$ / $22.94$& $60.11$ / $39.68$& $56.22$ / $39.73$
    \\
    % & PV-RCNN-C \cite{shi2020pv} & $52.18$ / $50.84$& $40.82$ / $20.69$& $52.86$ / $39.52$& $48.62$ / $37.02$
    % \\
    & PV-RCNN++ \cite{shi2023pv} & \textcolor{pi3det_red}{\hlred{$64.05$}} / \textcolor{pi3det_red}{\hlred{$57.01$}} & $47.54$ / $22.35$& $60.54$ / $40.10$& $57.38$ / $39.82$
    \\
    & PV-RCNN++-C \cite{shi2023pv} & $57.94$ / $50.56$& $40.75$ / $20.78$& $53.46$ / $40.00$& $50.72$ / $37.11$
    \\
    & VoxelRCNN-A \cite{deng2021voxel} & $63.00$ / \textcolor{pi3det_blue}{\hlblue{$56.98$}} & $46.78$ / \textcolor{pi3det_blue}{\hlblue{$23.30$}} & \textcolor{pi3det_blue}{\hlblue{$64.46$}} /\textcolor{pi3det_blue}{\hlblue{$42.76$}} & \textcolor{pi3det_blue}{\hlblue{$58.08$}} / \textcolor{pi3det_blue}{\hlblue{$41.01$}}
    \\
    & VoxelRCNN \cite{deng2021voxel} & $58.39$ / $51.11$ & \textcolor{pi3det_blue}{\hlblue{$48.30$}} / $21.61$& $60.29$ / $39.15$& $55.66$ / $37.29$
    \\
    &\textbf{PV-RCNN++ + Ours}& \textcolor{pi3det_blue}{\hlblue{$63.47$}} / $56.60$ & \textcolor{pi3det_red}{\hlred{$57.08$}} / \textcolor{pi3det_red}{\hlred{$31.09$}} & \textcolor{pi3det_red}{\hlred{$68.52$}} / \textcolor{pi3det_red}{\hlred{$47.92$}} & \textcolor{pi3det_red}{\hlred{$63.02$}} / \textcolor{pi3det_red}{\hlred{$45.20$}}
    \\
    \bottomrule
\end{tabular}}
\label{tab:det_task}
\vspace{-0.3cm}
\end{table}

We establish a benchmark on Pi3DET to evaluate the cross-platform performance of $18$ commonly-used 3D detectors by training all models on the vehicle set and testing them on vehicle, quadruped, and drone data (see Table~\ref{tab:det_task} and Appendix~\ref{sec:supp_c.2}). Detectors are categorized into grid-based, point-based, and grid-point-based. Although grid-point-based methods excel on vehicles, their performance declines on quadruped and drone platforms, where point-based detectors achieve more balanced results, demonstrating enhanced viewpoint robustness. Furthermore, we apply our RPJ to the top-performing detectors on the vehicle platform. While this augmentation slightly degrades performance on vehicles due to the introduction of unseen noises, it significantly boosts results on the other two platforms. Overall, our findings underscore that effective geometry alignment and robust point-based architectures are crucial for developing unified 3D detectors across diverse platforms.
\section{Conclusion}
\label{sec:conclusion}
In this work, we introduced \textbf{Pi3DET}, a large-scale dataset for cross-platform 3D detection that includes diverse samples from vehicle, drone, and quadruped platforms. We proposed a novel adaptation approach that transfers the knowledge of vehicle detectors to other platforms by aligning geometric and feature representations. Extensive experiments show that our method is superior in both cross-platform and cross-dataset 3D object detection. We also establish a cross-platform benchmark on current 3D detectors and provide insights to improve resilience to platform variations, which benefits the research on unified 3D detection systems operating reliably across diverse autonomous platforms.

\vspace{0.2cm}
\beginappendix

\startcontents[appendices]
\printcontents[appendices]{l}{1}{\setcounter{tocdepth}{2}}

\section{Pi3DET: Construction \& Statistics}
\label{sec_supp:dataset_statistics}

In this section, we briefly outline the overview of the proposed \textbf{Pi3DET} dataset, present detailed statistics, showcase representative examples, analyze cross-platform discrepancies, compare with existing 3D detection datasets, and describe the annotation toolkit used for precise 3D labeling.

\subsection{Overview}
\textbf{Pi3DET} is the first benchmark designed for 3D object detection across multiple robot platforms. Built upon M3ED \cite{chaney2023m3ed}, our dataset consists of $25$ sequences collected from three distinct platforms: \includegraphics[width=0.02\linewidth]{figures/icons/vehicle.png} \textbf{Vehicle}, \includegraphics[width=0.021\linewidth]{figures/icons/drone.png} \textbf{Drone}, and \includegraphics[width=0.021\linewidth]{figures/icons/quadruped.png} \textbf{Quadruped}. 

In each sequence, detailed $10$ Hz annotations are performed for vehicle and pedestrian targets, resulting in a total of $51{,}545$ annotated frames. The dataset spans a wide range of environmental conditions -- including both daytime and nighttime scenes -- and encompasses urban, suburban, and rural settings. This extensive and diverse benchmark offers a valuable resource for advancing cross-platform 3D object detection research.

\begin{table*}[t]
\centering
\caption{Summary of the platform-level and sequence-level statistics of the proposed \textbf{Pi3DET} dataset.}
\vspace{-0.1cm}
\resizebox{\linewidth}{!}{
\begin{tabular}{c|c|l|c|c|c|c}
    \toprule
    \textbf{Platform} & \textbf{~Condition~} & \textbf{Sequence} & \textbf{~\# of Frames~} & \textbf{~\# of Points (M)~} & \textbf{~\# of Vehicles~} & \textbf{~\# of Pedestrians~}
    \\\midrule\midrule
    \multirow{10}{*}{\makecell{\textcolor{pi3det_red}{\textbf{Vehicle}}\\\textcolor{pi3det_red}{$\mathbf{(8)}$}}} & \multirow{4}{*}{\makecell{Daytime\\$(4)$}} & city\_hall & $2,982$ & $26.61$ & $19,489$ & $12,199$
    \\
    & & penno\_big\_loop & $3,151$ & $33.29$ & $17,240$ & $1,886$
    \\
    & & rittenhouse & $3,899$ & $49.36$ & $11,056$ & $12,003$
    \\
    & & ucity\_small\_loop & $6,746$ & $67.49$ & $34,049$ & $34,346$
    \\\cmidrule{2-7}
    & \multirow{4}{*}{\makecell{Nighttime\\$(4)$}} & city\_hall & $2,856$ & $26.16$ & $12,655$ & $5,492$
    \\
    & & penno\_big\_loop & $3,291$ & $38.04$ & $8,068$ & $106$
    \\
    & & rittenhouse & $4,135$ & $52.68$ & $11,103$ & $14,315$
    \\
    & & ucity\_small\_loop & $5,133$ & $53.32$ & $18,251$ & $8,639$
    \\\cmidrule{2-7}
    & \multicolumn{2}{c|}{\textcolor{pi3det_red}{\textbf{Summary (Vehicle)}}} & \cellcolor{pi3det_red!12}\textcolor{pi3det_red}{$\mathbf{32{,}193}$} & \cellcolor{pi3det_red!12}\textcolor{pi3det_red}{$\mathbf{346.95}$} & \cellcolor{pi3det_red!12}\textcolor{pi3det_red}{$\mathbf{131{,}911}$} & \cellcolor{pi3det_red!12}\textcolor{pi3det_red}{$\mathbf{88{,}986}$}
    \\\midrule
    \multirow{9}{*}{\makecell{\textcolor{pi3det_blue}{\textbf{Drone}}\\\textcolor{pi3det_blue}{$\mathbf{(7)}$}}} & \multirow{4}{*}{\makecell{Daytime\\$(4)$}} & penno\_parking\_1 & $1,125$ & $8.69$ & $6,075$ & $115$
    \\
    & & penno\_parking\_2 & $1,086$ & $8.55$ & $5,896$ & $340$
    \\
    & & penno\_plaza & $678$ & $5.60$ & $721$ & $65$
    \\
    & & penno\_trees & $1,319$ & $11.58$ & $657$ & $160$
    \\\cmidrule{2-7}
    & \multirow{3}{*}{\makecell{Nighttime\\$(3)$}} & high\_beams & $674$ & $5.51$ & $578$ & $211$
    \\
    & & penno\_parking\_1 & $1,030$ & $9.42$ & $524$ & $151$
    \\
    & & penno\_parking\_2 & $1,140$ & $10.12$ & $83$ & $230$
    \\\cmidrule{2-7}
    & \multicolumn{2}{c|}{\textcolor{pi3det_blue}{\textbf{Summary (Drone)}}} & \cellcolor{pi3det_blue!12}\textcolor{pi3det_blue}{$\mathbf{7{,}052}$} & \cellcolor{pi3det_blue!12}\textcolor{pi3det_blue}{$\mathbf{59.47}$} & \cellcolor{pi3det_blue!12}\textcolor{pi3det_blue}{$\mathbf{14{,}534}$} & \cellcolor{pi3det_blue!12}\textcolor{pi3det_blue}{$\mathbf{1{,}272}$}
    \\\midrule
    \multirow{12}{*}{\makecell{\textcolor{pi3det_green}{\textbf{Quadruped}}\\\textcolor{pi3det_green}{$\mathbf{(10)}$}}} & \multirow{8}{*}{\makecell{Daytime\\$(8)$}} & art\_plaza\_loop & $1,446$ & $14.90$ & $0$ & $3,579$
    \\
    & & penno\_short\_loop & $1,176$ & $14.68$ & $3,532$ & $89$
    \\
    & & rocky\_steps & $1,535$ & $14.42$ & $0$ & $5,739$
    \\
    & & skatepark\_1 & $661$ & $12.21$ & $0$ & $893$
    \\
    & & skatepark\_2 & $921$ & $8.47$ & $0$ & $916$
    \\
    & & srt\_green\_loop & $639$ & $9.23$ & $1,349$ & $285$
    \\
    & & srt\_under\_bridge\_1 & $2,033$ & $28.95$ & $0$ & $1,432$
    \\
    & & srt\_under\_bridge\_2 & $1,813$ & $25.85$ & $0$ & $1,463$
    \\\cmidrule{2-7}
    & Nighttime & penno\_plaza\_lights & $755$ & $11.25$ & $197$ & $52$
    \\
    & $(2)$ & penno\_short\_loop & $1,321$ & $16.79$ & $904$ & $103$
    \\\cmidrule{2-7}
    & \multicolumn{2}{c|}{\textcolor{pi3det_green}{\textbf{Summary (Quadruped)}}} & \cellcolor{pi3det_green!12}\textcolor{pi3det_green}{$\mathbf{12{,}300}$} & \cellcolor{pi3det_green!12}\textcolor{pi3det_green}{$\mathbf{156.75}$} & \cellcolor{pi3det_green!12}\textcolor{pi3det_green}{$\mathbf{5{,}982}$} & \cellcolor{pi3det_green!12}\textcolor{pi3det_green}{$\mathbf{14{,}551}$}
    \\\midrule\midrule
    \textbf{All Three Platforms} & \multicolumn{2}{c|}{\multirow{2}{*}{\textbf{Summary (All)}}} & \multirow{2}{*}{$\mathbf{51{,}545}$} & \multirow{2}{*}{$\mathbf{563.17}$} & \multirow{2}{*}{$\mathbf{152{,}427}$} & \multirow{2}{*}{$\mathbf{104{,}809}$}
    \\
    $\mathbf{(25)}$ & \multicolumn{2}{c|}{} & & & 
    \\
    \bottomrule
\end{tabular}}
\label{tab:dataset_stats}
\end{table*}

\subsection{Dataset Statistics}
Table~\ref{tab:dataset_stats} summarizes the detailed statistics of the \textbf{Pi3DET} dataset. In total, our dataset comprises $25$ sequences collected from three robot platforms: Vehicle, Drone, and Quadruped. 
\begin{itemize}
    \item The \includegraphics[width=0.02\linewidth]{figures/icons/vehicle.png} \textbf{Vehicle} subset (eight sequences in total) contains $32{,}193$ frames with approximately $346.95$ million LiDAR points, along with $131,911$ vehicle and $88{,}986$ pedestrian annotations. 

    \item The \includegraphics[width=0.021\linewidth]{figures/icons/drone.png} \textbf{Drone} subset (seven sequences in total) contains $7{,}052$ frames, $59.47$ million points, $14{,}534$ vehicle annotations, and $1{,}272$ pedestrian annotations. 

    \item The \includegraphics[width=0.021\linewidth]{figures/icons/quadruped.png} \textbf{Quadruped} subset (ten sequences in total) contains $12{,}300$ frames with $156.75$ million points, $5{,}982$ vehicle annotations, and $14{,}551$ pedestrian annotations. 
\end{itemize}

Overall, \textbf{Pi3DET} consists of $\mathbf{51{,}545}$ frames and $\mathbf{563.17}$ million points, offering a diverse benchmark captured under varying conditions (daytime and nighttime) and across urban, suburban, and rural environments, thereby providing a comprehensive resource for real-world, cross-platform 3D object detection research.

For each platform in the \textbf{Pi3DET} dataset, we collect comprehensive statistics to characterize the data from multiple perspectives. Specifically, we compile point cloud distribution statistics including \(p^x\), \(p^y\), \(p^z\) coordinates and intensity values to capture spatial density and spread. In addition, we gather 3D object statistics, such as the number of objects per frame and the average number of points per bounding box, to assess detection challenges across varying environments. Finally, we documented 3D bounding box statistics, detailing dimensions such as length (\(l\)), width (\(w\)), and height (\(h\)). Details are provided in the following sections.

\subsection{Dataset Examples}
\label{sec_supp:dataset_examples}
\begin{figure*}
    \centering
    \includegraphics[width=\linewidth]{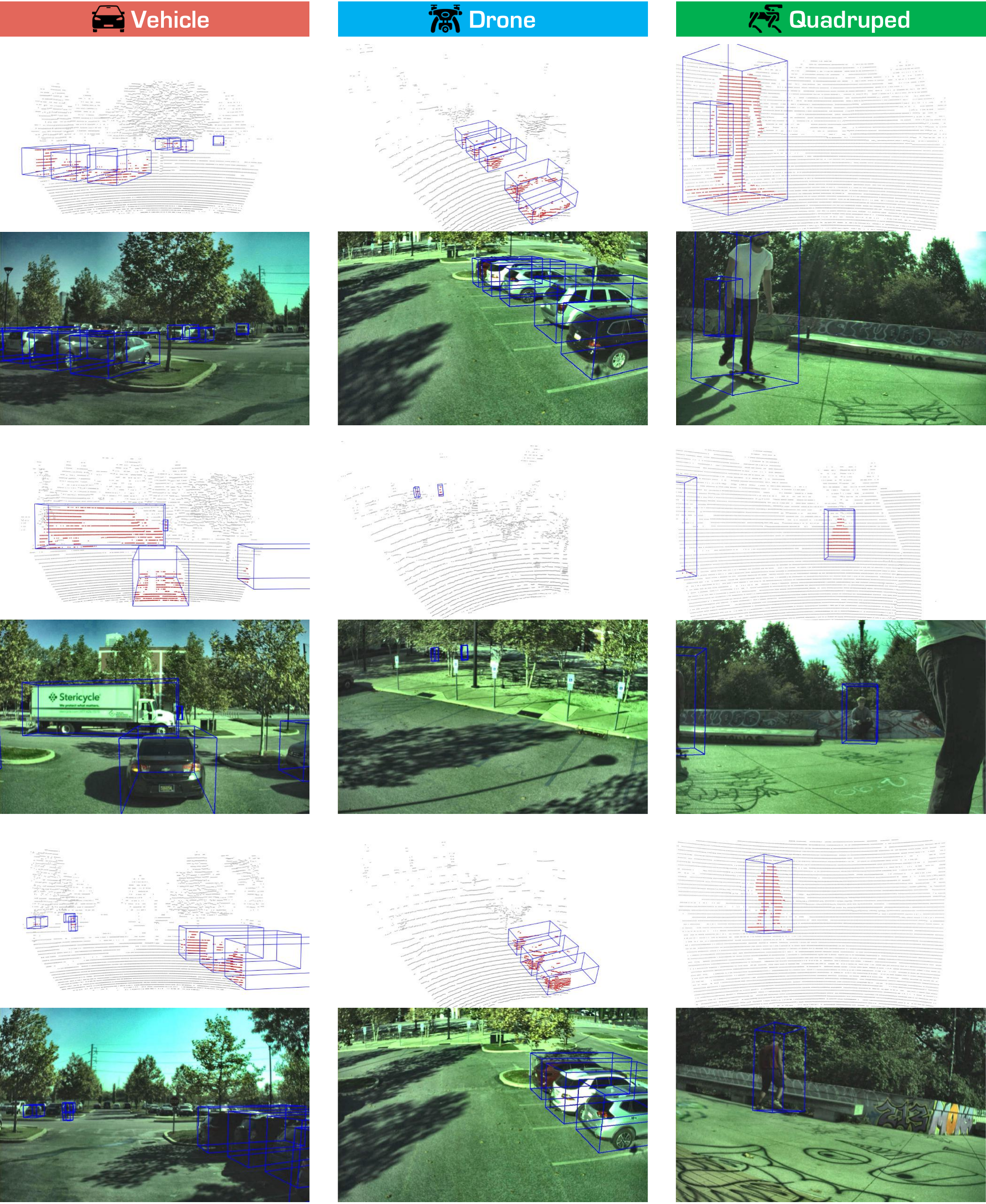}
    \vspace{-0.5cm}
    \caption{Examples of \textbf{3D object detection annotations} from 3D (LiDAR point cloud) and 2D (RGB image) in our \textbf{Pi3DET} dataset. We provide data from \textbf{three robot platforms}: \includegraphics[width=0.022\linewidth]{figures/icons/vehicle.png} \textbf{Vehicle}, \includegraphics[width=0.023\linewidth]{figures/icons/drone.png} \textbf{Drone}, and \includegraphics[width=0.023\linewidth]{figures/icons/quadruped.png} \textbf{Quadruped}. Best viewed in colors.}
    \label{fig:dataset_example_1}
\end{figure*}

\begin{figure*}
    \centering
    \includegraphics[width=\linewidth]{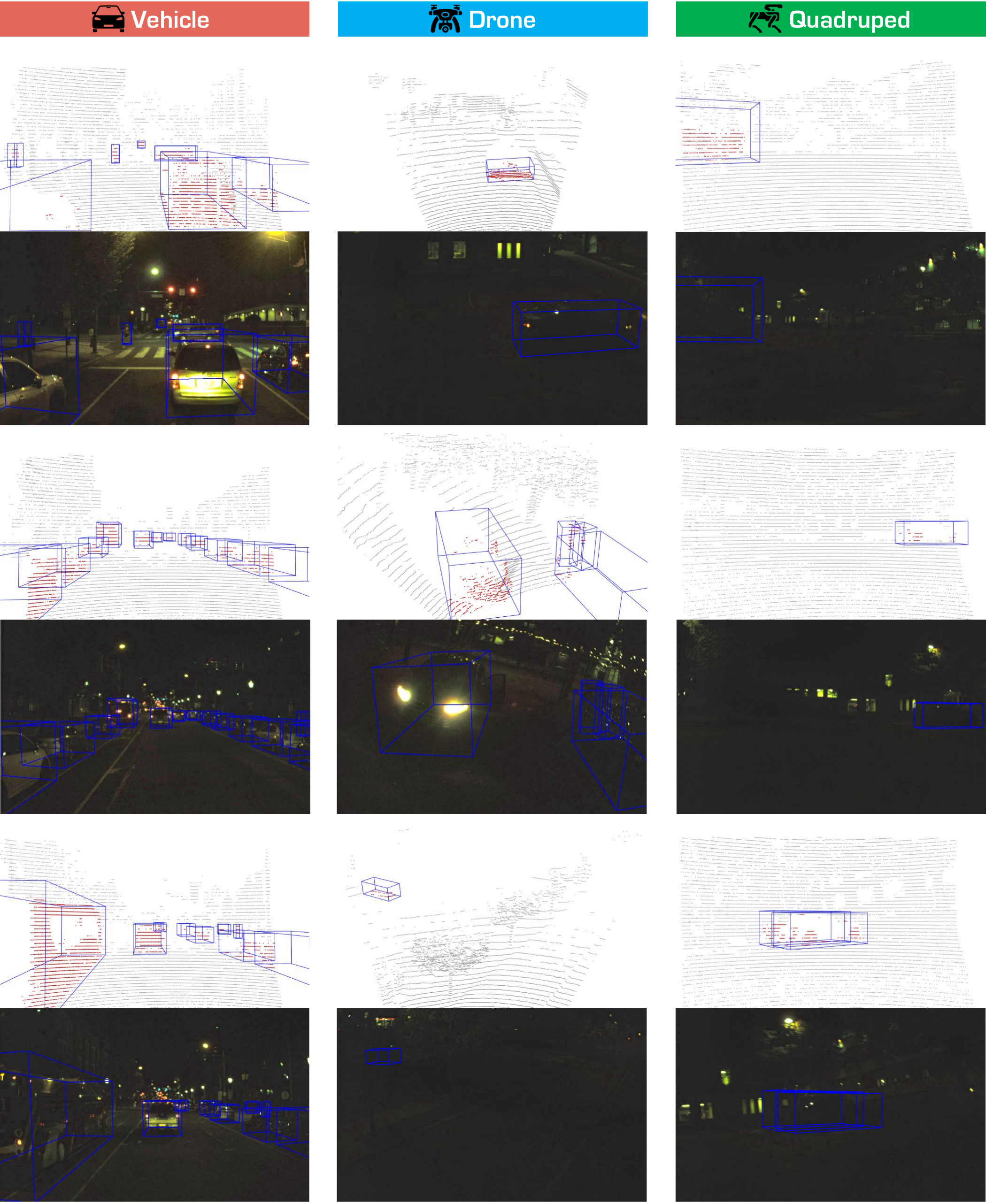}
    \vspace{-0.5cm}
    \caption{Examples of \textbf{3D object detection annotations} from 3D (LiDAR point cloud) and 2D (RGB image) in our \textbf{Pi3DET} dataset. We provide data from \textbf{three robot platforms}: \includegraphics[width=0.022\linewidth]{figures/icons/vehicle.png} \textbf{Vehicle}, \includegraphics[width=0.023\linewidth]{figures/icons/drone.png} \textbf{Drone}, and \includegraphics[width=0.023\linewidth]{figures/icons/quadruped.png} \textbf{Quadruped}. Best viewed in colors.}
    \label{fig:dataset_example_2}
\end{figure*}

\begin{figure*}
    \centering
    \includegraphics[width=\linewidth]{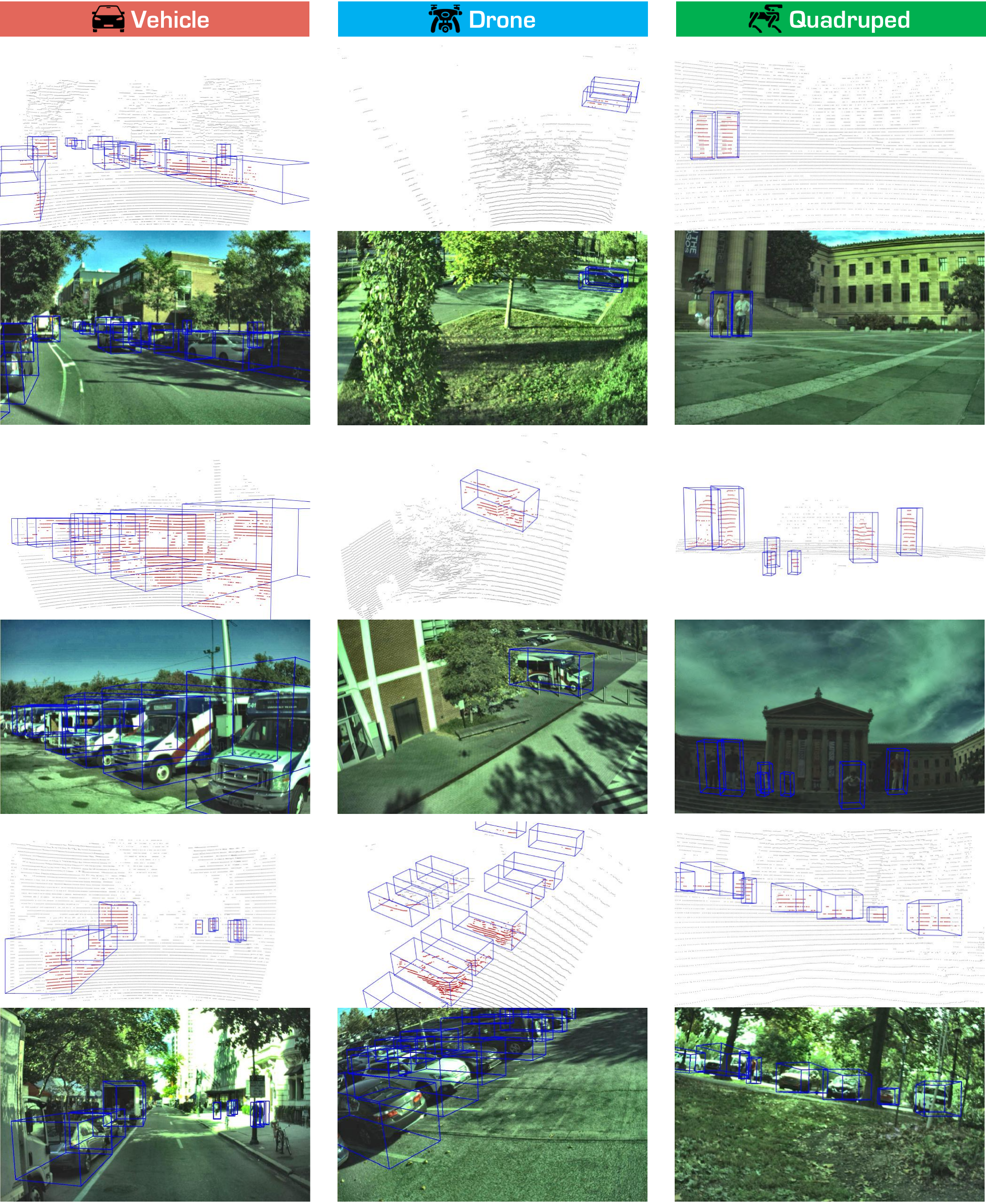}
    \vspace{-0.5cm}
    \caption{Examples of \textbf{3D object detection annotations} from 3D (LiDAR point cloud) and 2D (RGB image) in our \textbf{Pi3DET} dataset. We provide data from \textbf{three robot platforms}: \includegraphics[width=0.022\linewidth]{figures/icons/vehicle.png} \textbf{Vehicle}, \includegraphics[width=0.023\linewidth]{figures/icons/drone.png} \textbf{Drone}, and \includegraphics[width=0.023\linewidth]{figures/icons/quadruped.png} \textbf{Quadruped}. Best viewed in colors.}
    \label{fig:dataset_example_3}
\end{figure*}

\begin{figure*}
    \centering
    \includegraphics[width=\linewidth]{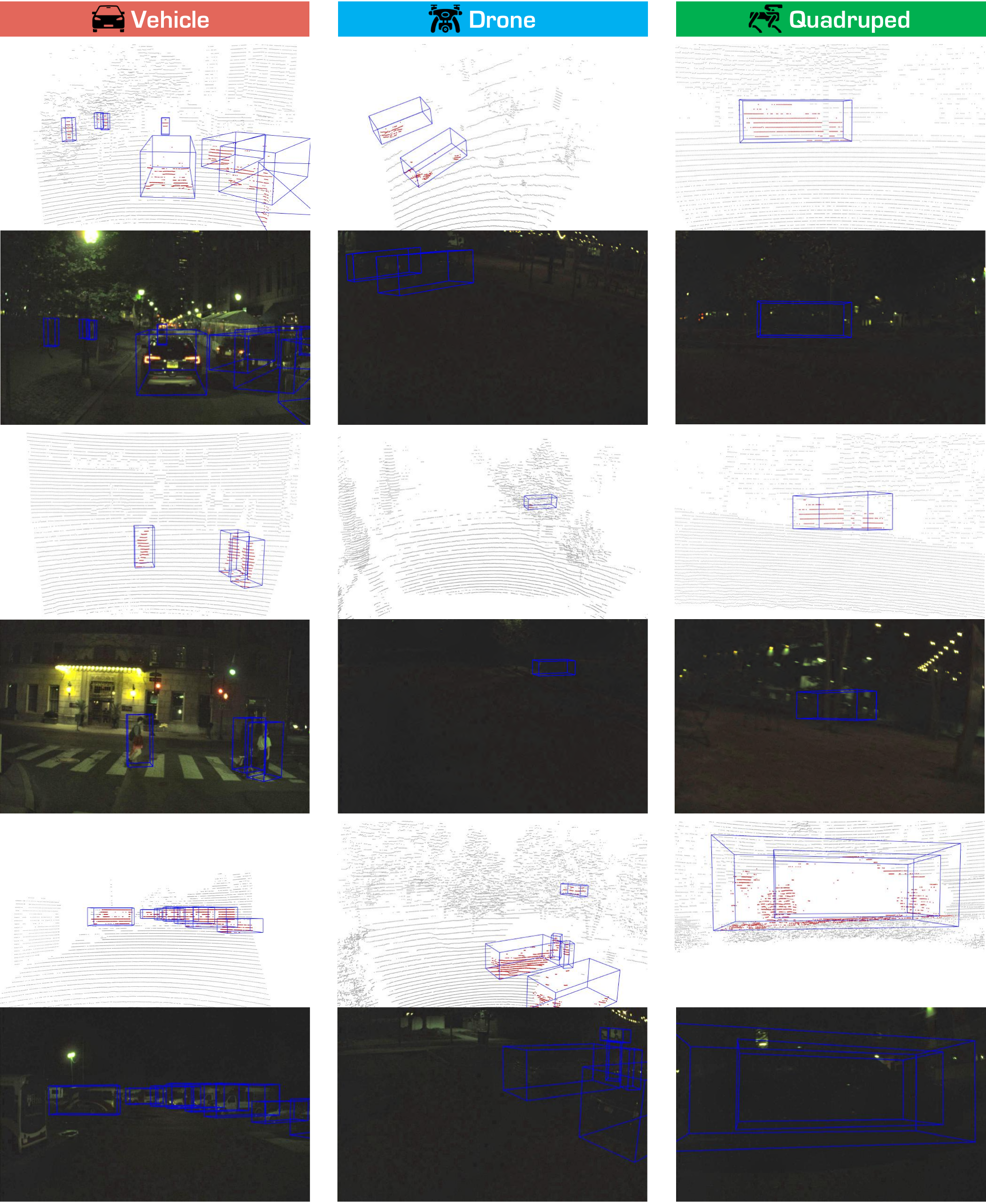}
    \vspace{-0.5cm}
    \caption{Examples of \textbf{3D object detection annotations} from 3D (LiDAR point cloud) and 2D (RGB image) in our \textbf{Pi3DET} dataset. We provide data from \textbf{three robot platforms}: \includegraphics[width=0.022\linewidth]{figures/icons/vehicle.png} \textbf{Vehicle}, \includegraphics[width=0.023\linewidth]{figures/icons/drone.png} \textbf{Drone}, and \includegraphics[width=0.023\linewidth]{figures/icons/quadruped.png} \textbf{Quadruped}. Best viewed in colors.}
    \label{fig:dataset_example_4}
\end{figure*}

In this section, we present some examples that demonstrate the rich diversity of the \textbf{Pi3DET} dataset. See Figure~\ref{fig:dataset_example_1} through Figure~\ref{fig:dataset_example_4} for details.

\textbf{Pi3DET} encompasses a wide range of scenes and temporal conditions. In particular, the quadruped platform is capable of operating in complex environments such as under bridges and on stairs, while the drone platform collects aerial views with significantly different imaging characteristics from the vehicle platform. 

Overall, the vehicle platform generally provides a slightly downward-facing view; the quadruped platform offers an upward view, yet its motion is highly dynamic and terrain-dependent, leading to a broader distribution of view angles; and the drone platform, although it typically captures targets below, exhibits considerable jitter and a wider range of view distributions due to its increased degrees of freedom.

Specifically, for the quadruped platform, Figure~\ref{fig:dataset_example_1} displays several scenes captured in a skatepark, where the quadruped is positioned very close to people, and the individuals appear taller than the platform. Figure~\ref{fig:dataset_example_3} further shows the quadruped traversing stairs and operating under bridges, where the terrain induces significant tilting of the ego coordinate system. These examples clearly demonstrate that the quadruped’s viewpoint is markedly different from that of the vehicle, leading to distinctly varied imaging effects.

For the drone platform, Figure~\ref{fig:dataset_example_1} and Figure~\ref{fig:dataset_example_3} illustrate sample frames captured during flight, showing that targets are predominantly located below the drone. The drone’s inherent jitter further contributes to imaging effects that differ substantially from those observed on the vehicle platform.

In addition, Figure~\ref{fig:dataset_example_2} and Figure~\ref{fig:dataset_example_4} showcase data collected under nighttime conditions across all three platforms. Collectively, these examples underscore the rich diversity of the Pi3DET dataset and highlight the unique challenges associated with cross-platform 3D object detection.

\begin{table*}
\centering
\caption{Summary of the \textbf{cross-platform} and \textbf{cross-dataset} discrepancies in existing 3D  detection datasets.}
\vspace{-0.1cm}
\resizebox{\linewidth}{!}{
\begin{tabular}{cl|c|c|c|c|c|c|c} 
    \toprule
    \multicolumn{2}{c|}{\multirow{2}{*}{\textbf{Dataset}}} & \textbf{Beam} & \textbf{Beam} & \textbf{Points} & \textbf{Training} & \textbf{Validation} & \multirow{2}{*}{\textbf{Night}} & \multirow{2}{*}{\textbf{Condition}}
    \\
    & & \textbf{Ways} & \textbf{Angles} & \textbf{per Scene} & \textbf{Frames} & \textbf{Frames} & 
    \\
    \midrule\midrule
    \multicolumn{2}{c|}{\textbf{nuScenes} \cite{caesar2020nuscenes}} & $32$ & $[-30.0, 10.0]$ & $\sim25$K  & $28,130$ & $6,019$ & Yes & Road 
    \\\midrule
    \multicolumn{2}{c|}{\textbf{KITTI} \cite{geiger2012we}} & $64$ & $[-23.6, 3.20]$ & $\sim118$K & $3,712$ & $3,769$ & No & Road 
    \\\midrule
    \multirow{3}{*}{\makecell{\textbf{Pi3DET}\\\textbf{{\small(Ours)}}}} 
    & Vehicle & \multirow{3}{*}{$64$} & \multirow{3}{*}{$[-22.5, 22.5]$} & $\sim110$K & $16,888$ & $15,305$ & Yes & Road 
    \\
    & Quadruped & & & $\sim87$K & $7,204$ & $5,096$  & Yes & Road, Stair, Under Bridge 
    \\
    & Drone & & & $\sim110$K & $3,584$ & $3,468$ & Yes & Air 
    \\
    \bottomrule
    \end{tabular}}
\label{tab:cross_dataset}
\end{table*}

\subsection{Cross-Platform Discrepancies}
Our statistical analyses and visualizations reveal that cross-platform discrepancies are primarily influenced by differences in the z-axis distribution, object geometry, and target bounding box characteristics. 

Table~\ref{tab:dataset_intensity_vehicle}, Table~\ref{tab:dataset_intensity_drone}, and \ref{tab:dataset_intensity_quadruped} show that while the distributions of \(x\), \(y\), and intensity are largely similar across platforms, significant differences emerge along the \(z\)-axis. This is likely attributable to variations in sensor mounting height and motion space: vehicles, with higher, fixed sensor mounts, tend to produce point clouds concentrated just below the sensor (with \(z\) values slightly below zero); quadruped platforms, operating at lower heights near the ground, generate point clouds with \(z\) values closer to zero; and drone platforms, which operate at even greater altitudes, yield broader \(Z\)-axis distributions that remain mostly below zero. 

\begin{wraptable}{r}{0.56\textwidth}
\centering
\vspace{-0.3cm}
\caption{Summary of notations defined in this work.}
\vspace{-0.2cm}
\resizebox{\linewidth}{!}{
\begin{tabular}{c|l}
    \toprule
    \textbf{Notation} & \textbf{Definition}
    \\\midrule\midrule
    $\beta$ & Platform type
    \\
    \textcolor{pi3det_red}{$\mathcal{S}$} & Symbol denoting the source platform
    \\
    \textcolor{pi3det_blue}{$\mathcal{T}$} & Symbol denoting the target platform
    \\
    $\mathcal{P}$ & LiDAR point cloud
    \\
    $\mathcal{B}$ & 3D bounding box
    \\
    $N$ & Total number of LiDAR point clouds 
    \\
    $M$ & Total number of 3D bounding boxes
    \\
    $(p^{x}, p^{y}, p^{z})$ & Point coordinates in X, Y, Z directions
    \\
    $(c^{x}, c^{y}, c^{z})$ & Center position of the 3D bounding box
    \\
    $l$ & Length of the 3D bounding box
    \\
    $w$ & Width of the 3D bounding box
    \\
    $h$ & Height of the 3D bounding box
    \\
    $\varphi$ & Heading angle of the 3D bounding box
    \\
    $\phi$ & Roll angle of the ego platform
    \\
    $\theta$ & pitch angle of the ego platform
    \\
    $\psi$ & Yaw angle of the ego platform
    \\
    $\mathbf{T}$ & Ego pose
    \\
    $\mathbf{R}$ & Ego rotation
    \\
    $\Delta \phi$ & Random jitter added to the roll angle
    \\
    $\Delta \theta$ & Random jitter added to the pitch angle
    \\
    $ \mathbf{F}^{\textcolor{pi3det_red}{\mathcal{S}}}$ & RoI feature from source platform
    \\
    $ \mathbf{F}^{\textcolor{pi3det_blue}{\mathcal{T}}}$ & RoI feature from target platform
    \\
    \bottomrule
\end{tabular}}
\label{tab:notations}
\vspace{-1cm}
\end{wraptable}

Furthermore, Table~\ref{tab:dataset_bbox_geo_vehicle}, Table~\ref{tab:dataset_bbox_geo_drone}, and Table~\ref{tab:dataset_bbox_geo_quadruped} show that vehicle targets typically measure around $4$–$5$ meters in length, $2$ meters in width, and $1.6$–$1.7$ meters in height (with pedestrians around $1.7$–$1.9$ meters). And the Vehicle platform also exhibits a wider range of object sizes (including larger vehicles like buses or trams exceeding $10$ meters in length). 

Analysis of the number of foreground objects and points per bounding box (Table~\ref{tab:dataset_bbox_vehicle}, Table~\ref{tab:dataset_bbox_drone}, Table~\ref{tab:dataset_bbox_quadruped}) further indicates that the Vehicle platform generally contains more diverse and numerous targets, while some sequences from the Drone and Quadruped platforms may include only pedestrian targets. 

In summary, our analyses demonstrate that differences in ego height and motion space significantly affect the \(z\)-axis distribution of LiDAR point clouds, leading to inconsistent object representations and spatial misalignments across platforms. These discrepancies pose considerable challenges for developing robust cross-platform 3D detection methods.

\subsection{Comparisons with Other Datasets}
In our experiments, we leverage two widely recognized datasets: nuScenes \cite{caesar2020nuscenes} and KITTI \cite{geiger2012we} to evaluate cross-platform and cross-dataset 3D object detection. Both datasets have distinct characteristics that contribute to the domain gap. Below is a summary of their key attributes:
\begin{itemize}
    \item \textbf{nuScenes} \cite{caesar2020nuscenes} is a large-scale autonomous driving dataset collected from urban environments in Boston and Singapore. It employs a $32$-beam LiDAR (Velodyne HDL-$32$E) alongside high-resolution cameras and radar to provide a comprehensive, multimodal view of complex urban scenes. The dataset encompasses approximately $1{,}000$ scenes, with each scene lasting around $20$ seconds, and includes roughly $28{,}130$ training frames, $6{,}019$ validation frames, and $6{,}008$ test frames. These frames capture a wide variety of weather conditions, traffic densities, and dynamic urban scenarios, making nuScenes a challenging benchmark for 3D object detection and tracking tasks.

    \item \textbf{KITTI} \cite{geiger2012we} is one of the pioneering datasets for autonomous driving research, widely recognized for its high-quality 3D annotations and real-world driving scenarios. Captured using a $64$-beam LiDAR (Velodyne HDL-$64$E) mounted on a vehicle, KITTI provides precise 3D point clouds over suburban and urban landscapes under relatively consistent weather conditions. The dataset is divided into roughly $7{,}481$ training frames and $7{,}518$ test frames, with detailed labels for objects such as vehicles, pedestrians, and cyclists. The comprehensive sensor data and annotations have established it as a fundamental benchmark for evaluating 3D object detection algorithms, despite its smaller scale compared to more recent datasets.
\end{itemize}

Table~\ref{tab:cross_dataset} provides an overview of key discrepancies across datasets and platforms. The nuScenes dataset, collected using a $32$-beam LiDAR, offers a balanced set of urban road scenes with both daytime and nighttime data. In contrast, KITTI, captured with a $64$-beam sensor, presents higher point density per scene but lacks nighttime data. 

Pi3DET spans three platforms, each utilizing a $64$-beam LiDAR with a uniform angular range of \([-22.5^\circ, 22.5^\circ]\). The Vehicle subset focuses on road environments with abundant training and validation frames, while the Quadruped subset captures more diverse terrains, including roads, stairs, and under bridges. The Drone subset, acquired in aerial environments, offers a comparable point density to the Vehicle subset. 

These differences highlight the diverse sensor configurations and environmental conditions, underscoring the challenges inherent in cross-dataset and cross-platform 3D detection.

\subsection{Cross-Platform Annotation Toolkit}
\label{sec:cross-platform-annotation-toolkit}
Our annotation process for Pi3DET is executed through a streamlined three-stage pipeline, which is described below.

\subsubsection{Pseudo-Label Generation} 
We pre-trained a diverse set of state-of-the-art 3D object detectors (PV-RCNN \cite{shi2020pv}, PV-RCNN++ \cite{shi2023pv}, Voxel-RCNN \cite{deng2021voxel}, IA-SSD \cite{zhang2022not}, CenterPoint \cite{yin2021center}, and SECOND \cite{yan2018second}) on external datasets such as Waymo \cite{sun2020scalability}, nuScenes \cite{caesar2020nuscenes}, and Lyft \cite{houston2021one}, and then used these models to infer initial pseudo-labels on the \textbf{Pi3DET} data.

\subsubsection{Pseudo-Label Optimization and Filtering}
We applied a kernel density estimation (KDE) algorithm to fuse predictions from multiple 3D object detectors and used the 3D multi-object tracking algorithm CTRL \cite{fan2023once} to ensure temporal consistency and to interpolate missed detections. 

In addition, we employed the vision foundation model Tokenize Anything (TA) \cite{pan2024tokenize} to project pseudo-labels onto corresponding RGB images and verify object categories within an open vocabulary. This step maps the TA outputs to the Pi3DET classes (Vehicle, Pedestrian), with mismatches flagged for manual review.

\subsubsection{Manual Refinement} 
Using the open-source 3D annotation platform Xtreme1\footnote{\url{https://github.com/xtreme1-io/xtreme1}.}, three annotators manually refined each frame on a per-box basis. This process, which included cross-validation among multiple annotators, ensured that the final annotations are both precise and consistent.

This comprehensive annotation toolkit integrates modules for data visualization, model pre-training, multi-object tracking, 3D bounding box editing, and vision model inference. Although our automated framework greatly reduced the manual workload, the inherent sparsity and irregularity of point cloud data required an average of over $30$ seconds of manual intervention per frame, culminating in \textbf{more than $\mathbf{500}$ hours of annotation effort} for the entire \textbf{Pi3DET} dataset.

\begin{figure*}[t]
    \centering
    \includegraphics[width=\textwidth]{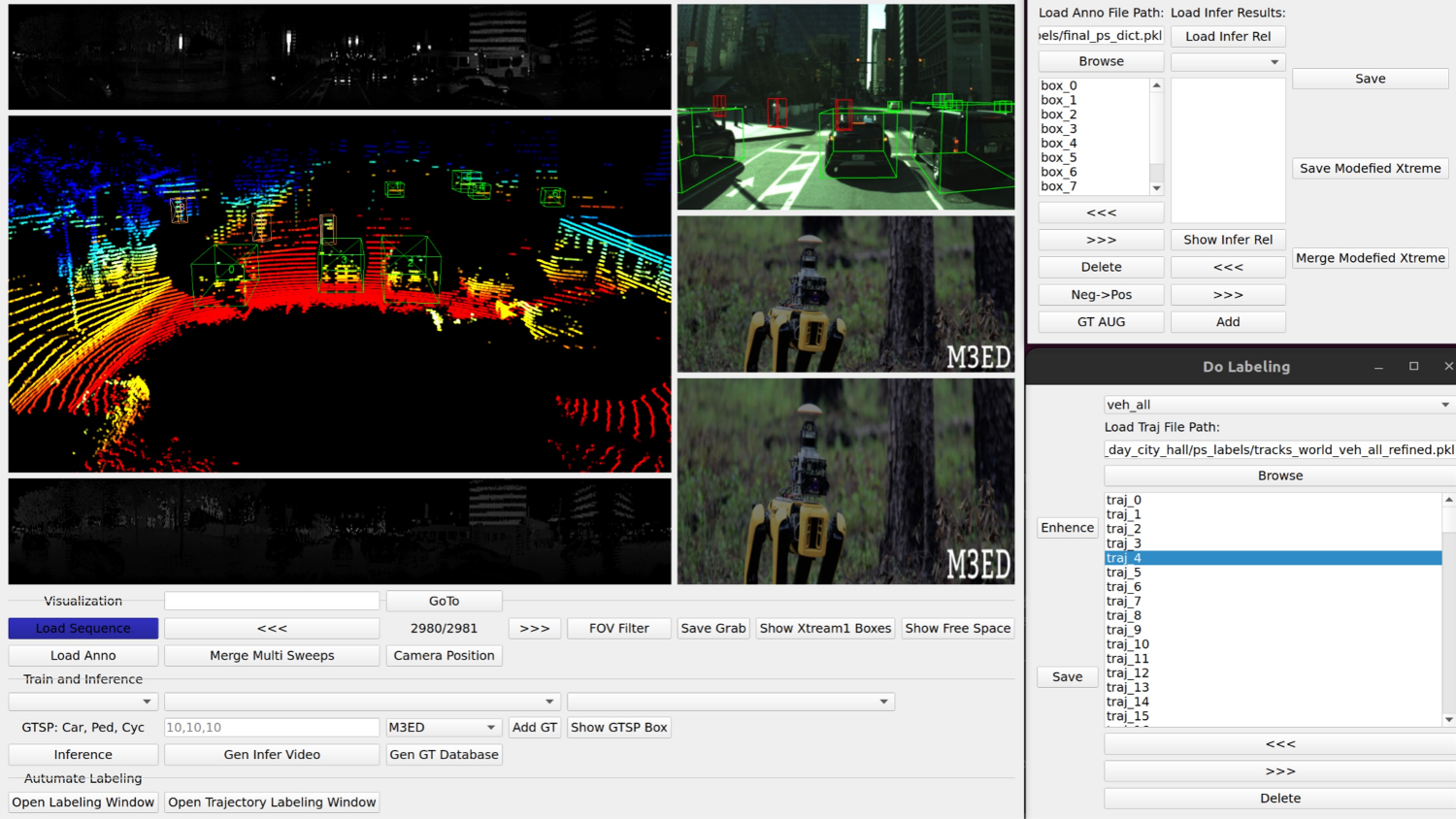}
    \vspace{-0.5cm}
    \caption{\textbf{Model Pre-Training Interface:} This interface enables the pre-training of various 3D detection models to generate initial pseudo labels for subsequent processing.}    
    \label{fig:tool_1}
\end{figure*}

Our annotation pipeline is further illustrated by several figures. Figure~\ref{fig:tool_1} depicts the pseudo-label generation process, where multiple pre-trained 3D detectors infer initial labels from the raw Pi3DET data. 

\begin{figure*}[t]
    \centering
    \includegraphics[width=\textwidth]{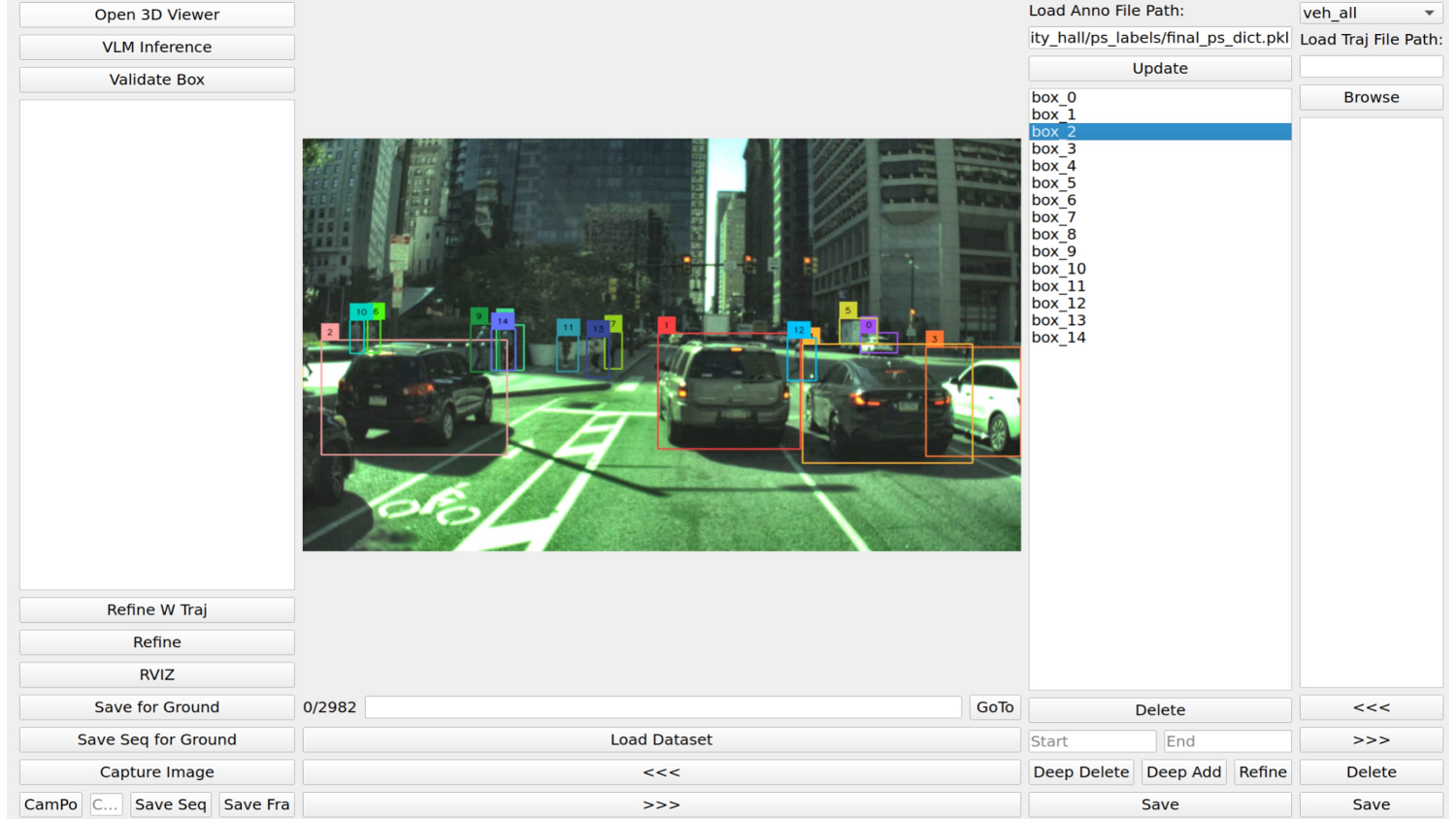}
    \vspace{-0.5cm}
    \caption{\textbf{Pseudo-Label Filtering Interface:} In this view, 3D bounding boxes are projected onto corresponding RGB images, facilitating efficient and convenient filtering of pseudo labels.}    \label{fig:tool_2}
\end{figure*}

\begin{figure*}[t]
    \centering
    \includegraphics[width=\textwidth]{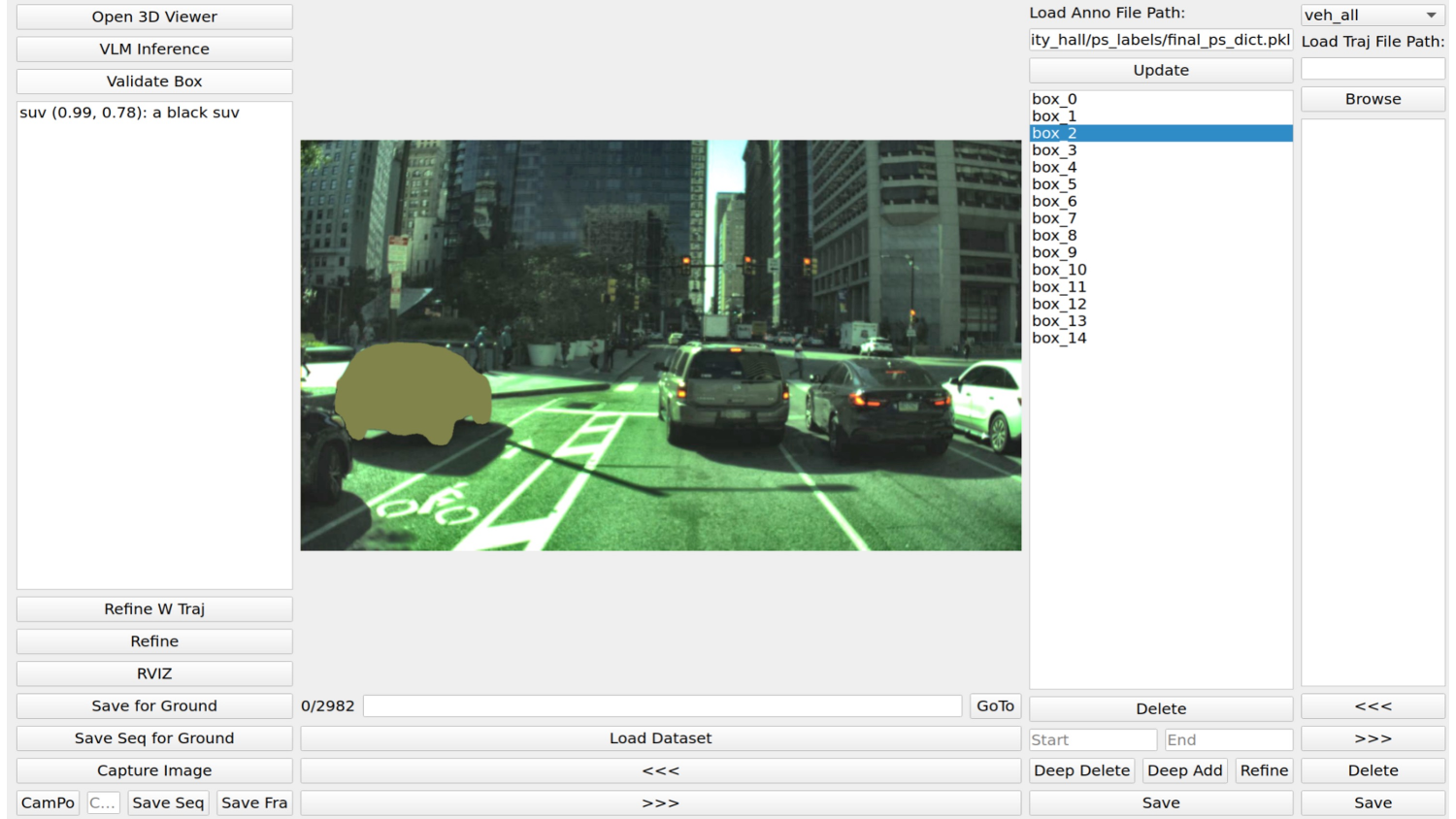}
    \vspace{-0.5cm}
    \caption{\textbf{Automatic Pseudo-Label Screening} with TA \cite{pan2024tokenize}. This interface employs a vision foundation model (Tokenize Anything) to automatically filter pseudo labels by verifying alignment with image content, with mismatched frames flagged for manual review.}    \label{fig:tool_3}
\end{figure*}

Figure~\ref{fig:tool_2} and Figure~\ref{fig:tool_3} demonstrate the pseudo-label optimization and filtering stage, highlighting how kernel density estimation and the CTRL tracking algorithm fuse detector outputs and maintain temporal consistency, while the Tokenize Anything model \cite{pan2024tokenize} verifies the projected labels on RGB images. 

\begin{figure*}[t]
    \centering
    \includegraphics[width=\textwidth]{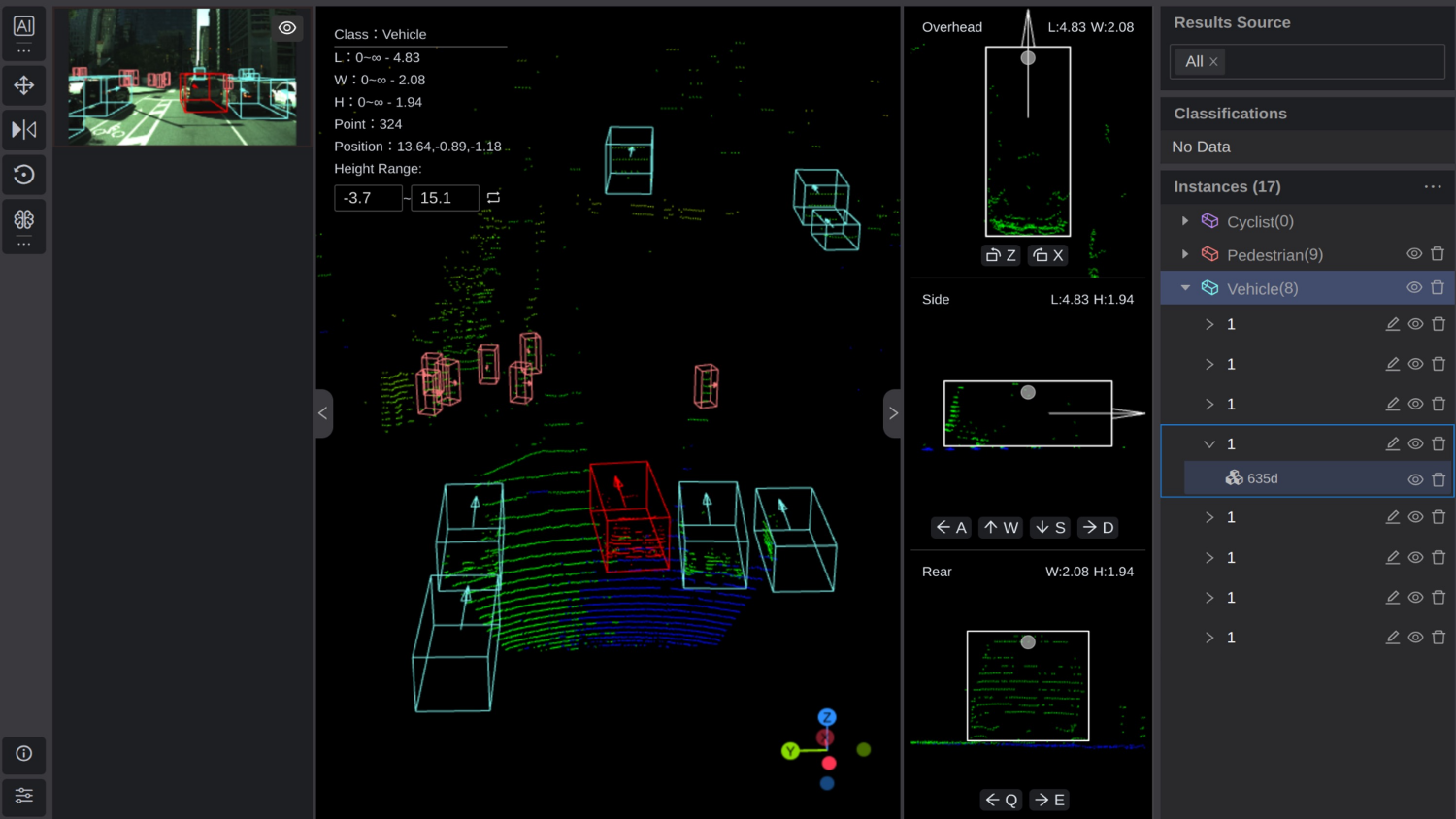}
    \vspace{-0.5cm}
    \caption{\textbf{Manual Refinement Interface:} Utilizing the open-source Xtreme1 platform, this interface allows annotators to perform detailed frame-by-frame and box-by-box corrections, ensuring high-quality final annotations.}    
    \label{fig:tool_4}
\end{figure*}

Finally, Figure~\ref{fig:tool_4} showcases the manual refinement interface provided by the Xtreme1 platform, where annotators conduct frame-by-frame corrections and cross-validation to ensure high annotation accuracy. These visualizations underscore the comprehensive and multi-faceted nature of our annotation toolkit, which has been instrumental in achieving a high-quality and consistent Pi3DET dataset.

\subsection{License}
The \textbf{Pi3DET} dataset and the associated benchmark are released under the Attribution-ShareAlike 4.0 International (CC BY-SA 4.0)\footnote{\url{https://creativecommons.org/licenses/by-sa/4.0/legalcode}.} license.

\section{Additional Implementation Details}
\label{sec:additional_implementation}

In this section, we provide additional implementation details to facilitate a thorough understanding and reproducibility of our work. We begin by describing the construction of our benchmark, which leverages data from three platforms in Pi3DET, as well as two widely used datasets (nuScenes \cite{caesar2020nuscenes} and KITTI \cite{geiger2012we}). 

Based on these sources, we construct a total of \textbf{eight} cross-platform and cross-dataset adaptation tasks. The cross-platform adaptation tasks involve various combinations of Vehicle, Drone, and Quadruped subsets from \textbf{Pi3DET}, while the cross-dataset tasks evaluate the domain gap between nuScenes and other vehicle data (Pi3DET and KITTI \cite{geiger2012we}).

Following the benchmark construction, we summarize the notations used throughout our work in Table~\ref{tab:notations} for better clarity. We then detail our training configurations and evaluation protocols, which include specific settings used for both detection and adaptation baselines. Finally, we provide an overview of the detection baselines and adaptation baselines employed in our experiments.

The subsequent subsections elaborate on these aspects in detail, ensuring that all experimental and implementation choices are clearly documented.

\subsection{Benchmark Construction}
\label{sec:supp_b.1}
Building upon three platforms from \textbf{Pi3DET}, as well as the other two datasets (nuScenes \cite{caesar2020nuscenes} and KITTI \cite{geiger2012we}), we construct a total of \textbf{eight} cross-platform and cross-dataset adaptation tasks. These tasks are summarized as follows.

\begin{itemize}
    \item \textbf{Cross-Platform Adaptation:}
    
    \begin{itemize}
    \item Pi3DET (Vehicle) $\rightarrow$ Pi3DET (Drone)
    
    \item nuScenes (Vehicle) $\rightarrow$ Pi3DET (Drone)

    \item Pi3DET (Vehicle) $\rightarrow$ Pi3DET (Quadruped)

    \item nuScenes (Vehicle) $\rightarrow$ Pi3DET (Quadruped)

    \item Pi3DET (Quadruped) $\rightarrow$ Pi3DET (Drone)
    
    \item Pi3DET (Drone) $\rightarrow$ Pi3DET (Quadruped)

    \end{itemize}

    \item \textbf{Cross-Dataset Adaptation:}

    \begin{itemize}
    \item nuScenes $\rightarrow$ Pi3DET (Vehicle)

    \item nuScenes $\rightarrow$ KITTI
    \end{itemize}
\end{itemize}

For the cross-platform adaptation tasks, we adopt PV-RCNN \cite{shi2020pv} and Voxel-RCNN \cite{deng2021voxel} as the base 3D detectors. These state-of-the-art detectors utilize anchor-based and center-based detection heads, respectively, thereby covering the most popular 3D detection settings and demonstrating the generality of our approach. 

For the cross-dataset adaptation tasks, we essentially employ the same configuration as in the cross-platform tasks; however, when KITTI \cite{geiger2012we} serves as the target dataset, we use the SECOND-IOU \cite{yan2018second, yan2018second} model, which is widely used in current cross-dataset methods to facilitate direct comparisons with reported results and highlight the effectiveness of our method. The data splits for each platform and dataset are summarized in Table~\ref{tab:cross_dataset}.

\subsection{Summary of Notations}
\label{sec:supp_b.2}
For better readability, the notations used in this work have been summarized in Table~\ref{tab:notations}.

\subsection{Training Configurations}
\label{subsec: training_configurations}

For all datasets, the detection range is fixed to \([-75.2\,\text{m}, 75.2\,\text{m}]\) along the \(X\) and \(Y\) axes and \([-2\,\text{m}, 4\,\text{m}]\) along the \(Z\) axis, with coordinate origins shifted to the ground plane. The voxel size is consistently set to \((0.1\,\text{m}, 0.1\,\text{m}, 0.15\,\text{m})\) across datasets. Data augmentation is widely adopted during both pre-training and self-training; this includes random world flipping, scaling, and rotation, as well as random object rotation. In addition, Pi3DET-Net incorporates Random Platform Jitter, where rotations around the \(x\) and \(y\) axes (\(\Delta \phi\)) are uniformly sampled from \([-5^\circ, +5^\circ]\).

Our pre-training framework is built upon the open-source OpenPCDet project\footnote{\url{https://github.com/open-mmlab/OpenPCDet}.} and is executed using two NVIDIA Titan RTX GPUs. For cross-platform tasks, 3D detectors are initially pre-trained on nuScenes with optimization settings that include a batch size of $4$ per GPU for $20$ epochs, use of the Adam optimizer with an initial learning rate of $0.01$, weight decay of $0.001$, and momentum of $0.9$. 

For cross-platform tasks on Pi3DET, we extend the pre-training to $40$ epochs. In the cross-dataset adaptation tasks, we use the detector weights pre-trained on nuScenes from the ST3D++ framework\footnote{\url{https://github.com/CVMI-Lab/ST3D}.} to ensure fairness in comparisons.

\subsection{Evaluation Protocols}
\label{sec:supp_b.4}
We follow \cite{yang2021st3d} and adopt the KITTI evaluation metric for the common category \textit{Vehicle} (referred to as \textit{Car} in the KITTI and nuScenes dataset). Our evaluation protocol uses the official KITTI criteria, reporting average precision (AP) in both bird’s-eye view (BEV) and 3D over $40$ recall positions. Mean average precision is computed with an IoU threshold of $0.7$ for cars and $0.5$ for pedestrians and cyclists. For all tasks and datasets, the prediction confidence threshold for 3D detectors is set to $0.2$.

For 3D IoU, given a predicted 3D box \(B_p\) and its corresponding ground truth \(B_{gt}\), the IoU is calculated as:

\begin{align}
\text{IoU} = \frac{\text{Vol}(B_p \cap B_{gt})}{\text{Vol}(B_p \cup B_{gt})}~,
\end{align}

For BEV, the IoU is computed similarly using the 2D projections of the 3D boxes onto the ground plane.

The average precision (AP) is computed as follows:
\begin{align}
\mathrm{AP} = \frac{1}{40} \sum_{i=1}^{40} p_{\mathrm{interp}}(r_i)~,
\end{align}
where \( r_i \) represents the \( i \)-th recall threshold (typically evenly spaced over the recall range), and \( p_{\mathrm{interp}}(r_i) \) is the interpolated precision defined as follows:
\begin{align}
p_{\mathrm{interp}}(r_i) = \max_{\tilde{r} \ge r_i} p(\tilde{r})~,
\end{align}
with \( p(\tilde{r}) \) denoting the precision at recall \( \tilde{r} \)~.

\subsection{Summary of Detection Baselines}
\label{sec:supp_b.5}
The following 3D object detection methods are used as baselines in our \textbf{Pi3DET} benchmark.
\begin{itemize}
    \item \textbf{PV-RCNN} \cite{shi2020pv}
    is a two-stage 3D detection framework that effectively combines voxel-based and point-based representations. In the first stage, the model aggregates voxel features into keypoints via a voxel set abstraction module, which enables efficient proposal generation. In the second stage, PV-RCNN employs a RoI grid pooling module that leverages point-wise features to refine the candidate proposals, thereby achieving high localization accuracy and robust performance.

    \item \textbf{Voxel-RCNN} \cite{deng2021voxel}
    is another two-stage detector that primarily relies on voxel representations. It integrates a voxel feature encoder for both proposal generation and refinement, enabling precise region proposal extraction from high-dimensional sparse data. The design emphasizes efficient voxel-based processing, reducing computational overhead while maintaining competitive accuracy in 3D object detection.

    \item \textbf{SECOND} \cite{yan2018second}, also termed as Sparsely Embedded Convolutional Detection, is a one-stage 3D detector that capitalizes on sparse convolutional networks to process voxelized point clouds. By converting irregular point cloud data into a structured voxel representation, SECOND applies sparse convolution operations to efficiently extract features and directly predict object classes and bounding boxes in a single forward pass. This design achieves a favorable trade-off between detection speed and accuracy, making it a popular baseline in many 3D detection studies. Following the design proposed in ST3D++ \cite{yang2022st3d++}, we improve the SECOND detector by incorporating an additional IoU head to estimate the IoU between object proposals and their corresponding ground truths, naming the modified detector \textbf{SECOND-IoU}. 
\end{itemize}

In our experiments, PV-RCNN and Voxel-RCNN are two-stage detectors that respectively employ anchor-based and center-based detection heads, while SECOND is a one-stage detector. This comprehensive setting covers a broad range of popular 3D detection designs, thereby demonstrating the generality of our proposed approach.

\subsection{Summary of Adaptation Baselines}
\label{sec:supp_b.6}

The following cross-domain 3D object detection methods are used as baselines in our \textbf{Pi3DET} benchmark.
\begin{itemize}
    \item \textbf{ST3D} \cite{yang2021st3d}
    is a self-training pipeline designed for cross-dataset adaptation on 3D object detection from point clouds. ST3D consists of three key components: 1) Random Object Scaling (ROS), which mitigates source domain bias by randomly scaling 3D objects during pre-training; 2) Quality-Aware Triplet Memory Bank (QTMB), which generates high-quality pseudo labels by assessing localization quality and avoiding ambiguous examples; and 3) Curriculum Data Augmentation (CDA), which progressively increases the intensity of data augmentation to prevent overfitting to easy examples and improve the ability to handle hard cases. ST3D iteratively improves the detector on the target domain by alternating between pseudo label generation and model training, achieving state-of-the-art performance on multiple 3D object detection datasets, even surpassing fully supervised results in some cases.
    
    \item \textbf{ST3D++} \cite{yang2022st3d++}
    introduces a holistic pseudo-label denoising pipeline to reduce noise in pseudo-label generation and mitigate the negative impacts of noisy pseudo labels on model training. The pipeline consists of three key components: 1) Random Object Scaling (ROS), which reduces object scale bias during pre-training; 2) Hybrid Quality-Aware Triplet Memory (HQTM), which improves the quality and stability of pseudo labels through a hybrid scoring criterion and memory ensemble; and 3) Source-Assisted Self-Denoised Training (SASD) and Curriculum Data Augmentation (CDA), which rectify noisy gradient directions and prevent overfitting to easy examples. ST3D++ achieves state-of-the-art performance on multiple 3D object detection datasets, even surpassing fully supervised results in some cases, and demonstrates robustness across various categories such as cars, pedestrians, and cyclists. The method is model-agnostic and can be integrated with different 3D detection architectures.

    \item \textbf{MS3D++} \cite{tsai2024ms3d++}
    is a multi-source self-training framework designed for cross-dataset 3D object detection. The method addresses the significant performance drop ($70$-$90$\%) that occurs when 3D detectors are deployed in unfamiliar domains due to variations in lidar types, geography, or weather. MS3D++ generates high-quality pseudo-labels by leveraging an ensemble of pre-trained detectors from multiple source domains, which are then fused using Kernel-density estimation Box Fusion (KBF) to improve domain generalization. Temporal refinement is applied to ensure consistency in box localization and object classification. The framework also includes a multi-stage self-training process to iteratively improve pseudo-label quality, balancing precision and recall. Experimental results on datasets like Waymo \cite{sun2020scalability}, nuScenes \cite{caesar2020nuscenes}, and Lyft \cite{houston2021one} demonstrate that MS3D++ achieves state-of-the-art performance, comparable to training with human-annotated labels, particularly in Bird's Eye View (BEV) evaluation for both low and high-density lidar. The approach is highly versatile, allowing easy integration with various 3D detector architectures and data augmentation techniques without modifying the inference runtime of the detector.

    \item \textbf{ReDB} \cite{chen2023revisiting}
    aims to generate reliable, diverse, and class-balanced pseudo labels to iteratively guide self-training on a target dataset with a different distribution. The framework includes a cross-domain examination (CDE) to assess pseudo label reliability, an overlapped boxes counting (OBC) metric to ensure geometric diversity, and a class-balanced self-training strategy to address inter-class imbalance.
\end{itemize}

\section{Additional Experimental Analyses}
\label{sec:additional_experiments}

\begin{wrapfigure}{r}{0.56\textwidth}
% \begin{figure}[t]
    \centering
    \includegraphics[width=\linewidth]{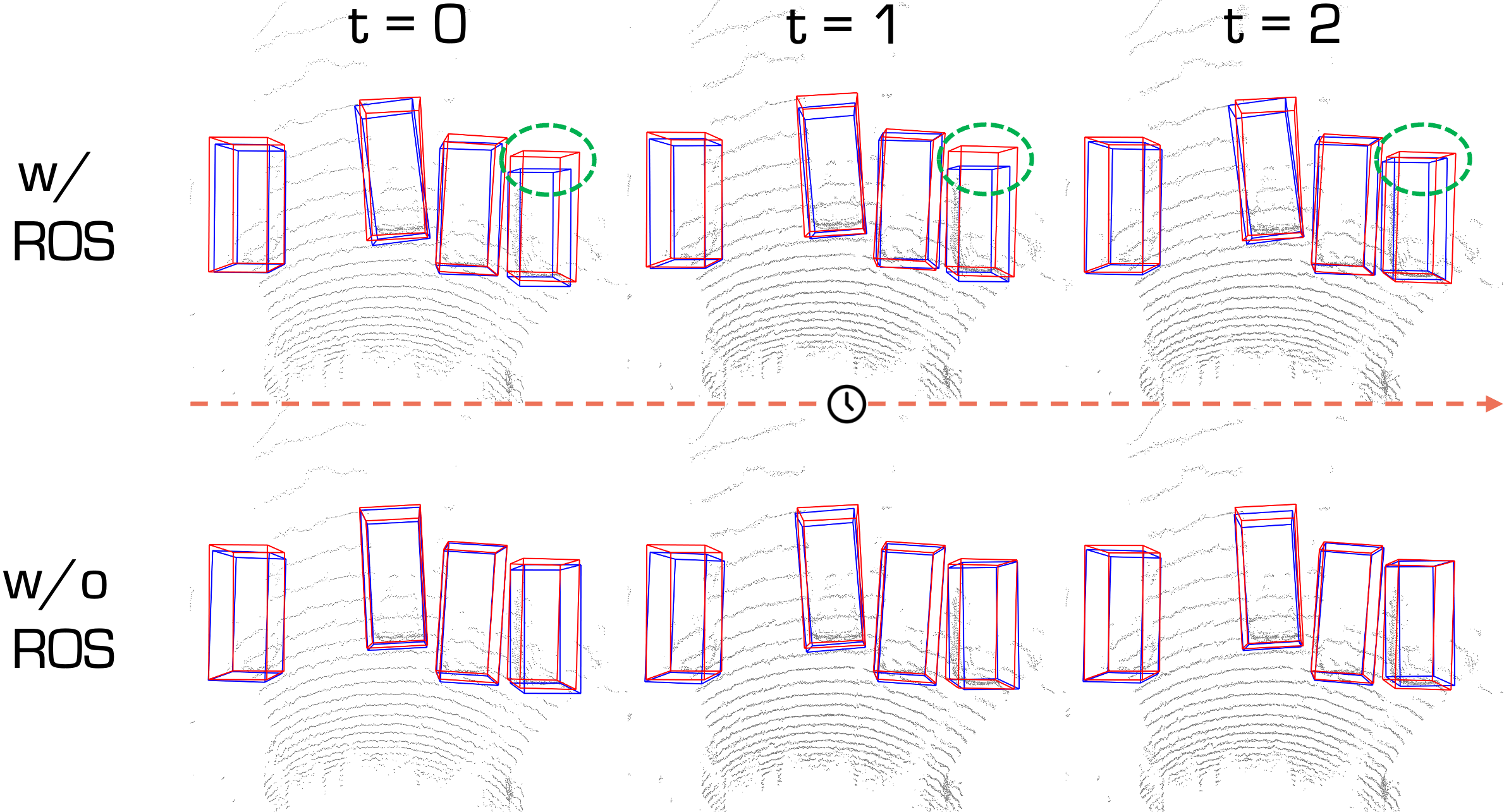}
    \vspace{-0.6cm}
    \caption{Comparisons of inference results in a continuous static scene using PV-RCNN with and without ROS. The \textcolor{red}{\textbf{red boxes}} indicate ground truth, while the \textcolor{blue}{\textbf{blue boxes}} denote predictions from the detector. Despite the ego vehicle and surrounding objects remaining static, the ROS-pretrained PV-RCNN yields variable predictions, whereas the model without ROS produces much more stable and consistent outputs.}
    \label{fig:ros_adverse_effects}
    \vspace{-1cm}
\end{wrapfigure}

In this section, we present additional results to complement the findings reported in the main paper. First, we provide further quantitative results that reinforce our evaluation of cross-platform and cross-dataset adaptation performance. Next, we offer qualitative results with visual examples that highlight both the strengths and potential weaknesses of our approach. Finally, we analyze failure cases to identify specific scenarios where our method struggles, thereby offering insights for future improvements.

\subsection{Additional Quantitative Results}
\label{sec:supp_c.1}

\subsubsection{Adverse Effects of Random Object Scaling (ROS)}
Random Object Scaling (ROS) is a data augmentation technique introduced in ST3D \cite{yang2021st3d} for cross-dataset 3D object detection. The primary goal of ROS is to enhance the diversity of foreground objects in the source domain by randomly scaling the sizes of ground-truth bounding boxes. This augmentation strategy aims to mitigate the bias inherent in object size distributions, thereby improving the detector’s ability to extract robust foreground features. 

\begin{table}[htbp]
\centering
\begin{minipage}{0.462\linewidth}
  \centering
% \usepackage{multirow}
% \usepackage{booktabs}

% \begin{wraptable}{l}{0.6\textwidth}
% \begin{table}
\centering
\caption{Ablation study on effects of the random object scaling (ROS) operation on the pseudo label quality.}
\vspace{-0.2cm}
\resizebox{\linewidth}{!}{
\begin{tabular}{l|c|c|c} 
\toprule
    \textbf{Method} & \textbf{ROS} & \textbf{AP@0.70} & \textbf{AP@0.50}         
    \\ 
    \midrule\midrule
    \multirow{2}{*}{PV-RCNN \cite{shi2020pv}} & \textcolor{pi3det_red}{\xmark} & $37.84$ / $30.20$ & $39.83$ / $39.28$  
    \\
    & \textcolor{pi3det_blue}{\cmark} & $17.96$ / $12.02$ & $29.26$ / $24.58$  
    \\ 
    \midrule
    \multirow{2}{*}{SECOND-IOU \cite{yan2018second}} & \textcolor{pi3det_red}{\xmark} & $32.47$ / $28.21$ & $38.76$ / $37.25$  
    \\
    &  \textcolor{pi3det_blue}{\cmark}  & $19.75$ / $10.40$ & $36.14$ / $31.94$  
    \\
    \bottomrule
\end{tabular}}
\label{tab:ros_setting}
% \vspace{-0.2cm}
% \end{wraptable}
\end{minipage}
\hfill
\begin{minipage}{0.5\linewidth}
  \centering
  \centering
\caption{Ablation study on the effect of different angle setups in the proposed Random Platform Jitter (RPJ).}
\vspace{-0.2cm}
\resizebox{\linewidth}{!}{
\begin{tabular}{c|cc|cc} 
\toprule
\multirow{2}{*}{\textbf{Angle}} & \multicolumn{2}{c|}{\textbf{Vehicle to Quadruped}} & \multicolumn{2}{c}{\textbf{Vehicle to Drone}}
\\
& \textbf{AP@0.70} & \textbf{AP@0.50} & \textbf{AP@0.70} & \textbf{AP@0.50}             
\\\midrule\midrule
$\pm0^\circ$  & $38.61$ / $26.84$ & $40.64$ / $39.22$    & $57.29$ / $36.62$ & $58.92$ / $56.19$  \\
$\pm3^\circ$                    & $40.87$ / $28.46$ & $44.14$ / $41.21$    & $61.95$ / $40.52$ & $63.89$ / $59.75$  \\
$\pm5^\circ$                    & $42.54$ / $30.03$ & $46.54$ / $43.02$    & $56.47$ / $34.69$ & $56.23$ / $54.65$  \\
$\pm8^\circ$                    & $30.55$ / $21.41$ & $35.29$ / $30.88$    & $41.03$ / $26.17$ & $48.65$ / $42.58$  \\
\bottomrule
\end{tabular}}
\label{tab:rpj_setting}
\end{minipage}
\end{table}

In cross-dataset tasks such as nuScenes \cite{caesar2020nuscenes} $\rightarrow$ KITTI \cite{geiger2012we}, Waymo \cite{sun2020scalability} $\rightarrow$ KITTI \cite{geiger2012we}, and Waymo \cite{sun2020scalability} $\rightarrow$ nuScenes \cite{caesar2020nuscenes}, ROS has demonstrated considerable benefits and has been adopted by subsequent methods, including ST3D++ \cite{yang2022st3d++} and ReDB \cite{chen2023revisiting}.

However, our experiments on the Pi3DET dataset reveal that ROS has a deleterious effect on pseudo-label quality, particularly in high-frequency annotated data. Pi3DET is annotated at $10$ Hz, meaning that in consecutive frames, although the LiDAR point clouds exhibit subtle variations due to sensor noise and slight motion, the positions and sizes of foreground objects remain essentially constant. 

Under these conditions, ROS inadvertently exaggerates minor variations in object size, causing the detector to produce inconsistent predictions across similar frames. For instance, when evaluating a nuScenes $\rightarrow$ Pi3DET (Vehicle) cross-dataset task, we observed that PV-RCNN \cite{shi2020pv} and SECOND-IoU \cite{yan2018second} models pre-trained with ROS experienced performance drops of approximately $60\%$ and $63\%$ respectively in $AP_{3D}$, as detailed in Table~\ref{tab:ros_setting}. 

\begin{table}
% \begin{wraptable}{l}{0.7\textwidth}
\centering
\caption{
    \textbf{Comparisons among state-of-the-art 3D detection algorithms for nuScenes $\rightarrow$ Pi3DET (Vehicle) adaptation}. We report the average precision (AP) in ``BEV / 3D'' at the IoU thresholds of $0.70$ and $0.50$, respectively. Symbol $\ddagger$ denotes algorithms \textit{w.o.} ROS. All scores are given in percentage (\%). The \textcolor{pi3det_red}{\hlred{Best}} and \textcolor{pi3det_blue}{\hlblue{Second Best}} scores under each metric are highlighted in \textcolor{pi3det_red}{\hlred{Red}} and \textcolor{pi3det_blue}{\hlblue{Blue}}, respectively.
}
\vspace{-0.2cm}
\resizebox{0.8\linewidth}{!}{
\begin{tabular}{c|l|cc|cc} 
    \toprule
    \multirow{2}{*}{\textbf{Setting}} & \multirow{2}{*}{\textbf{Method}} & \multicolumn{2}{c|}{PV-RCNN \cite{shishaoshuai2020pv}} & \multicolumn{2}{c}{Voxel RCNN-C \cite{deng2021voxel}}
    \\
    & & \textbf{AP@0.70} & \textbf{AP@0.50} & \textbf{AP@0.70} & \textbf{AP@0.50}
    \\ 
    \midrule\midrule
    \multirow{11}{*}{\rotatebox{90}{nuScenes \cite{caesar2020nuscenes} $\rightarrow$ Pi3DET}} & \textcolor{gray}{Source Dataset} & \textcolor{gray}{$37.84$ / $30.20$} & \textcolor{gray}{$39.83$ / $39.28$} & \textcolor{gray}{$45.13$ / $34.14$} & \textcolor{gray}{$53.27$ / $51.20$}
    \\\cmidrule{2-6}
    & SN \cite{wang2020train} & $23.23$ / $14.91$ & $38.27$ / $33.51$ & - / -         & - / -
    \\
    & ST3D \cite{yang2021st3d} & $55.40$ / $37.92$ & $63.26$ / $57.67$ & $50.89$ / $39.10$ & $56.83$ / $55.32$
    \\
    & ST3D$^{\ddagger}$ \cite{yang2021st3d} & $56.42$ / $44.40$ & $64.11$ / $58.37$ & $52.55$ / $40.47$ & $58.75$ / $56.1$
    \\
    & ST3D++ \cite{yang2022st3d++} & $58.55$ / $47.19$ & $60.23$ / $59.72$ & $54.48$ / $43.99$ & $60.03$ / $57.46$
    \\
    & ST3D++$^{\ddagger}$ \cite{yang2022st3d++} & $58.93$ / $47.34$ & $60.75$ / $60.33$ & $53.83$ / $44.16$ & $59.59$ / $57.05$
    \\
    & REDB \cite{chen2023revisiting} & $51.65$ / $43.50$ & $58.70$ / $52.70$ & - / -         & - / -
    \\
    & MS3D++ \cite{tsai2024ms3d++} & \textcolor{pi3det_blue}{\hlblue{$59.48$}} / \textcolor{pi3det_blue}{\hlblue{$50.83$}} & \textcolor{pi3det_blue}{\hlblue{$65.92$}} / \textcolor{pi3det_blue}{\hlblue{$64.89$}} & \textcolor{pi3det_blue}{\hlblue{$56.14$}} / \textcolor{pi3det_blue}{\hlblue{$47.61$}} & \textcolor{pi3det_blue}{\hlblue{$62.58$}} / \textcolor{pi3det_red}{\hlred{$61.50$}}
    \\
    & \textbf{Pi3DET-Net} & \textcolor{pi3det_red}{\hlred{$64.29$}} / \textcolor{pi3det_red}{\hlred{$54.76$}} & \textcolor{pi3det_red}{\hlred{$66.77$}} / \textcolor{pi3det_red}{\hlred{$66.21$}} & \textcolor{pi3det_red}{\hlred{$57.12$}} / \textcolor{pi3det_red}{\hlred{$48.98$}} & \textcolor{pi3det_red}{\hlred{$63.36$}} / \textcolor{pi3det_blue}{\hlblue{$61.03$}}
    \\\cmidrule{2-6}
    & Target Platform & $70.48$ / $62.77$ & $75.28$ / $70.13$ & $68.47$ / $58.44$ & $73.29$ / $68.56$ 
    \\
    \bottomrule
\end{tabular}}
\label{tab:nusc_v_pi3det_v_benchmark}
\end{table}

Further analysis indicates that the adverse effects of ROS are mainly due to its sensitivity to high-frequency data. As illustrated in Figure~\ref{fig:ros_adverse_effects}, in a continuous scene where both the ego vehicle and surrounding objects are static, the ROS augmentation leads to varying outputs even though the actual scene remains unchanged, whereas detectors without ROS produce much more temporally stable pseudo labels. This inconsistency in the predictions results in a higher rate of false negatives and false positives during pseudo-label generation, thereby misleading the subsequent self-training process.

Consequently, to ensure a fair comparison and maintain stable pseudo label quality, we opted not to apply ROS during the pre-training phase for all our experiments on the Pi3DET dataset. For completeness, we also evaluated variants of ST3D \cite{yang2021st3d} and ST3D++ \cite{yang2022st3d++} without ROS during self-training. Our findings underscore that, while ROS can be beneficial in datasets with lower annotation frequencies, its application in high-frequency scenarios like Pi3DET can be counterproductive, and thus must be carefully reconsidered for such settings.

\subsubsection{Ablation Study on Random Platform Jitter (RPJ)}
In our analysis of cross-platform LiDAR imaging discrepancies, we identified ego motion -- specifically, sensor jitter -- as a key factor induced by different platform dynamics. Vehicles typically travel on smooth, gently sloping roads, so their 6D ego poses (relative to the world coordinate system) exhibit minimal or gradual changes in pitch and roll. 

In contrast, the Quadruped platform, although also operating on the ground, experiences significant variations in pitch and roll due to mechanical vibrations and unique actions (such as crouching, standing, and turning). The Drone platform, with its greater degrees of freedom, exhibits an even broader distribution of view angles. This motivated our use of Random Platform Jitter during pre-training to simulate these dynamic variations.

To explore the impact of jitter augmentation, we experimented with three settings for randomly rotating the scene around the \(x\) and \(y\) axes: \(\pm3^\circ\), \(\pm5^\circ\), and \(\pm8^\circ\). Our experiments were conducted using the PV-RCNN model on two cross-platform tasks: Pi3DET (Vehicle) \(\rightarrow\) Pi3DET (Quadruped) and Pi3DET (Vehicle) \(\rightarrow\) Pi3DET (Drone). The results are shown in Table~\ref{tab:rpj_setting}. 

We observed that a jitter range of \(\pm5^\circ\) yields a $3.2$ AP@$0.7$ gain for the Vehicle-to-Quadruped task, while a smaller range of \(\pm3^\circ\) is more effective for the Vehicle-to-Drone task, resulting in a $4.3$ AP@$0.7$ gain. We note that larger jitter angles, such as \(\pm8^\circ\), can cause the point cloud to exceed the pre-defined detection range, limiting their practical utility.

These results indicate that the optimal jitter setting is task-specific and likely depends on the intrinsic sensor placement and motion characteristics of the platform. We believe that while the current settings are effective for the Pi3DET benchmark, other platforms may require tailored augmentation parameters. Moreover, our findings highlight a broader challenge: truly robust 3D detectors should ideally be invariant to viewpoint changes, yet current state-of-the-art models, due to their reliance on regularized point cloud representations, often lose genuine viewpoint robustness from the outset. Future research should continue to explore methods that overcome these limitations.

\subsubsection{Ablation Results on Cross-Dataset Task}
Table~\ref{tab:nusc_v_pi3det_v_benchmark} summarizes our cross-dataset adaptation results for the nuScenes $\rightarrow$ Pi3DET (Vehicle) task. In this setting, we compare several state-of-the-art methods using two base detectors, PVRCNN \cite{shi2020pv} and VoxelRCNN \cite{deng2021voxel}, and report AP in both BEV and 3D at IoU thresholds of $0.70$ and $0.50$. The table reveals several key observations: the Source Only model, trained solely on the nuScenes dataset, suffers from a considerable performance drop when directly applied to the Pi3DET (Vehicle) target dataset, underscoring the significant domain shift. 

In contrast, adaptation methods such as ST3D and ST3D++ markedly improve performance by leveraging self-training strategies. Our proposed method, Pi3DET-Net, achieves the highest AP scores among the compared methods on both PVRCNN and VoxelRCNN settings. For instance, under the PV-RCNN configuration, Pi3DET-Net attains an AP of $64.29\%$ in BEV and $54.76\%$ in 3D (at an IoU of $0.70$), which is substantially higher than the other methods, and it significantly narrows the gap to the fully supervised target performance. Overall, our method closes a large portion of the performance gap between the Source Only baseline and the Oracle (fully supervised) model.

\subsubsection{Cross-Platform 3D Detection Benchmark}
\begin{table*}
\centering
\caption{
    \textbf{Cross-platform 3D detection benchmark}. We report the average precision (AP) in ``BEV / 3D'' at the IoU thresholds of $0.7$. All scores are given in percentage (\%). "-C", "-A" mean detectors with Anchor-based or Center-based detection head.
}
\vspace{-0.2cm}
\resizebox{0.96\linewidth}{!}{
\begin{tabular}{c|l|c|c|c|c} 
    \toprule
    \multirow{2}{*}{\textbf{Category}} & \multirow{2}{*}{\textbf{Method}} & Vehicle & Quadruped & Drone & \multirow{2}{*}{\textbf{Average}}
    \\
    && \textbf{AP@0.50} & \textbf{AP@0.50} & \textbf{AP@0.50} & 
    \\ 
    \midrule\midrule
    \multirow{9}{*}{Grid-Based Detector} & PointPillar \cite{lang2019pointpillars} & $61.39$ / $59.86$& $46.81$ / $37.46$& $56.13$ / $49.03$& $54.78$ / $48.78$
    \\
    & SECOND-IOU \cite{yan2018second} & $62.95$ / $60.31$& $54.63$ / $44.31$& $60.02$ / $56.43$& $59.20$ / $53.68$
    \\
    & CenterPoint \cite{yin2021center} & $62.48$ / $60.91$& $52.79$ / $40.88$& $60.90$ / $53.38$& $58.72$ / $51.72$
    \\
    & PillarNet \cite{shi2022pillarnet} & $60.12$ / $58.57$& $46.88$ / $36.36$& $53.82$ / $46.29$& $53.61$ / $47.07$
    \\
    & Part A$^*$ \cite{shi2020points} & \textcolor{pi3det_red}{\hlred{$64.41$}} / \textcolor{pi3det_red}{\hlred{$63.10$}}& \textcolor{pi3det_blue}{\hlblue{$56.07$}} / \textcolor{pi3det_blue}{\hlblue{$46.89$}}& \textcolor{pi3det_blue}{\hlblue{$65.24$}} / \textcolor{pi3det_blue}{\hlblue{$57.37$}}& \textcolor{pi3det_blue}{\hlblue{$61.91$}} / \textcolor{pi3det_blue}{\hlblue{$55.79$}}
    \\
    & Transfusion-L \cite{bai2022transfusion} & $59.28$ / $56.77$& $52.41$ / $38.41$& $59.74$ / $48.63$& $57.14$ / $47.94$
    \\
    & HEDNet \cite{zhang2023hednet} & $57.14$ / $54.38$& $50.56$ / $35.77$& $58.05$ / $46.52$& $55.25$ / $45.56$
    \\
    & SAFNet \cite{kong2024safnet} & $53.01$ / $50.48$& $48.95$ / $36.68$& $59.80$ / $48.77$& $54.00$ / $45.31$
    \\
    &\textbf{Part A$^*$ + Ours}& \textcolor{pi3det_blue}{\hlblue{$63.21$}} / \textcolor{pi3det_blue}{\hlblue{$61.47$}}& \textcolor{pi3det_red}{\hlred{$57.26$}} / \textcolor{pi3det_red}{\hlred{$49.16$}}& \textcolor{pi3det_red}{\hlred{$67.82$}} / \textcolor{pi3det_red}{\hlred{$60.01$}}& \textcolor{pi3det_red}{\hlred{$62.76$}} / \textcolor{pi3det_red}{\hlred{$56.88$}}
    \\
    \midrule
    \multirow{5}{*}{Point-Based Detector} & PointRCNN \cite{shi2019pointrcnn} & $51.71$ / $51.04$& $48.45$ / $41.50$& $59.10$ / $52.31$& $53.09$ / $48.28$
    \\
    & 3DSSD \cite{yang20203dssd} & $52.72$ / $51.98$& $52.68$ / $43.07$& $62.32$ / $54.63$& $55.91$ / $49.89$
    \\
    & IA-SSD \cite{zhang2022not} & \textcolor{pi3det_red}{\hlred{$58.62$}} / \textcolor{pi3det_red}{\hlred{$57.61$}}& \textcolor{pi3det_red}{\hlred{$68.77$}} / \textcolor{pi3det_red}{\hlred{$56.65$}}& \textcolor{pi3det_red}{\hlred{$69.50$}} / \textcolor{pi3det_red}{\hlred{$60.10$}}& \textcolor{pi3det_red}{\hlred{$65.63$}} / \textcolor{pi3det_red}{\hlred{$58.12$}}
    \\
    & DBQ-SSD \cite{yang2022dbq} & $54.28$ / $53.87$& \textcolor{pi3det_blue}{\hlblue{$62.89$}} / \textcolor{pi3det_blue}{\hlblue{$54.77$}}& \textcolor{pi3det_blue}{\hlblue{$65.63$}} / $58.74$& \textcolor{pi3det_blue}{\hlblue{$60.93$}} / \textcolor{pi3det_blue}{\hlblue{$55.79$}}
    \\
    &\textbf{PointRCNN + Ours}& \textcolor{pi3det_blue}{\hlblue{$57.80$}} / \textcolor{pi3det_blue}{\hlblue{$57.23$}}& $49.76$ / $45.83$& $62.53$ / \textcolor{pi3det_blue}{\hlblue{$59.25$}}& $56.70$ / $54.10$
    \\
    \midrule
    \multirow{7}{*}{Grid-Point Detector} & PV-RCNN \cite{shi2020pv} & $67.02$ / $66.57$& $56.37$ / \textcolor{pi3det_blue}{\hlblue{$57.64$}}& $67.19$ / $59.66$& $63.53$ / \textcolor{pi3det_blue}{\hlblue{$61.29$}}
    \\
    & PV-RCNN-C \cite{shi2020pv} & $60.24$ / $60.08$& $51.58$ / $42.12$& $53.77$ / $52.66$& $55.20$ / $51.62$
    \\
    & PV-RCNN++-A \cite{shi2023pv} & \textcolor{pi3det_blue}{\hlblue{$67.59$}} / \textcolor{pi3det_red}{\hlred{$67.20$}}& \textcolor{pi3det_blue}{\hlblue{$57.91$}} / $47.95$& \textcolor{pi3det_blue}{\hlblue{$67.78$}} / \textcolor{pi3det_blue}{\hlblue{$60.14$}}& $64.43$ / $58.43$
    \\
    & PV-RCNN++-C \cite{shi2023pv} & $60.37$ / $60.20$& $51.45$ / $48.39$& $61.90$ / $53.12$& $57.91$ / $53.90$
    \\
    & VoxelRCNN-A \cite{deng2021voxel} & \textcolor{pi3det_red}{\hlred{$70.32$}} / \textcolor{pi3det_blue}{\hlblue{$66.27$}}& $57.31$ / $51.50$& $67.66$ / $59.62$& \textcolor{pi3det_blue}{\hlblue{$65.10$}} / $59.13$
    \\
    & VoxelRCNN \cite{deng2021voxel} & $60.21$ / $60.03$& $52.29$ / $49.04$& $61.91$ / $59.86$& $58.14$ / $56.31$
    \\
    &\textbf{PV-RCNN++ + Ours} & $66.33$ / $65.90$& \textcolor{pi3det_red}{\hlred{$68.15$}} / \textcolor{pi3det_red}{\hlred{$59.20$}}& \textcolor{pi3det_red}{\hlred{$70.47$}} / \textcolor{pi3det_red}{\hlred{$67.43$}}& \textcolor{pi3det_red}{\hlred{$68.32$}} / \textcolor{pi3det_red}{\hlred{$64.18$}}
    \\
    \bottomrule
\end{tabular}}
\label{tab:detectors_ap05}
\end{table*}

In this section, we detail our cross-platform detection benchmark built on the Pi3DET dataset and analyze the performance of several state-of-the-art 3D detection algorithms under the AP@$0.5$ metric. We evaluate detectors from three design paradigms -- point-based, grid-based, and point-grid-based -- to comprehensively assess their cross-platform performance.

\noindent\textbf{Dataset Settings.} 
For our experiments, we select the \texttt{penno big loop} sequence from the Vehicle platform as the training set, which contains a large number of Vehicle targets to ensure robust feature learning. The test set comprises three platforms: the Vehicle platform uses the \texttt{city hall} sequence, the Quadruped platform uses the \texttt{penno short loop} sequence, and the Drone platform uses the \texttt{penno parking 1} and \texttt{penno parking 2} sequences. These test sequences were collected in scenes similar to those in the training set to provide a fair evaluation of cross-platform detection performance.

\noindent\textbf{Implementation Details.} 
Our training framework is also built upon the open-source OpenPCDet project\footnote{\url{https://github.com/open-mmlab/OpenPCDet}} and is executed using two NVIDIA Titan RTX GPUs. For training, 3D detectors are optimized with a batch size of $4$ per GPU over $40$ epochs, using the Adam optimizer with an initial learning rate of $0.01$, weight decay of $0.001$, and momentum of $0.9$. The data augmentation strategy remains consistent with that used for both cross-platform and cross-dataset tasks. In our experiments, we selected the best-performing detector for each category and further enhanced its performance by incorporating Random Platform Jitter. Specifically, we set the rotation range for the Quadruped platform to \(\pm5^\circ\) and for the Drone platform to \(\pm3^\circ\).

Table~\ref{tab:detectors_ap05} presents the AP@$0.5$ results for various detectors. Our analysis yields several key findings: 
\begin{itemize}
    \item Under the AP@$0.5$ setting, all detectors show improved performance on the Quadruped and Drone platforms, sometimes approaching or even surpassing the results obtained on the Vehicle platform. This indicates that while these detectors have good recall, they still struggle to accurately regress the geometric parameters of the target bounding boxes.

    \item Detectors that combine grid-based and point-based representations continue to perform well under the AP@$0.5$ metric, suggesting that the hybrid approach of leveraging both regular (grid) and irregular (point cloud) representations is a highly effective strategy for building high-performance 3D detectors.

    \item Point-based detectors exhibit relatively balanced performance across platforms, with some even achieving higher AP@$0.5$ scores on Quadruped and Drone platforms than on the Vehicle platform. For example, IA-SSD achieves an AP@$0.5$ on the Drone platform that is approximately $2.5\%$ higher than on the Vehicle platform, indicating that architectures based on raw point cloud inputs tend to be less sensitive to viewpoint changes.

    \item Although IA-SSD shows significantly lower AP@$0.7$ performance compared to PointRCNN on the Vehicle platform, its AP@$0.5$ performance is notably higher -- especially on the Quadruped and Drone platforms. This suggests that the semantic feature extraction branch in IA-SSD plays a key role in overcoming viewpoint variations.

    \item We further evaluated the best-performing models across the different detector types by incorporating our proposed Random Platform Jitter (RPJ) data augmentation. Our experiments indicate that RPJ, while causing a slight decrease in performance on the Vehicle platform, significantly enhances cross-platform performance. Specifically, for the Part A$^*$ model, RPJ improved the average BEV/3D AP by $0.85\%$ and $1.1\%$, respectively; PointRCNN saw gains of $3.6\%$ and $5.8\%$, while PV-RCNN++ improved by $3.9\%$ and $5.8\%$.
\end{itemize} 
    
    These results demonstrate that although RPJ may slightly reduce performance on the source domain, it effectively boosts cross-platform detection performance by enhancing the model's robustness to diverse viewing conditions.

Overall, the experimental results under the AP@$0.5$ setting reveal that although current detectors exhibit strong recall, they often lack the precision needed to accurately regress bounding box geometries across different platforms. The combination of diverse detector architectures and the RPJ augmentation provides a promising pathway for improving cross-platform 3D detection, offering valuable insights for future research in this challenging domain.

\subsection{Additional Qualitative Results}
\label{sec:supp_c.2}
\begin{figure*}
    \centering
    \includegraphics[width=0.9\linewidth]{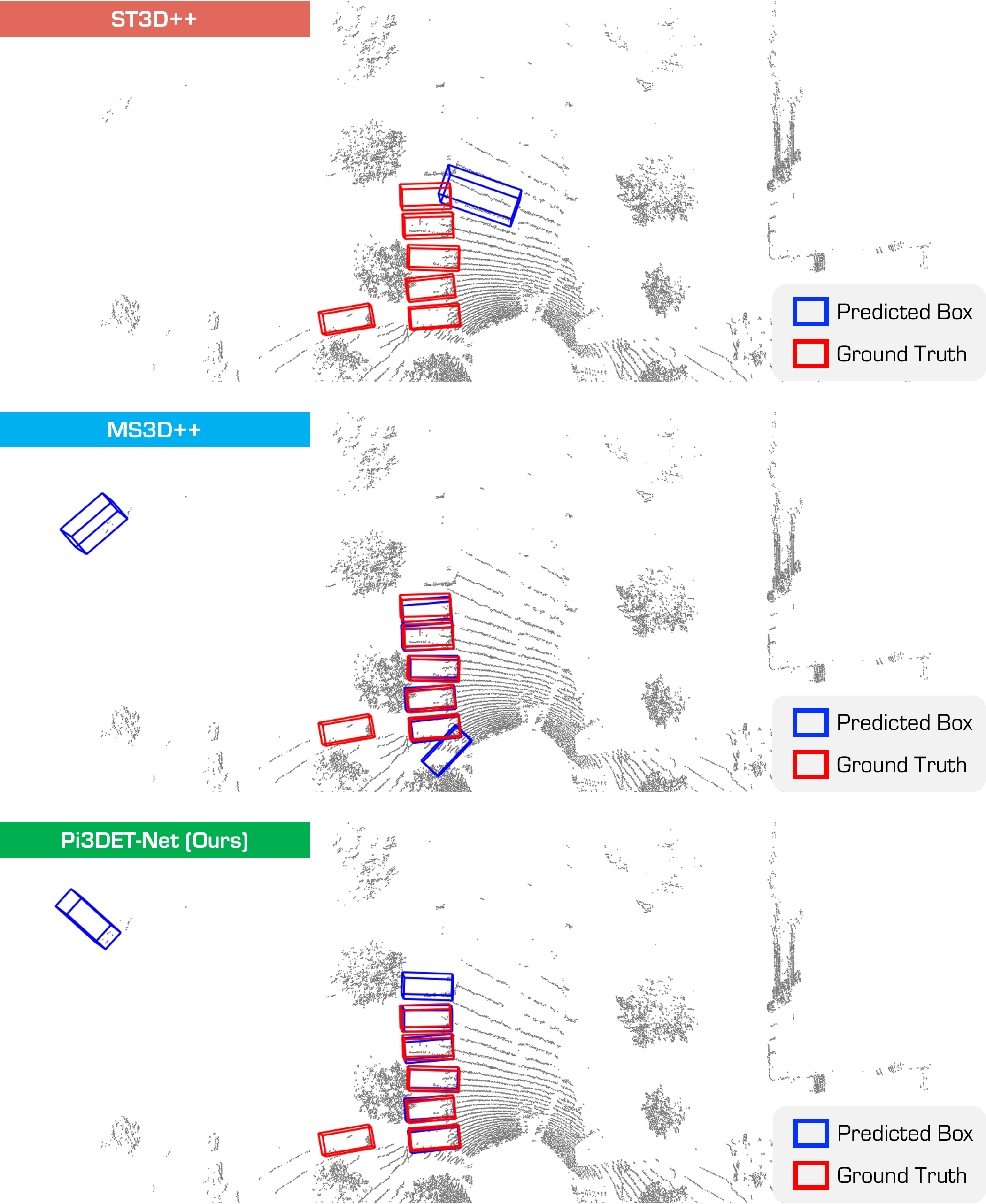}
    \caption{Qualitative results from state-of-the-art methods. We compare \textbf{Pi3DET-Net} with ST3D++ \cite{yang2022st3d++} and MS3D++ \cite{tsai2024ms3d++}. The figure illustrates predictions from methods that are adapted from \textbf{Pi3DET (Vehicle)} to \textbf{Pi3DET (Drone)}. Best viewed in colors.}
    \label{fig:visualization_1}
\end{figure*}

\begin{figure*}
    \centering
    \includegraphics[width=0.9\linewidth]{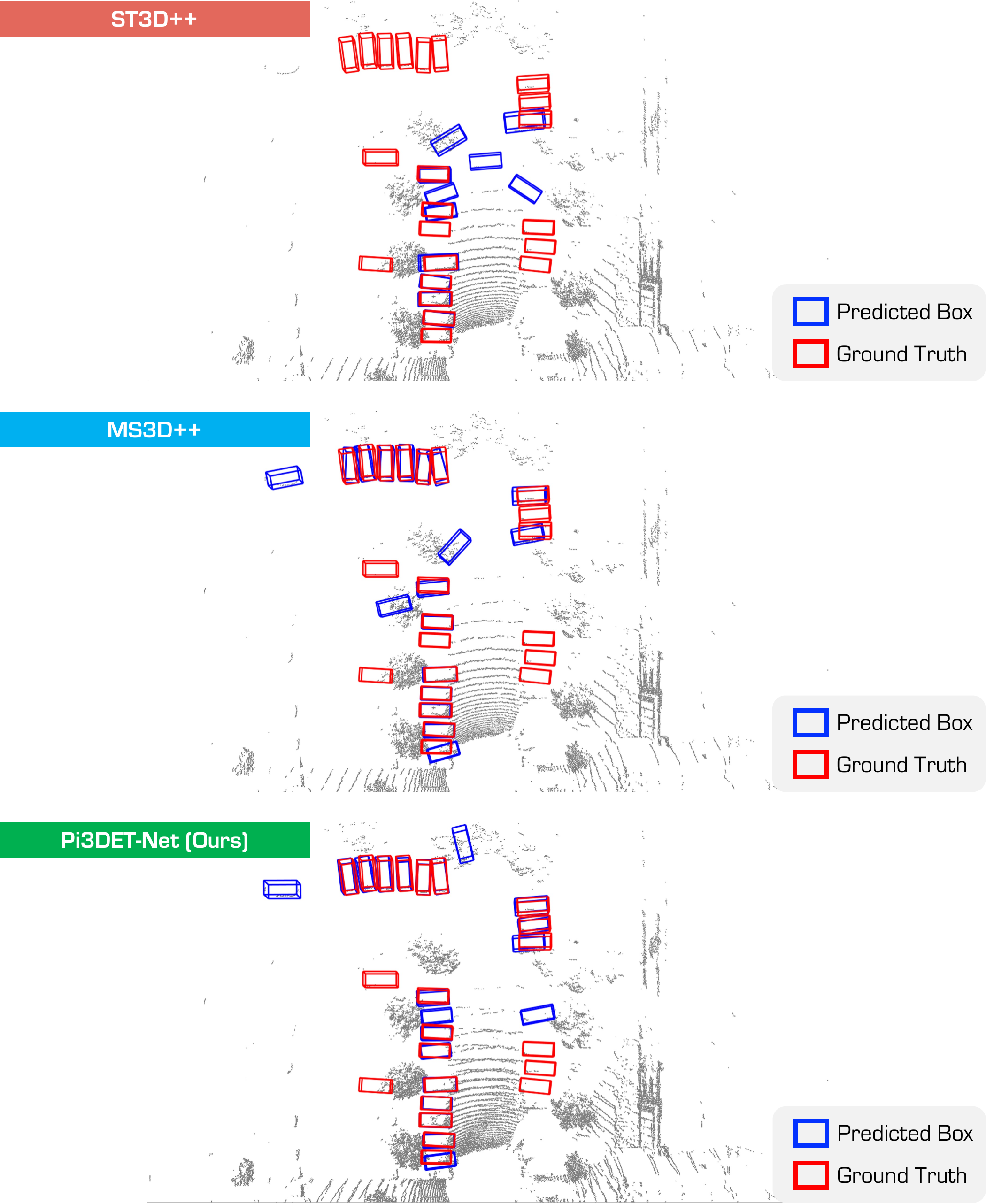}
    \caption{Qualitative results from state-of-the-art methods. We compare \textbf{Pi3DET-Net} with ST3D++ \cite{yang2022st3d++} and MS3D++ \cite{tsai2024ms3d++}. The figure illustrates predictions from methods that are adapted from \textbf{Pi3DET (Vehicle)} to \textbf{Pi3DET (Drone)}. Best viewed in colors.}
    \label{fig:visualization_2}
\end{figure*}

\begin{figure*}
    \centering
    \includegraphics[width=0.9\linewidth]{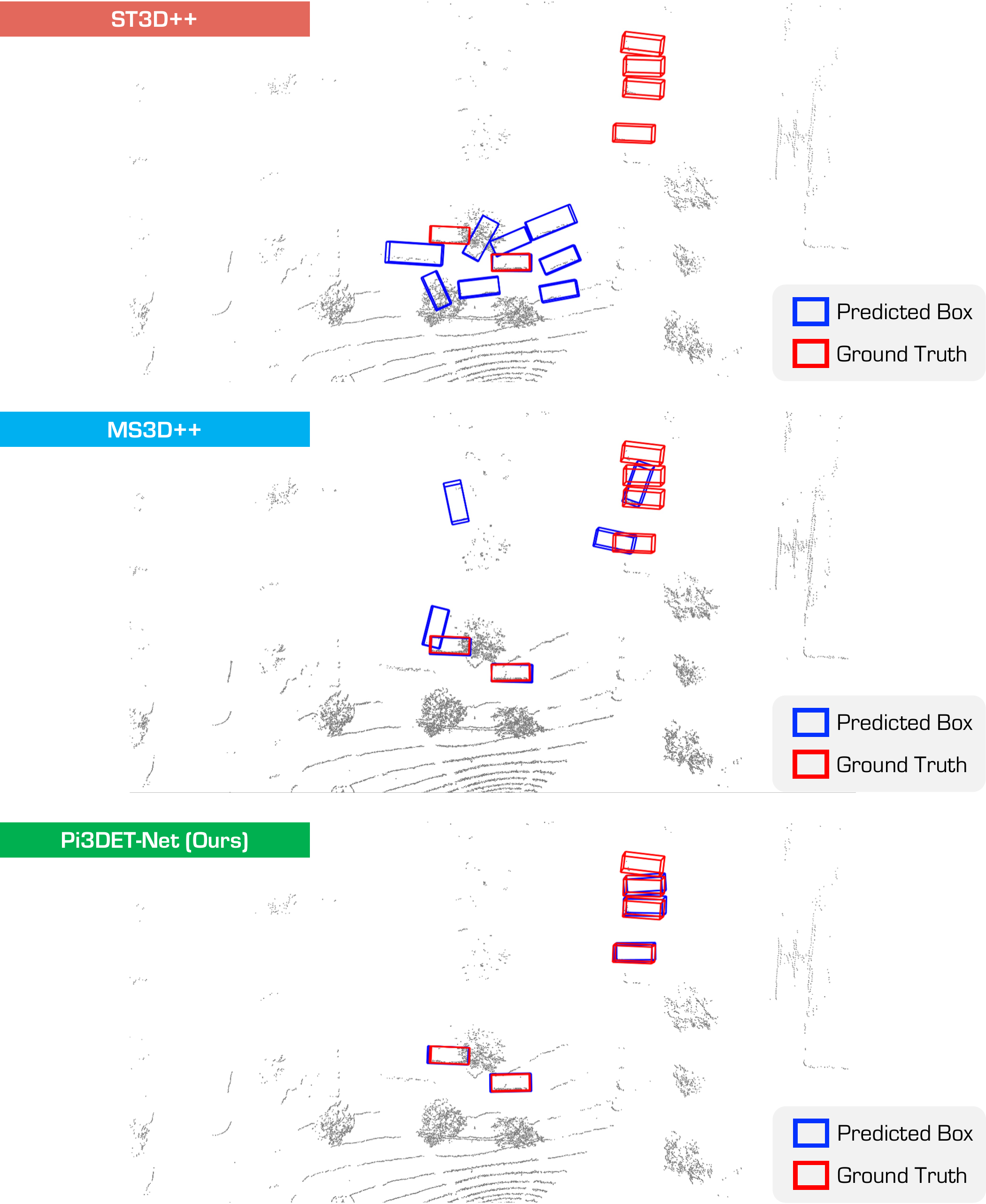}
    \caption{Qualitative results from state-of-the-art methods. We compare \textbf{Pi3DET-Net} with ST3D++ \cite{yang2022st3d++} and MS3D++ \cite{tsai2024ms3d++}. The figure illustrates predictions from methods that are adapted from \textbf{Pi3DET (Vehicle)} to \textbf{Pi3DET (Quadruped)}. Best viewed in colors.}
    \label{fig:visualization_3}
\end{figure*}

\begin{figure*}
    \centering
    \includegraphics[width=0.9\linewidth]{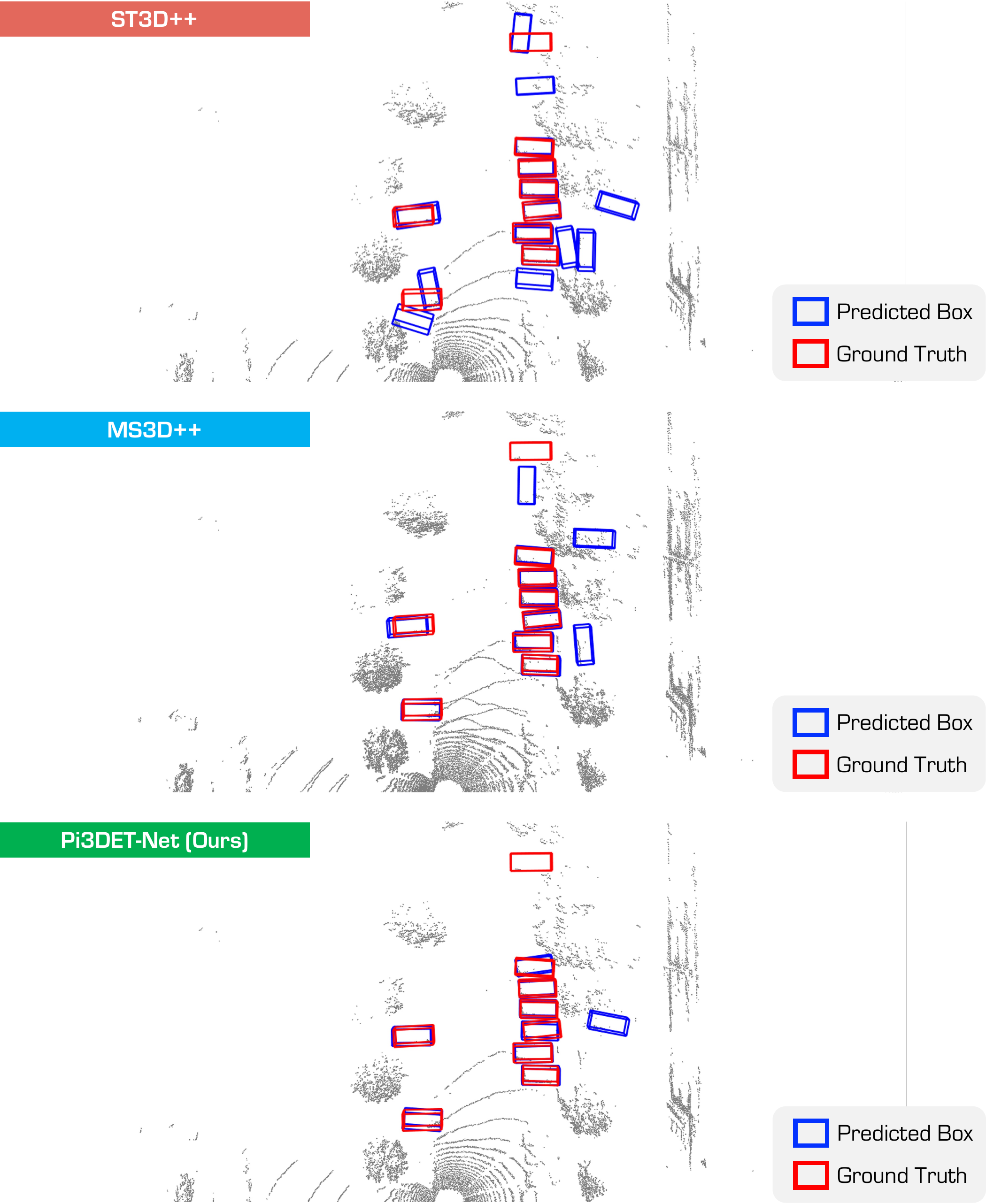}
    \caption{Qualitative results from state-of-the-art methods. We compare \textbf{Pi3DET-Net} with ST3D++ \cite{yang2022st3d++} and MS3D++ \cite{tsai2024ms3d++}. The figure illustrates predictions from methods that are adapted from \textbf{Pi3DET (Vehicle)} to \textbf{Pi3DET (Quadruped)}. Best viewed in colors.}
    \label{fig:visualization_4}
\end{figure*}

\begin{figure*}
    \centering
    \includegraphics[width=0.9\linewidth]{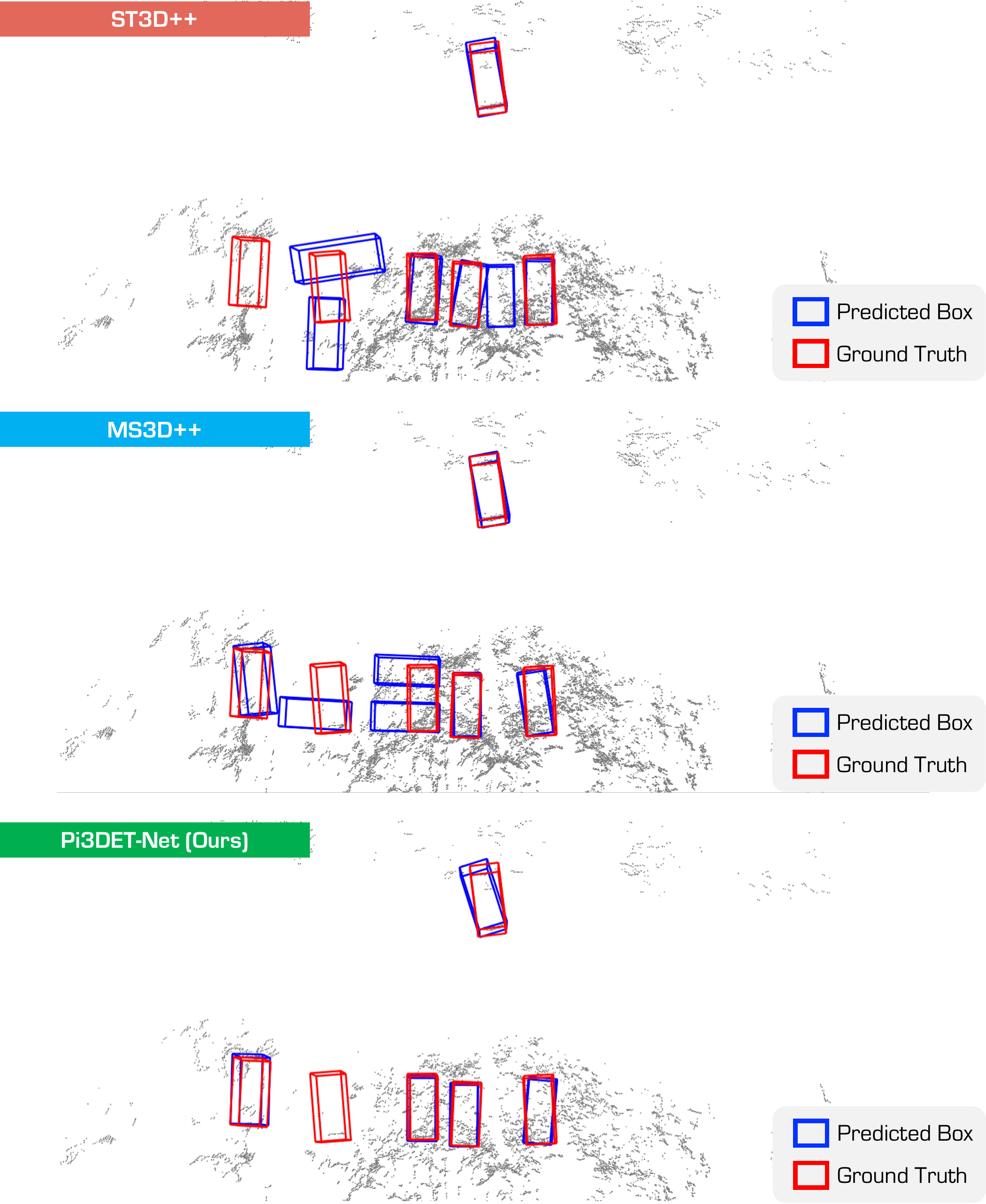}
    \caption{Qualitative results from state-of-the-art methods. We compare \textbf{Pi3DET-Net} with ST3D++ \cite{yang2022st3d++} and MS3D++ \cite{tsai2024ms3d++}. The figure illustrates predictions from methods that are adapted from \textbf{Pi3DET (Drone)} to \textbf{Pi3DET (Quadruped)}. Best viewed in colors.}
    \label{fig:visualization_5}
\end{figure*}

\begin{figure*}
    \centering
    \includegraphics[width=0.9\linewidth]{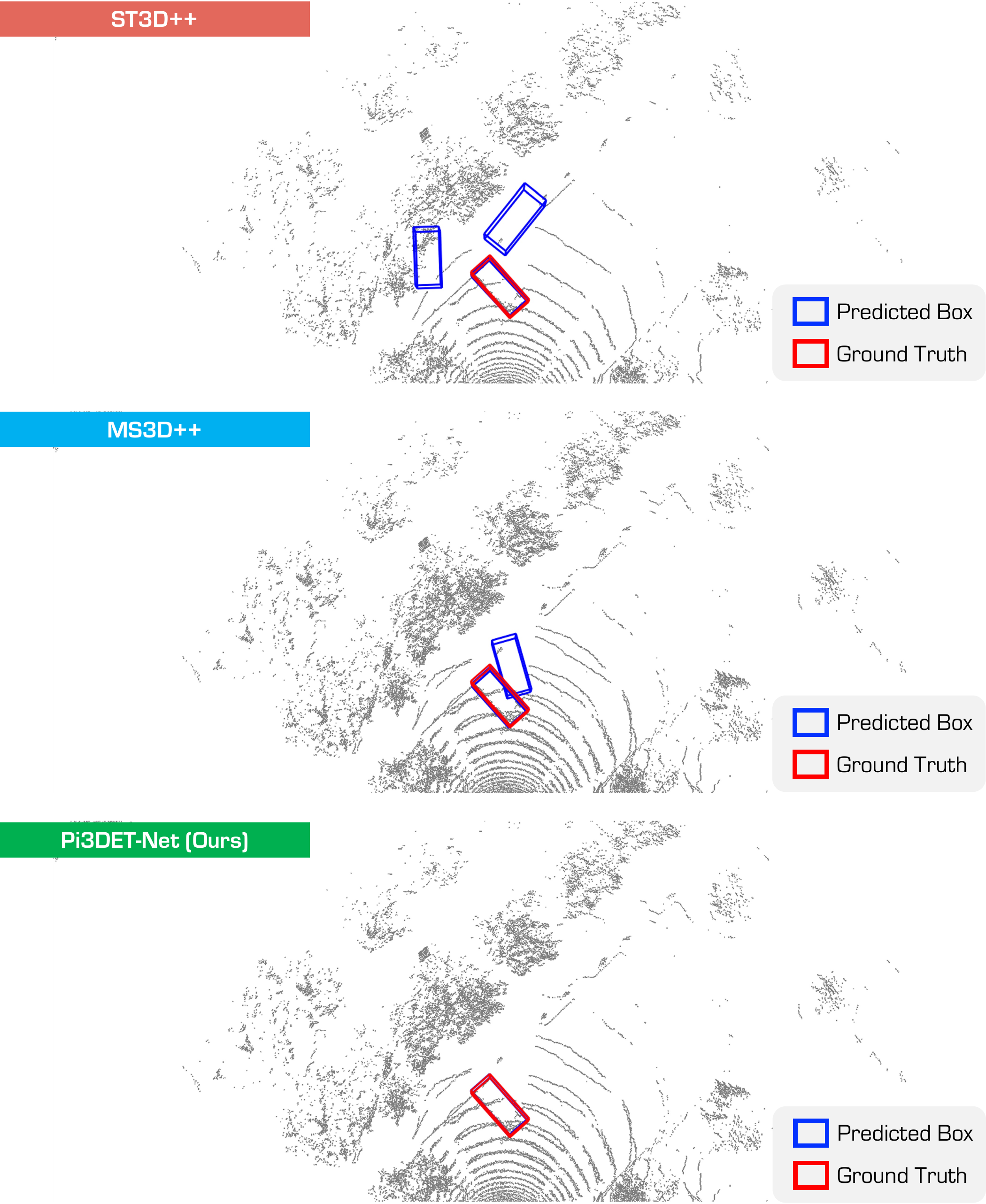}
    \caption{Qualitative results from state-of-the-art methods. We compare \textbf{Pi3DET-Net} with ST3D++ \cite{yang2022st3d++} and MS3D++ \cite{tsai2024ms3d++}. The figure illustrates predictions from methods that are adapted from \textbf{Pi3DET (Drone)} to \textbf{Pi3DET (Quadruped)}. Best viewed in colors.}
    \label{fig:visualization_6}
\end{figure*}

In this section, we present qualitative visualizations for six cross-platform adaptation tasks to further analyze the effectiveness of our proposed method, Pi3DET-Net (see Figure~\ref{fig:visualization_1} through Figure~\ref{fig:visualization_6}). 

We compare our results against two state-of-the-art cross-dataset approaches, ST3D++ \cite{yang2022st3d++} and MS3D++ \cite{tsai2024ms3d++}. Overall, Pi3DET-Net consistently delivers superior detection performance across all tasks. For example, in Figure~\ref{fig:visualization_1}, ST3D++ fails to detect a target in one scenario, whereas Pi3DET-Net successfully captures the target in its entirety; in contrast, MS3D++ tends to produce false positives. 

Similarly, Figure~\ref{fig:visualization_3} illustrates that while both ST3D++ and MS3D++ generate numerous false positives, our method maintains high precision and recall. These qualitative observations, combined with our quantitative analyses, highlight the significant advantages of Pi3DET-Net in cross-platform detection tasks.

\subsection{Failure Cases}
\label{sec:supp_c.3}
Although Pi3DET-Net introduces effective strategies to enhance viewpoint robustness in cross-platform detection tasks, certain failure cases reveal limitations and challenges that remain to be addressed. 

In some scenarios, when the platform viewpoint becomes excessively distorted, Pi3DET-Net tends to miss detections, as illustrated in Figure~\ref{fig:visualization_3}. This suggests that further improvements in aligning platform feature domains are necessary. Additionally, the method still struggles with long-distance detection; sparse targets at far ranges exhibit significant deviations in feature distribution under viewpoint transformations, leading to degraded performance. 

Furthermore, Pi3DET-Net does not achieve true viewpoint invariance; it fundamentally relies on the underlying performance of the base detector. Current state-of-the-art detectors typically depend on regularizing point clouds, which involves pre-defining a sensing range. When significant viewpoint changes occur, for example, a $10^\circ$ downward tilt can reduce the effective sensing range to under $20$ meters due to increased vertical drop in the point cloud (As illustrated in our example Figure~\ref{fig:dataset_example_3}.), resulting in fewer points being captured within the detection range. 

In future work, based on Pi3DET, we plan to develop more effective data augmentation strategies and leverage the intrinsic robustness of point-based approaches to design detectors that achieve true viewpoint invariance without relying on pre-defined sensing ranges.
\section{Broader Impact}
In this section, we discuss the broader impact of our proposed Pi3DET dataset and the Pi3DET-Net framework, highlighting its contributions to robot perception and beyond. Additionally, we outline potential limitations and areas for future improvements.

\subsection{Potential Societal Impact}

The Pi3DET dataset and Pi3DET-Net framework hold significant promise for advancing robotic perception and enhancing the safety and efficiency of autonomous systems. By providing a comprehensive benchmark for cross-platform 3D detection, our work can foster the development of detectors that perform robustly in diverse real-world environments. 

This progress is critical for a wide array of applications, from autonomous driving and delivery robotics to search and rescue operations, ultimately contributing to improved safety, reduced operational risks, and more efficient resource utilization. 

Moreover, the availability of a multi-platform dataset may accelerate innovation in related fields such as surveillance, environmental monitoring, and assistive technologies.

\subsection{Potential Limitations}
Despite the promising results, several limitations warrant consideration. First, the effectiveness of Pi3DET-Net is still largely dependent on the underlying performance of base detectors, which may constrain its applicability across various sensor types or operational conditions. Second, the current approach relies on predefined sensing ranges and data augmentation strategies (\emph{e.g.}, Random Platform Jitter), which may not generalize optimally to platforms with significantly different sensor configurations or motion patterns.

\subsection{Future Directions}
Looking ahead, we plan to further enhance cross-platform robustness by exploring novel data augmentation techniques that reduce dependency on fixed sensing ranges and better capture the dynamics of varying platform motions. 

In addition, future work will investigate more intrinsically viewpoint-invariant detection architectures, potentially leveraging advances in point-based feature extraction to overcome the limitations of regularized representations. We also aim to extend our framework to other modalities and domains, such as multi-modal sensor fusion detection, to further advance the state of autonomous perception. 

Ultimately, we hope that the Pi3DET dataset and our findings will serve as a foundation for developing truly platform-agnostic 3D detection systems.
\section{Public Resources Used}
\label{sec:public-resources-used}

In this section, we acknowledge the use of the following public resources, during the course of this work.

\subsection{Public Codebase Used}
We acknowledge the use of the following public codebase, during the course of this work:
\begin{itemize}
    \item MMEngine\footnote{\url{https://github.com/open-mmlab/mmengine}.} \dotfill Apache License 2.0
    
    \item MMCV\footnote{\url{https://github.com/open-mmlab/mmcv}.} \dotfill Apache License 2.0
    
    \item MMDetection\footnote{\url{https://github.com/open-mmlab/mmdetection}.} \dotfill Apache License 2.0
    
    \item MMDetection3D\footnote{\url{https://github.com/open-mmlab/mmdetection3d}.} \dotfill Apache License 2.0
    
    \item OpenPCSeg\footnote{\url{https://github.com/PJLab-ADG/OpenPCSeg}.} \dotfill Apache License 2.0
    \item OpenPCDet\footnote{\url{https://github.com/open-mmlab/OpenPCDet}.} \dotfill Apache License 2.0
    \item xtreme1\footnote{\url{https://github.com/xtreme1-io/xtreme1}.} \dotfill Apache License 2.0
\end{itemize}

\subsection{Public Datasets Used}
We acknowledge the use of the following public datasets, during the course of this work:
\begin{itemize}
    \item M3ED\footnote{\url{https://m3ed.io}.}\dotfill CC BY-SA 4.0

    \item nuScenes\footnote{\url{https://www.nuscenes.org/nuscenes}.} \dotfill CC BY-NC-SA 4.0
    
    \item KITTI\footnote{\url{http://www.cvlibs.net/datasets/kitti}.} \dotfill CC BY-NC-SA 3.0.
\end{itemize}

\subsection{Public Implementations Used}
\begin{itemize}
    \item nuscenes-devkit\footnote{\url{https://github.com/nutonomy/nuscenes-devkit}.} \dotfill Apache License 2.0
    
    \item waymo-open-dataset\footnote{\url{https://github.com/waymo-research/waymo-open-dataset}.} \dotfill Apache License 2.0

    \item Open3D\footnote{\url{http://www.open3d.org}.} \dotfill MIT License  
    \item PyTorch\footnote{\url{https://pytorch.org}.} \dotfill BSD License  
    \item ROS Humble\footnote{\url{https://docs.ros.org/en/humble}.} \dotfill Apache License 2.0
    
    \item TorchSparse\footnote{\url{https://github.com/mit-han-lab/torchsparse}.} \dotfill MIT License
\end{itemize}

\clearpage

\clearpage

\clearpage

\clearpage

\clearpage
\begin{table*}[t]
\centering
\caption{Summary of \textbf{point cloud distribution statistics} ($x$, $y$, $z$, and intensity) of the \includegraphics[width=0.0225\linewidth]{figures/icons/vehicle.png} \textbf{Vehicle} data from \textbf{Pi3DET}.}
\vspace{-0.2cm}
\resizebox{\linewidth}{!}{
\begin{tabular}{c|c|l|c}
    \toprule
    \textbf{Platform} & \textbf{Condition} & \textbf{Sequence} & \textbf{Point Cloud Distributions (X, Y, Z, Intensity)}
    \\\midrule\midrule
    \multirow{66}{*}{\makecell{\textcolor{pi3det_red}{~~\textbf{Vehicle}~~}\\\textcolor{pi3det_red}{$\mathbf{(8)}$}}} & \multirow{16}{*}{\makecell{Daytime\\$(4)$}} & city\_hall & \begin{minipage}[b]{1.05\columnwidth}\centering\raisebox{-.47\height}{\includegraphics[width=\linewidth]{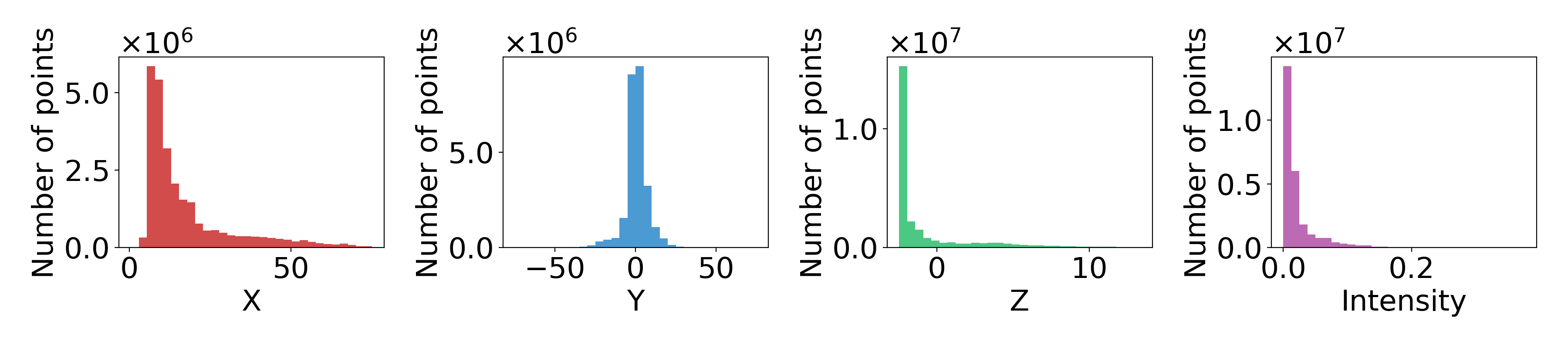}}\end{minipage}
    \\\cmidrule{3-4}
    & & penno\_big\_loop & \begin{minipage}[b]{1.05\columnwidth}\centering\raisebox{-.47\height}{\includegraphics[width=\linewidth]{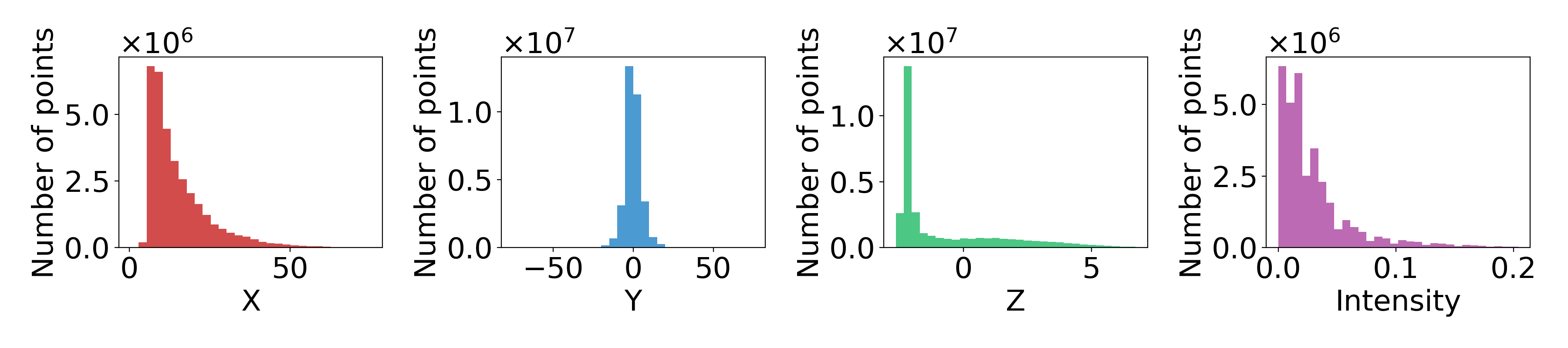}}\end{minipage}
    \\\cmidrule{3-4}
    & & rittenhouse & \begin{minipage}[b]{1.05\columnwidth}\centering\raisebox{-.47\height}{\includegraphics[width=\linewidth]{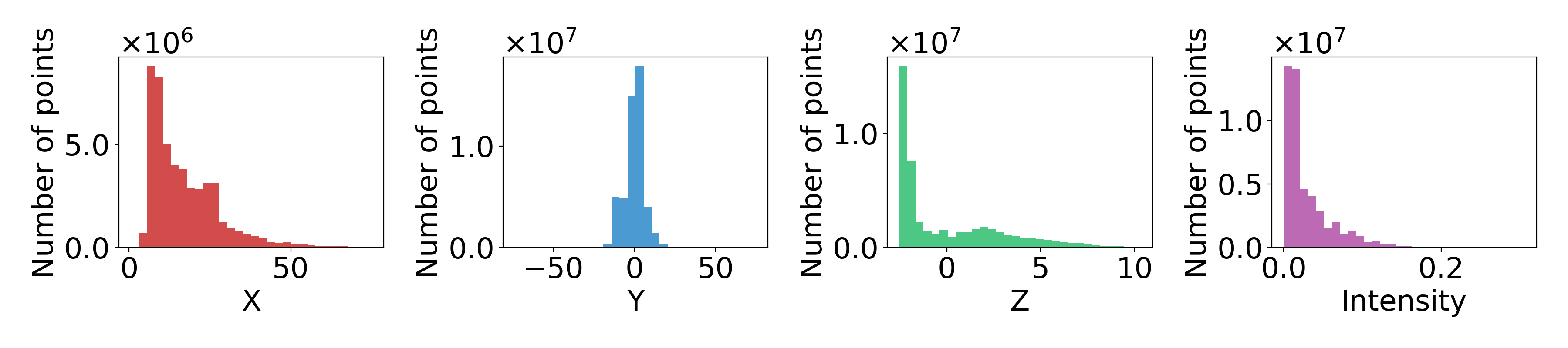}}\end{minipage}
    \\\cmidrule{3-4}
    & & ucity\_small\_loop & \begin{minipage}[b]{1.05\columnwidth}\centering\raisebox{-.47\height}{\includegraphics[width=\linewidth]{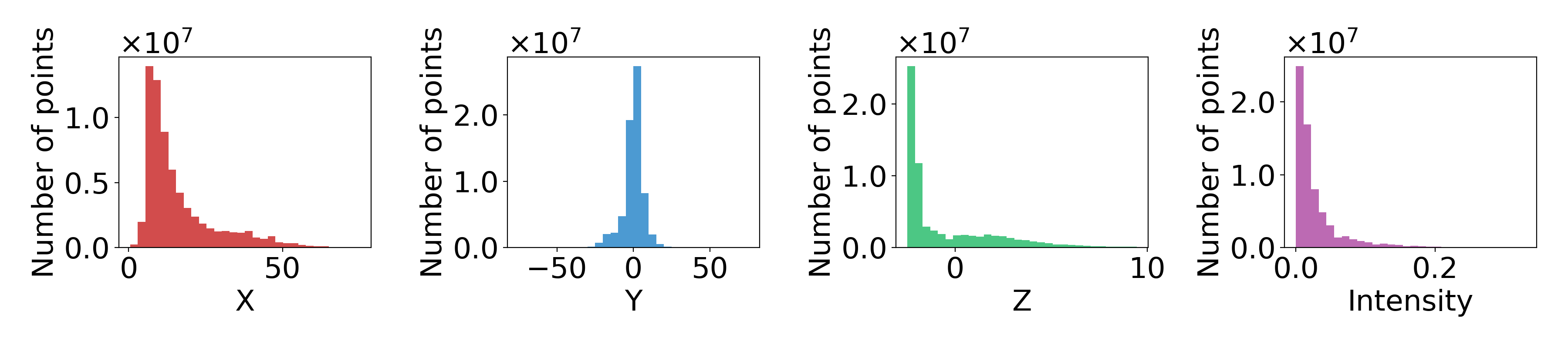}}\end{minipage}
    \\\cmidrule{2-4}
    & \multirow{16}{*}{\makecell{~~Nighttime~~\\$(4)$}} & city\_hall & \begin{minipage}[b]{1.05\columnwidth}\centering\raisebox{-.47\height}{\includegraphics[width=\linewidth]{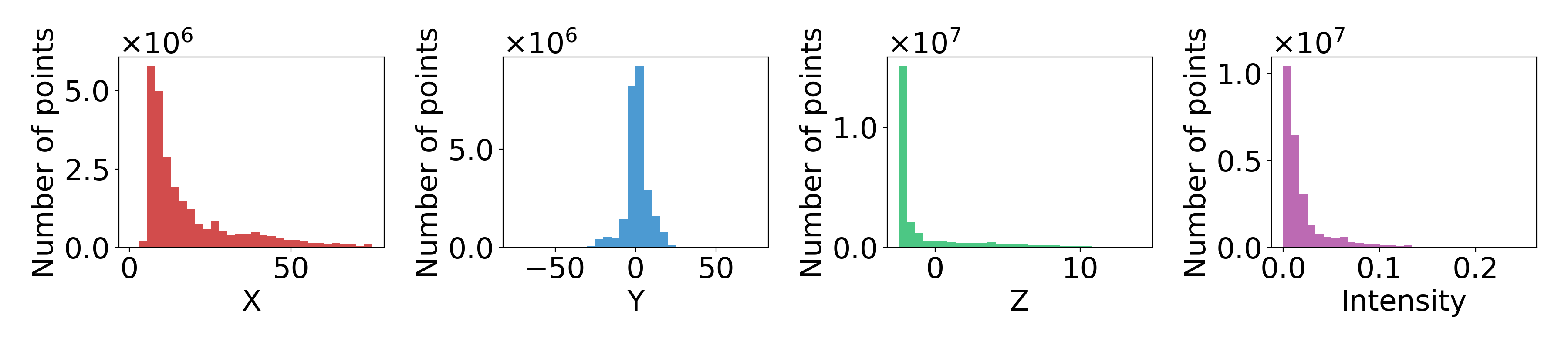}}\end{minipage}
    \\\cmidrule{3-4}
    & & penno\_big\_loop & \begin{minipage}[b]{1.05\columnwidth}\centering\raisebox{-.47\height}{\includegraphics[width=\linewidth]{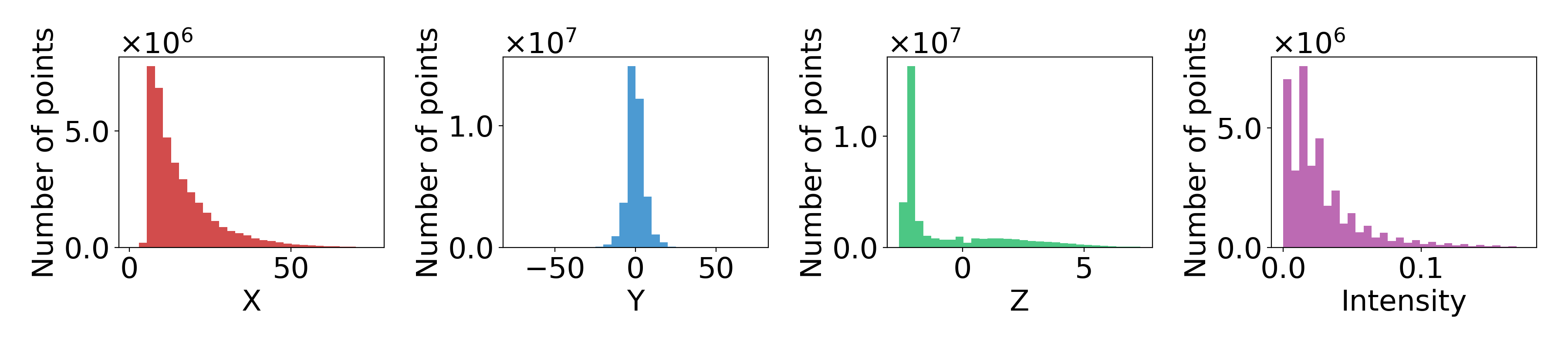}}\end{minipage}
    \\\cmidrule{3-4}
    & & rittenhouse & \begin{minipage}[b]{1.05\columnwidth}\centering\raisebox{-.47\height}{\includegraphics[width=\linewidth]{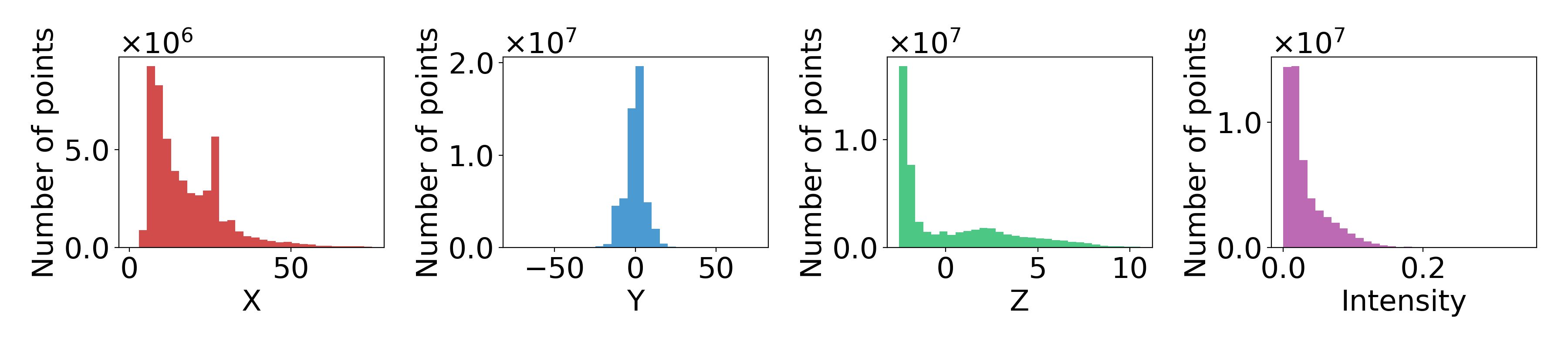}}\end{minipage}
    \\\cmidrule{3-4}
    & & ucity\_small\_loop~~~~ & \begin{minipage}[b]{1.05\columnwidth}\centering\raisebox{-.47\height}{\includegraphics[width=\linewidth]{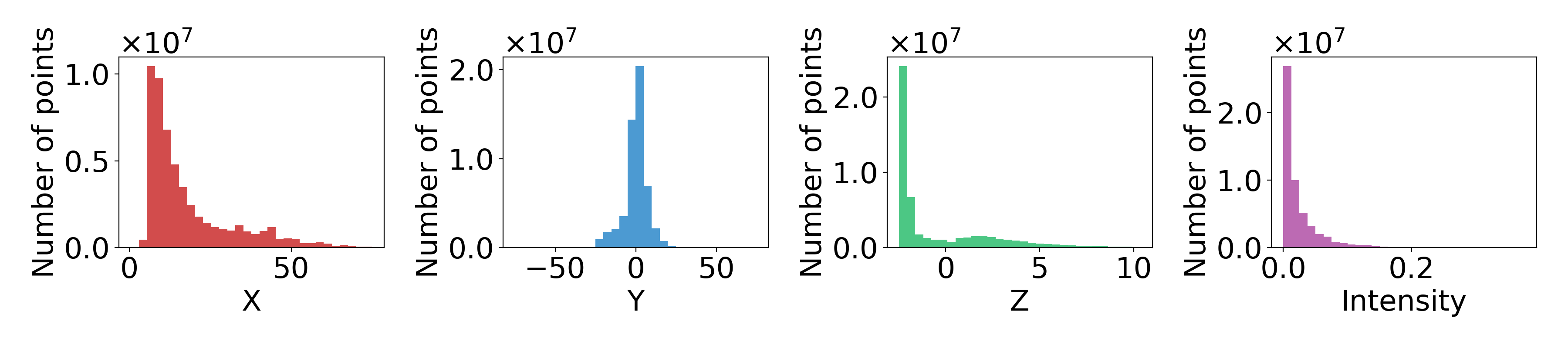}}\end{minipage}
    \\
    \bottomrule
\end{tabular}}
\label{tab:dataset_intensity_vehicle}
\end{table*}

\begin{table*}[t]
\centering
\caption{Summary of \textbf{point cloud distribution statistics} ($x$, $y$, $z$, and intensity) of the \includegraphics[width=0.023\linewidth]{figures/icons/drone.png} \textbf{Drone} data from \textbf{Pi3DET}.}
\vspace{-0.2cm}
\resizebox{\linewidth}{!}{
\begin{tabular}{c|c|l|c}
    \toprule
    \textbf{Platform} & \textbf{Condition} & \textbf{Sequence} & \textbf{Point Cloud Distributions (X, Y, Z, Intensity)}
    \\\midrule\midrule
    \multirow{56}{*}{\makecell{\textcolor{pi3det_blue}{~~\textbf{Drone}~~}\\\textcolor{pi3det_blue}{$\mathbf{(7)}$}}} & \multirow{16}{*}{\makecell{Daytime\\$(4)$}} & penno\_parking\_1 & \begin{minipage}[b]{1.05\columnwidth}\centering\raisebox{-.47\height}{\includegraphics[width=\linewidth]{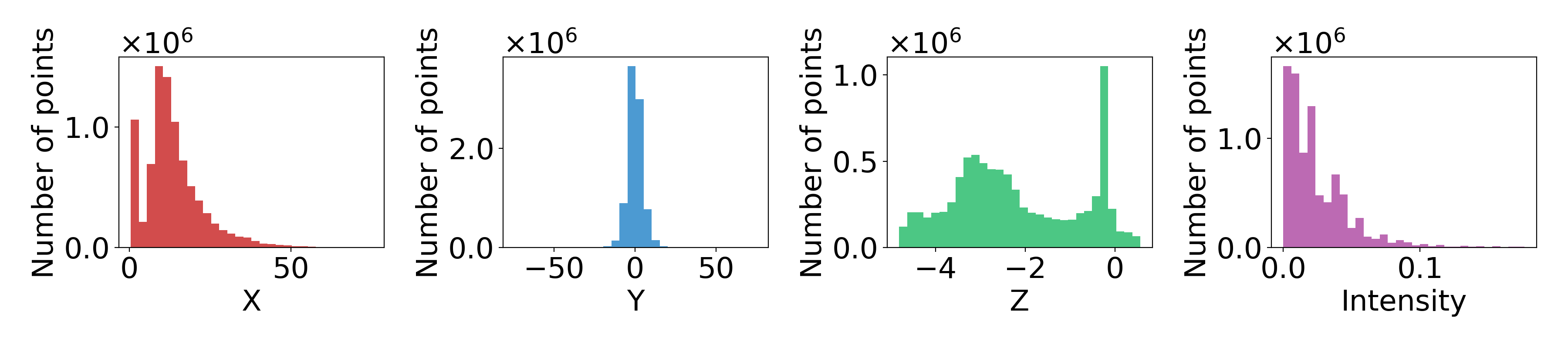}}\end{minipage}
    \\\cmidrule{3-4}
    & & penno\_parking\_2 & \begin{minipage}[b]{1.05\columnwidth}\centering\raisebox{-.47\height}{\includegraphics[width=\linewidth]{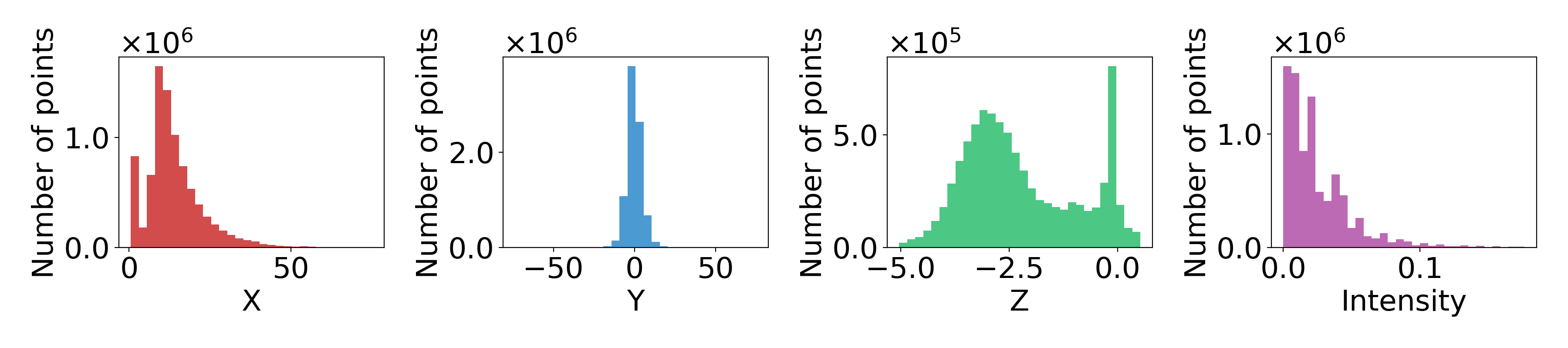}}\end{minipage}
    \\\cmidrule{3-4}
    & & penno\_plaza & \begin{minipage}[b]{1.05\columnwidth}\centering\raisebox{-.47\height}{\includegraphics[width=\linewidth]{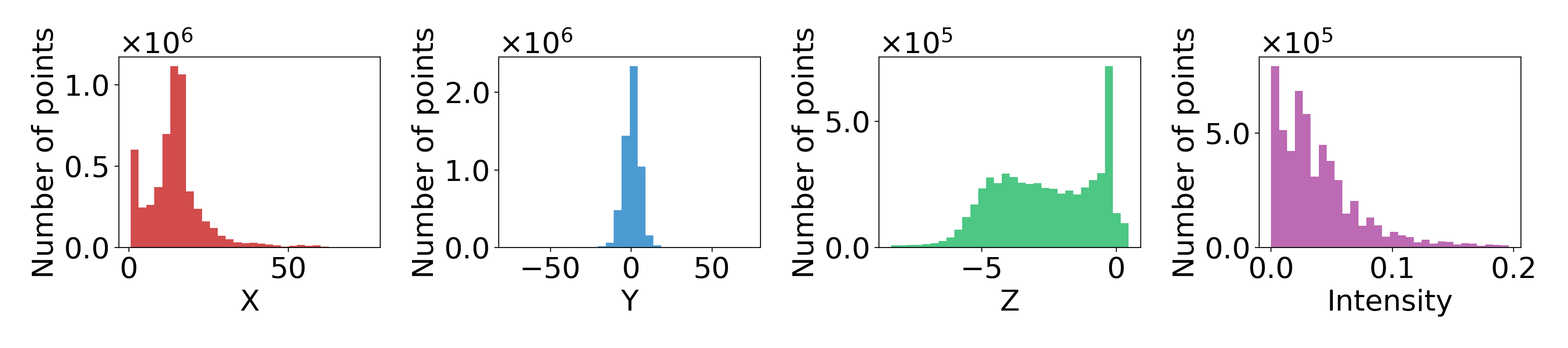}}\end{minipage}
    \\\cmidrule{3-4}
    & & penno\_trees & \begin{minipage}[b]{1.05\columnwidth}\centering\raisebox{-.47\height}{\includegraphics[width=\linewidth]{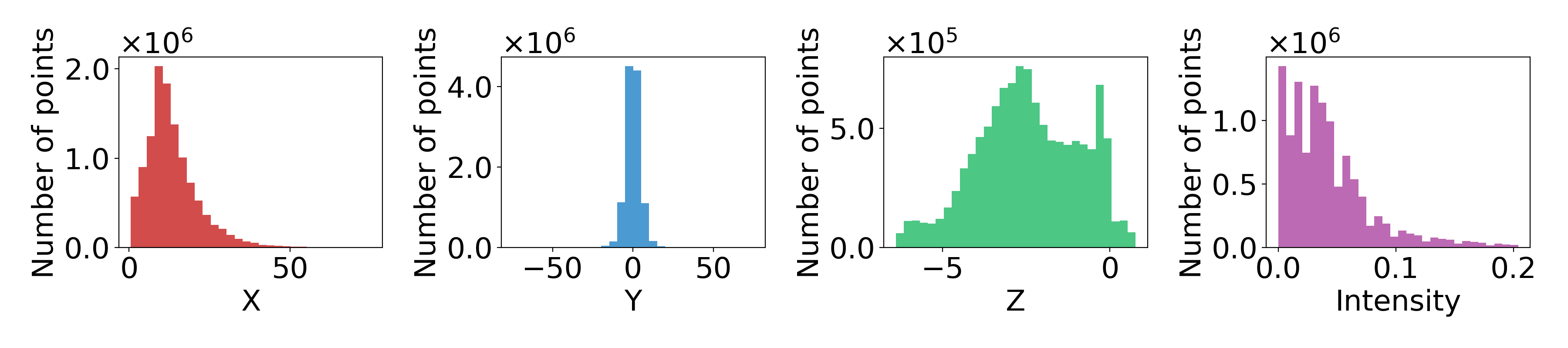}}\end{minipage}
    \\\cmidrule{2-4}
    & \multirow{12}{*}{\makecell{~~Nighttime~~\\$(3)$}} & high\_beams & \begin{minipage}[b]{1.05\columnwidth}\centering\raisebox{-.47\height}{\includegraphics[width=\linewidth]{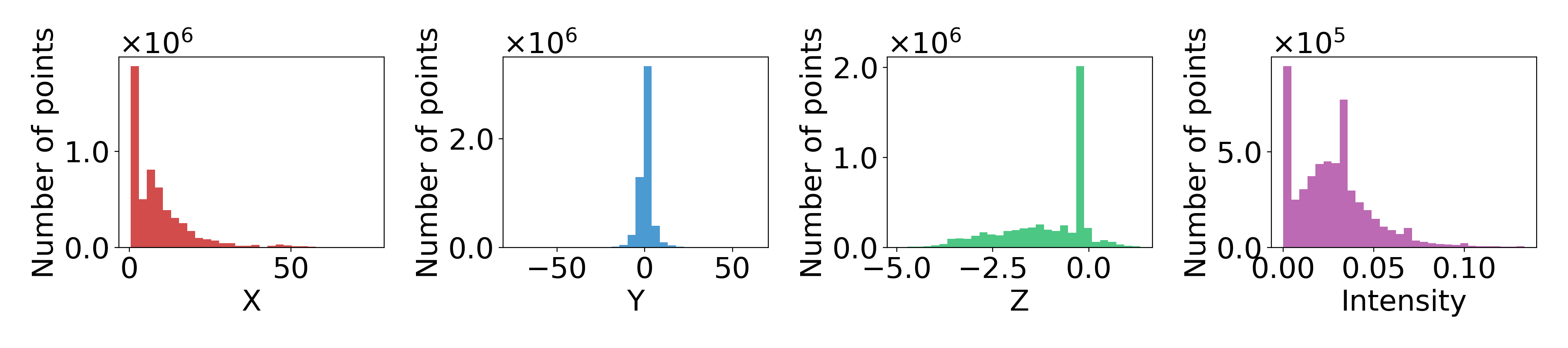}}\end{minipage}
    \\\cmidrule{3-4}
    & & penno\_parking\_1 & \begin{minipage}[b]{1.05\columnwidth}\centering\raisebox{-.47\height}{\includegraphics[width=\linewidth]{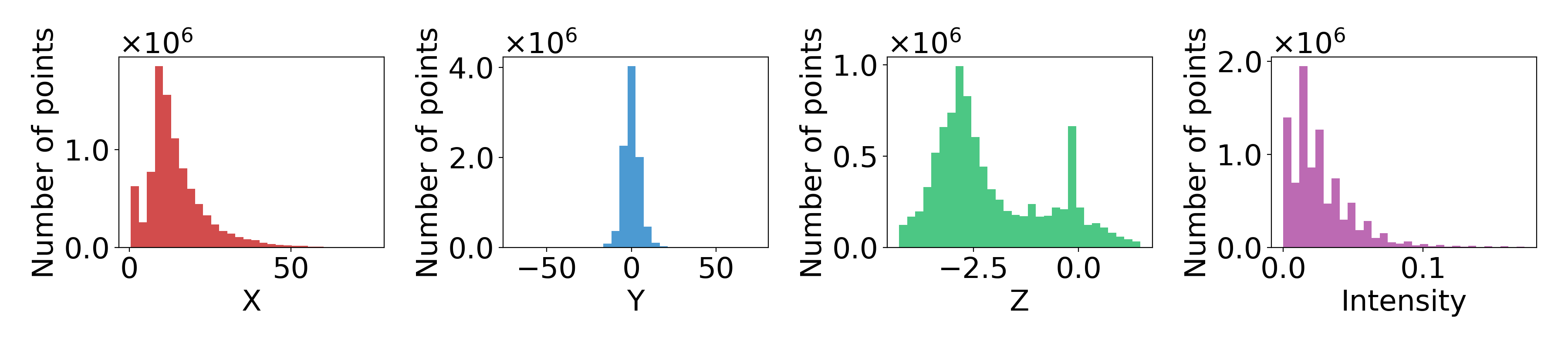}}\end{minipage}
    \\\cmidrule{3-4}
    & & penno\_parking\_2~~~~~ & \begin{minipage}[b]{1.05\columnwidth}\centering\raisebox{-.47\height}{\includegraphics[width=\linewidth]{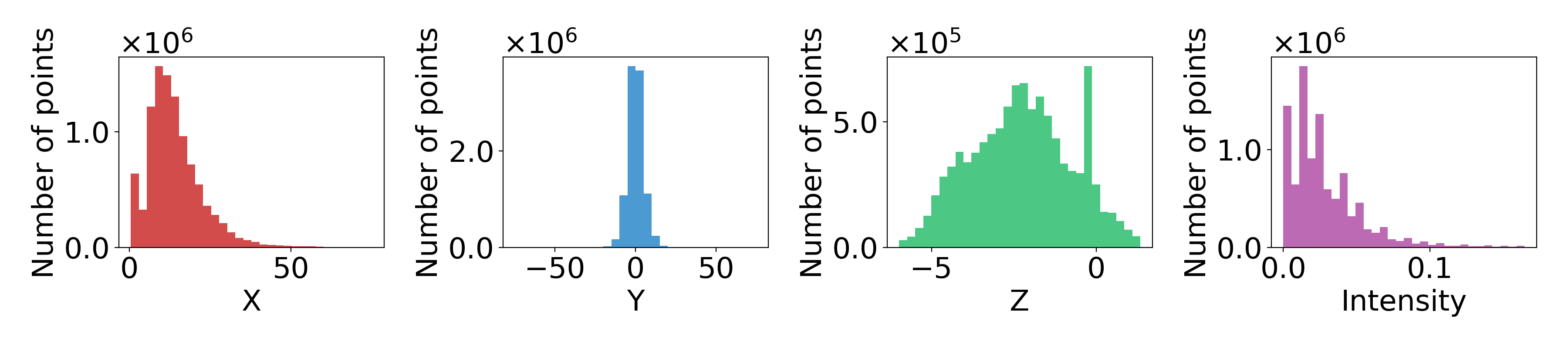}}\end{minipage}
    \\
    \bottomrule
\end{tabular}}
\label{tab:dataset_intensity_drone}
\end{table*}

\begin{table*}[t]
\centering
\caption{Summary of \textbf{point cloud distribution statistics} ($x$, $y$, $z$, and intensity) of the \includegraphics[width=0.023\linewidth]{figures/icons/quadruped.png} \textbf{Quadruped} data from \textbf{Pi3DET}.}
\vspace{-0.2cm}
\resizebox{.8\linewidth}{!}{
\begin{tabular}{c|c|l|c}
    \toprule
    \textbf{Platform} & \textbf{Condition} & \textbf{Sequence} & \textbf{Point Cloud Distributions (X, Y, Z, Intensity)}
    \\\midrule\midrule
    \multirow{85}{*}{\makecell{\textcolor{pi3det_green}{\textbf{Quadruped}}\\\textcolor{pi3det_green}{$\mathbf{(10)}$}}} & \multirow{36}{*}{\makecell{Daytime\\$(8)$}} & art\_plaza\_loop & \begin{minipage}[b]{1.05\columnwidth}\centering\raisebox{-.47\height}{\includegraphics[width=\linewidth]{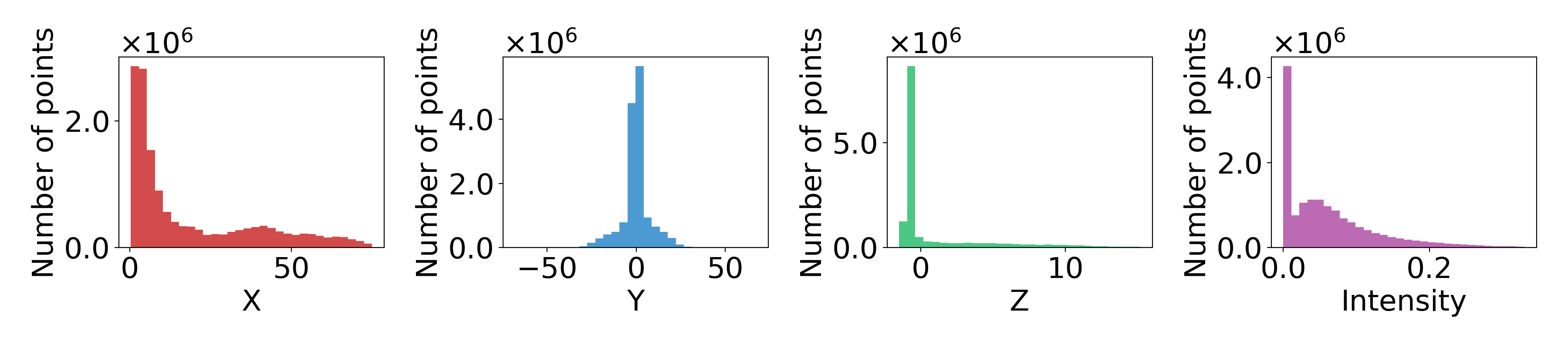}}\end{minipage}
    \\\cmidrule{3-4}
    & & penno\_short\_loop & \begin{minipage}[b]{1.05\columnwidth}\centering\raisebox{-.47\height}{\includegraphics[width=\linewidth]{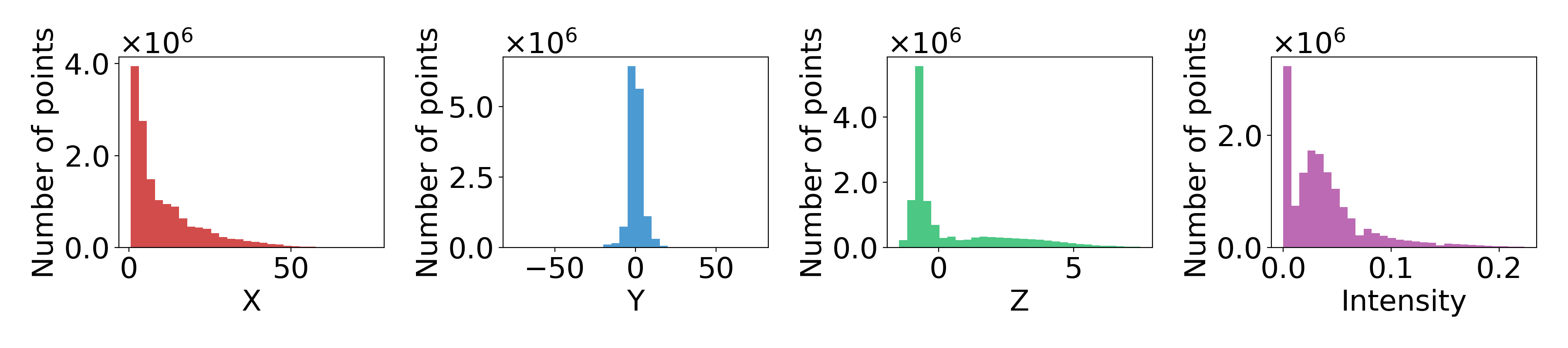}}\end{minipage}
    \\\cmidrule{3-4}
    & & rocky\_steps & \begin{minipage}[b]{1.05\columnwidth}\centering\raisebox{-.47\height}{\includegraphics[width=\linewidth]{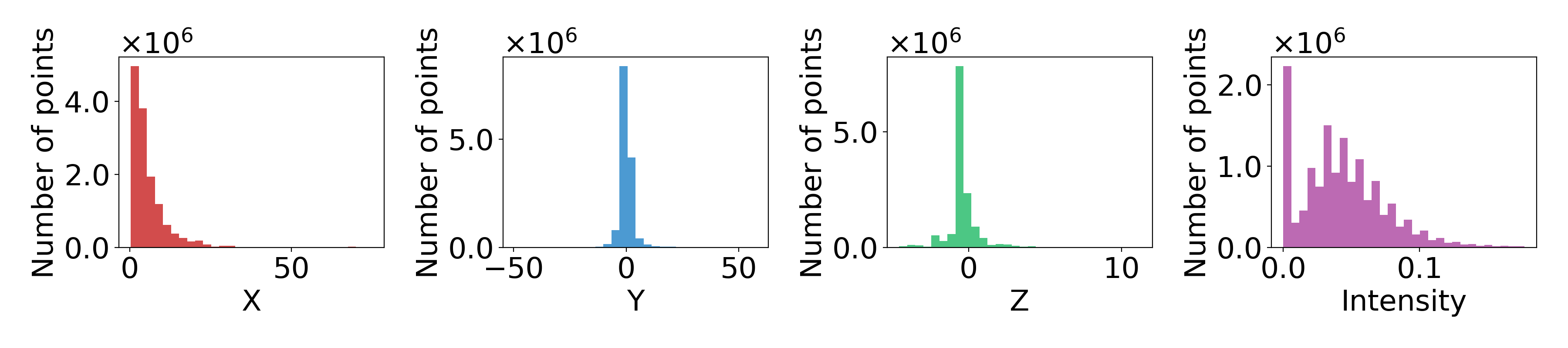}}\end{minipage}
    \\\cmidrule{3-4}
    & & skatepark\_1 & \begin{minipage}[b]{1.05\columnwidth}\centering\raisebox{-.47\height}{\includegraphics[width=\linewidth]{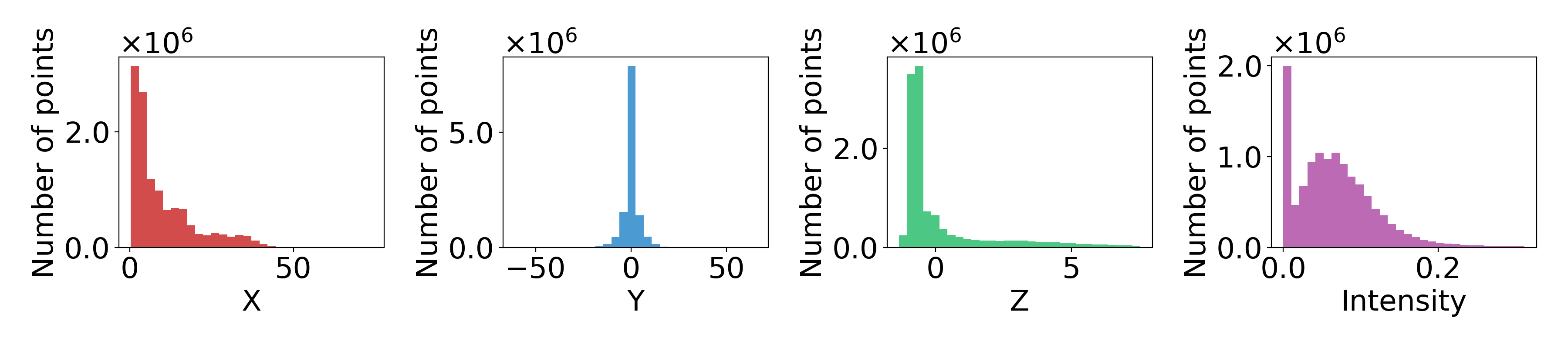}}\end{minipage}
    \\\cmidrule{3-4}
    & & skatepark\_2 & \begin{minipage}[b]{1.05\columnwidth}\centering\raisebox{-.47\height}{\includegraphics[width=\linewidth]{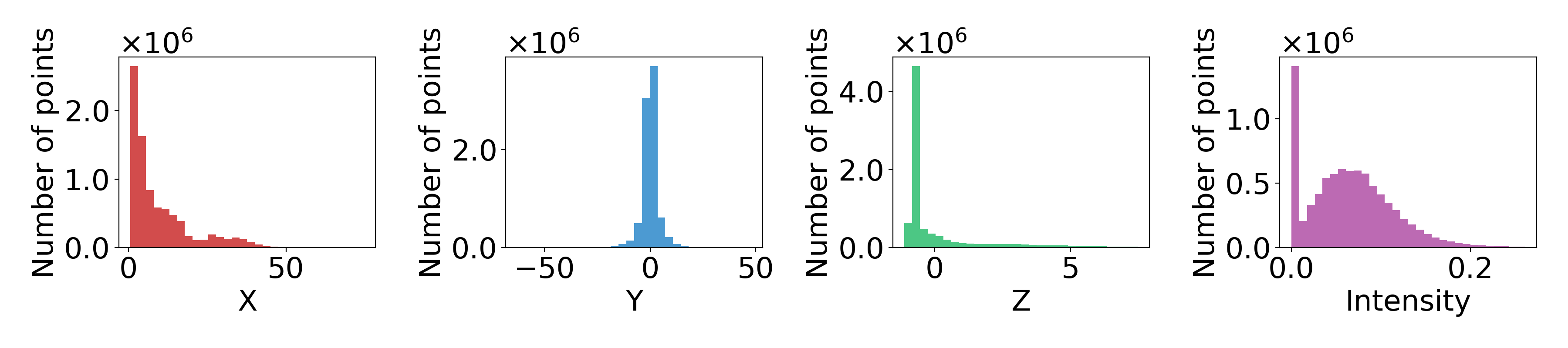}}\end{minipage}
    \\\cmidrule{3-4}
    & & srt\_green\_loop & \begin{minipage}[b]{1.05\columnwidth}\centering\raisebox{-.47\height}{\includegraphics[width=\linewidth]{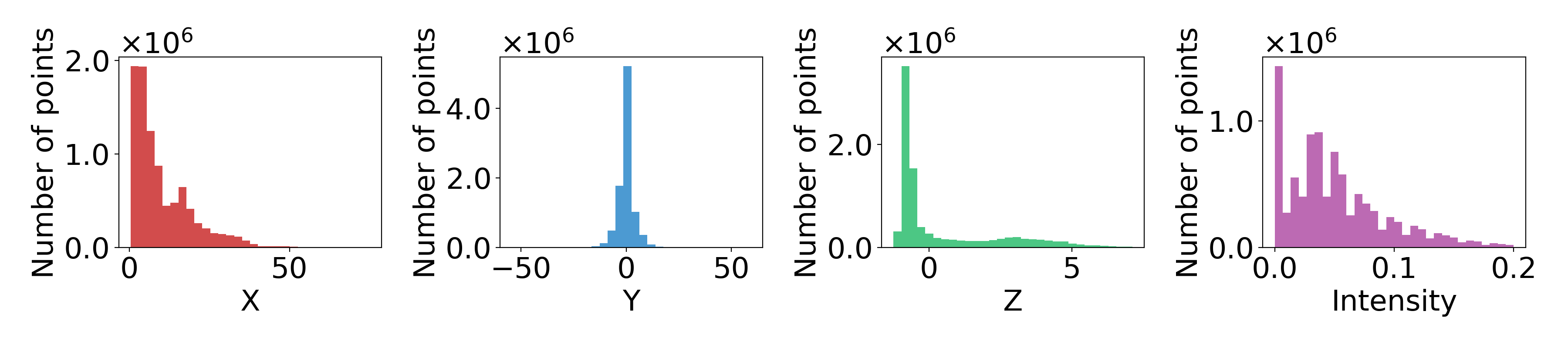}}\end{minipage}
    \\\cmidrule{3-4}
    & & srt\_under\_bridge\_1 & \begin{minipage}[b]{1.05\columnwidth}\centering\raisebox{-.47\height}{\includegraphics[width=\linewidth]{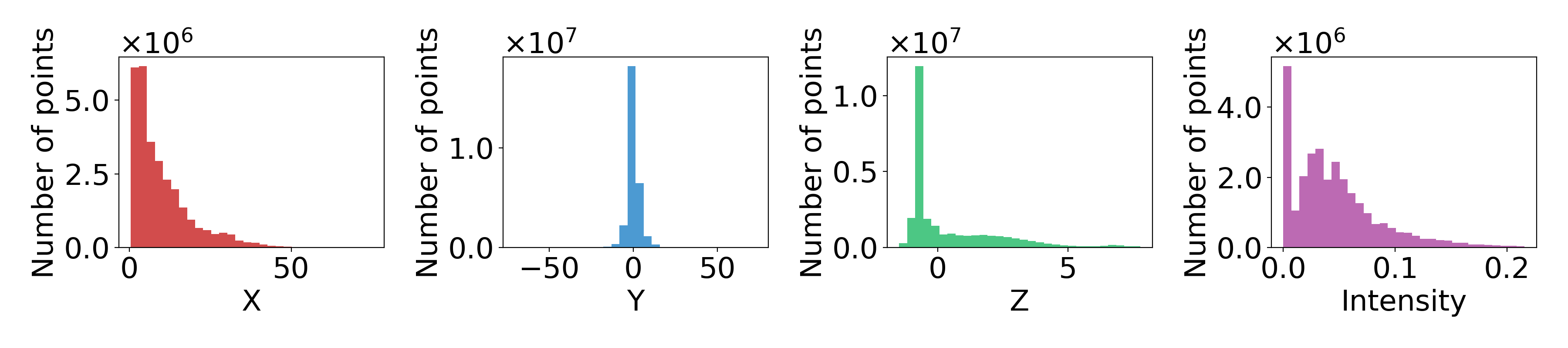}}\end{minipage}
    \\\cmidrule{3-4}
    & & srt\_under\_bridge\_2~~~~~ & \begin{minipage}[b]{1.05\columnwidth}\centering\raisebox{-.47\height}{\includegraphics[width=\linewidth]{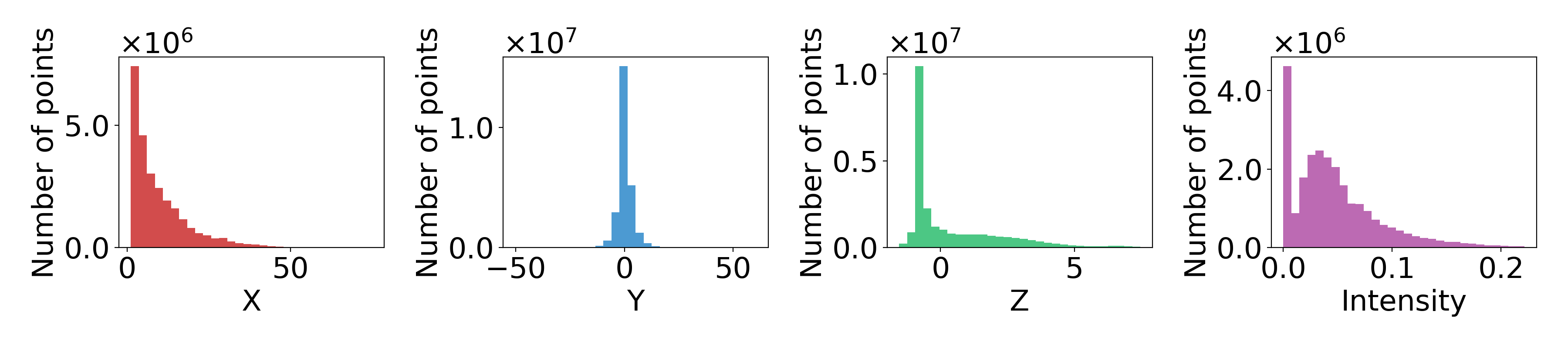}}\end{minipage}
    \\\cmidrule{2-4}
    & \multirow{6}{*}{\makecell{~~Nighttime~~\\$(2)$}} & penno\_plaza\_lights & \begin{minipage}[b]{1.05\columnwidth}\centering\raisebox{-.47\height}{\includegraphics[width=\linewidth]{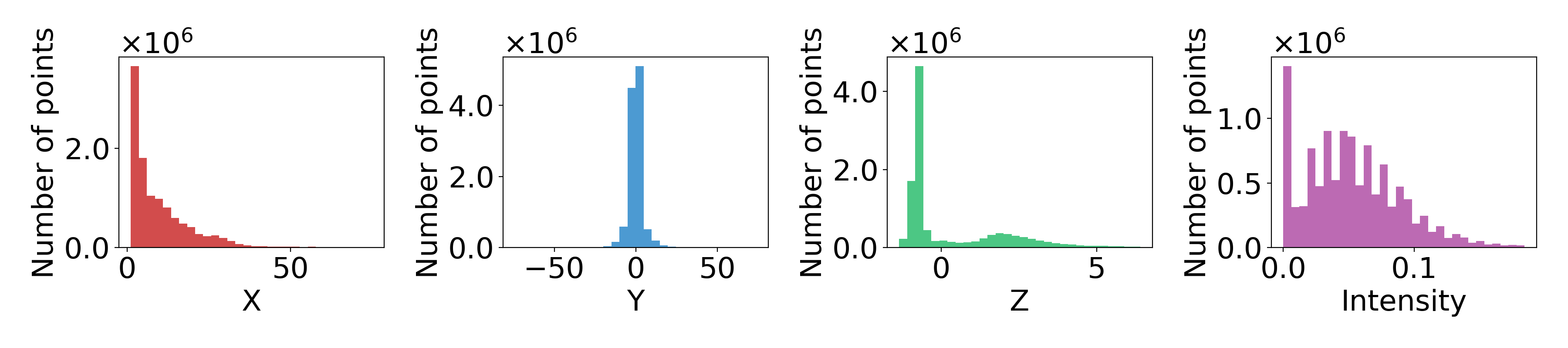}}\end{minipage}
    \\\cmidrule{3-4}
    & & penno\_short\_loop & \begin{minipage}[b]{1.05\columnwidth}\centering\raisebox{-.47\height}{\includegraphics[width=\linewidth]{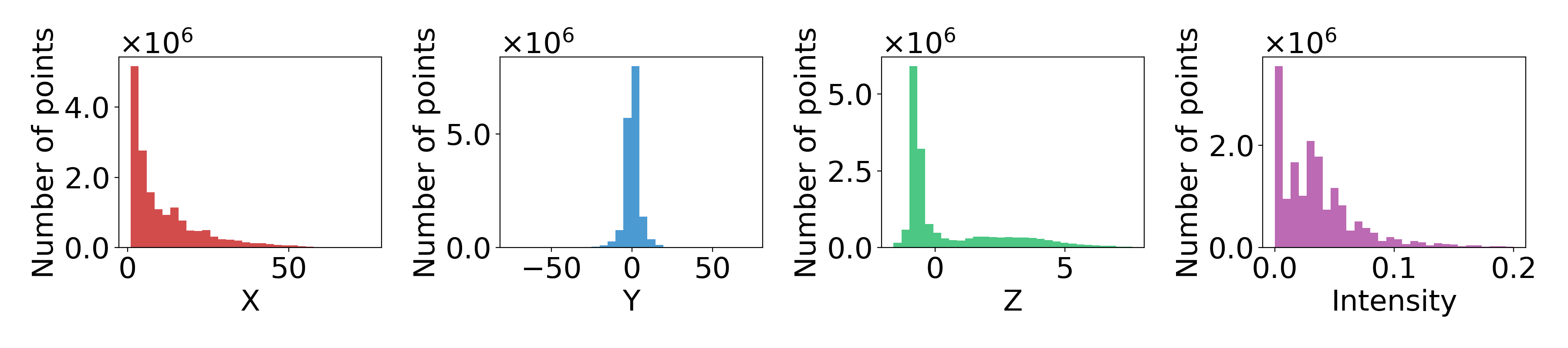}}\end{minipage}
    \\
    \bottomrule
\end{tabular}}
\label{tab:dataset_intensity_quadruped}
\end{table*}

\begin{table*}[t]
\centering
\caption{Summary of \textbf{3D bounding box statistics} (length $L$, width $W$, height $H$) of the \includegraphics[width=0.0225\linewidth]{figures/icons/vehicle.png} \textbf{Vehicle} data from \textbf{Pi3DET}.}
\vspace{-0.2cm}
\resizebox{\linewidth}{!}{
\begin{tabular}{c|c|l|c}
    \toprule
    \textbf{Platform} & \textbf{Condition} & \textbf{Sequence} & \textbf{3D Box Statistics of Veh. (Left) and Ped. (Right)}
    \\\midrule\midrule
    \multirow{66}{*}{\makecell{\textcolor{pi3det_red}{~~\textbf{Vehicle}~~}\\\textcolor{pi3det_red}{$\mathbf{(8)}$}}} & \multirow{16}{*}{\makecell{Daytime\\$(4)$}} & city\_hall & \begin{minipage}[b]{0.95\columnwidth}\centering\raisebox{-.47\height}{\includegraphics[width=\linewidth]{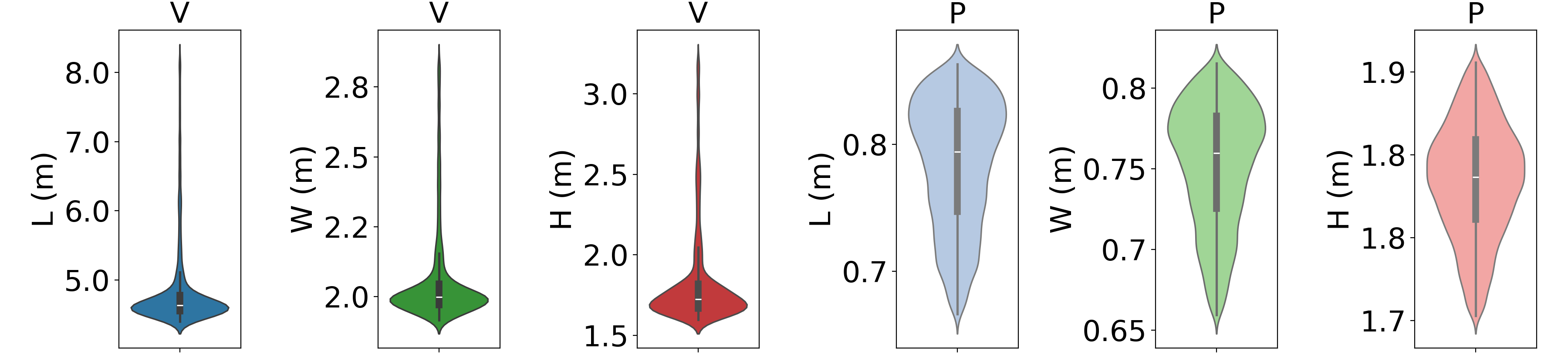}}\end{minipage}
    \\\cmidrule{3-4}
    & & penno\_big\_loop & \begin{minipage}[b]{0.95\columnwidth}\centering\raisebox{-.47\height}{\includegraphics[width=\linewidth]{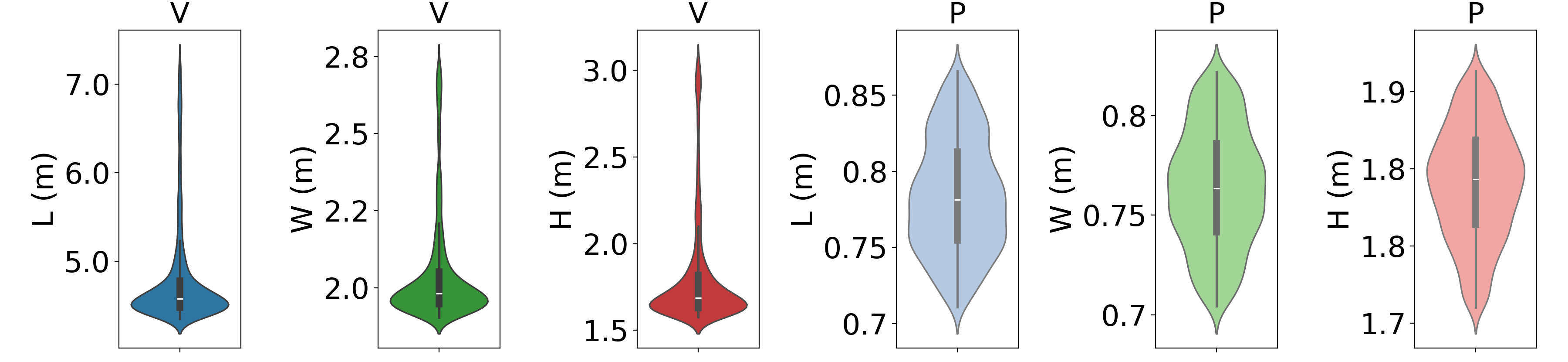}}\end{minipage}
    \\\cmidrule{3-4}
    & & rittenhouse & \begin{minipage}[b]{0.95\columnwidth}\centering\raisebox{-.47\height}{\includegraphics[width=\linewidth]{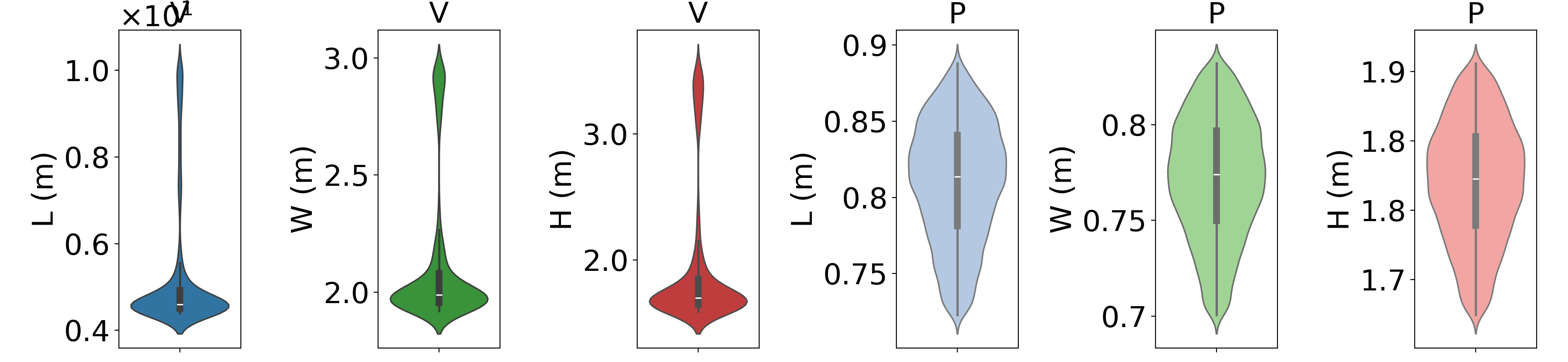}}\end{minipage}
    \\\cmidrule{3-4}
    & & ucity\_small\_loop & \begin{minipage}[b]{0.95\columnwidth}\centering\raisebox{-.47\height}{\includegraphics[width=\linewidth]{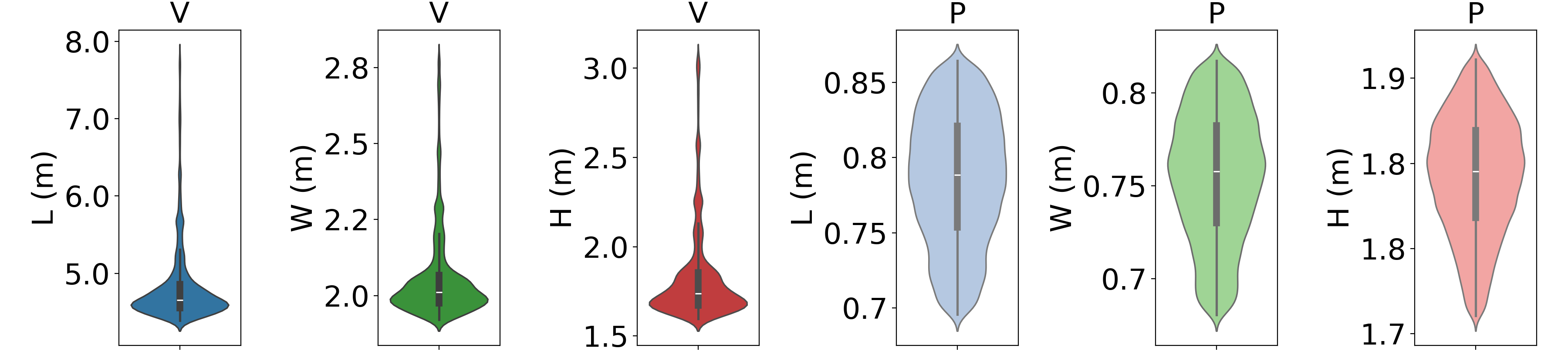}}\end{minipage}
    \\\cmidrule{2-4}
    & \multirow{16}{*}{\makecell{~~Nighttime~~\\$(4)$}} & city\_hall & \begin{minipage}[b]{0.95\columnwidth}\centering\raisebox{-.47\height}{\includegraphics[width=\linewidth]{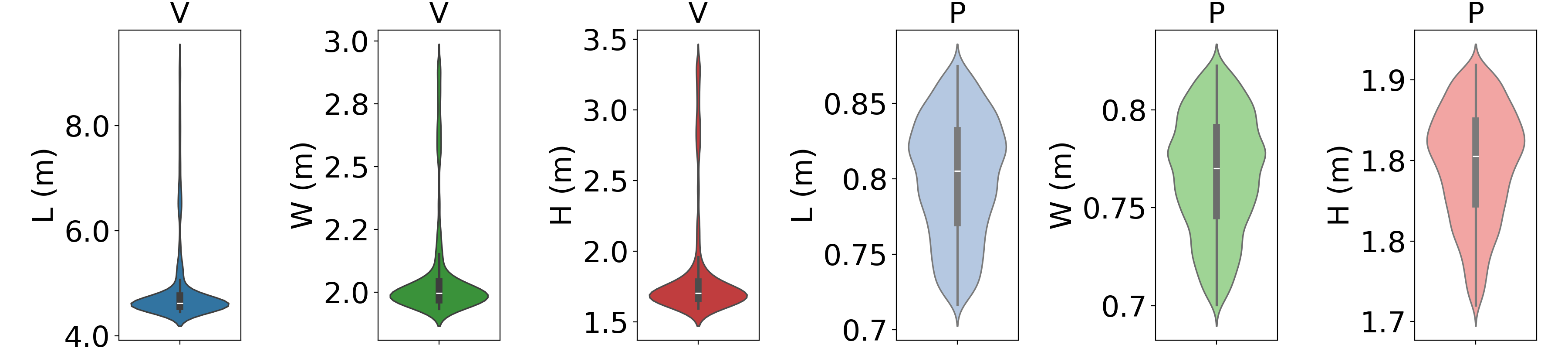}}\end{minipage}
    \\\cmidrule{3-4}
    & & penno\_big\_loop & \begin{minipage}[b]{0.95\columnwidth}\centering\raisebox{-.47\height}{\includegraphics[width=\linewidth]{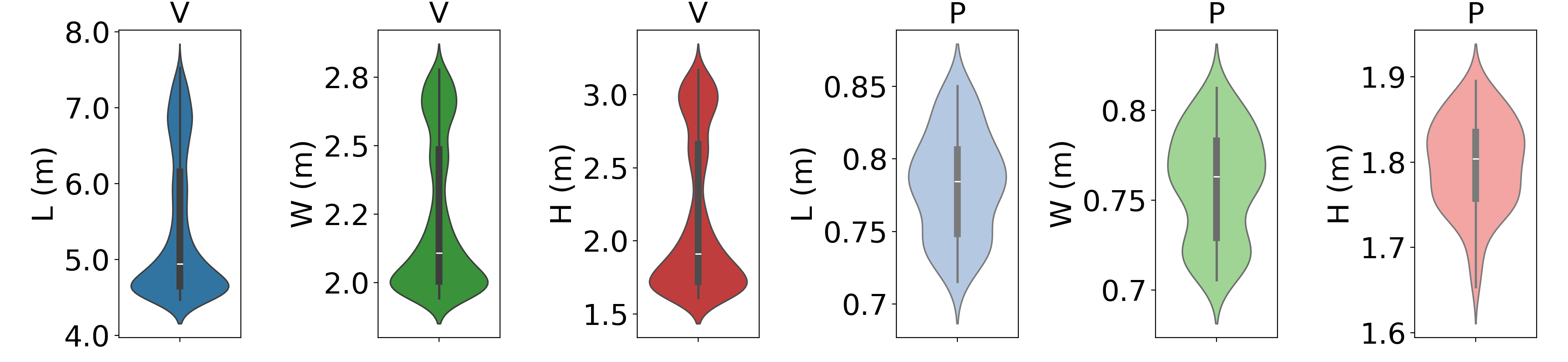}}\end{minipage}
    \\\cmidrule{3-4}
    & & rittenhouse & \begin{minipage}[b]{0.95\columnwidth}\centering\raisebox{-.47\height}{\includegraphics[width=\linewidth]{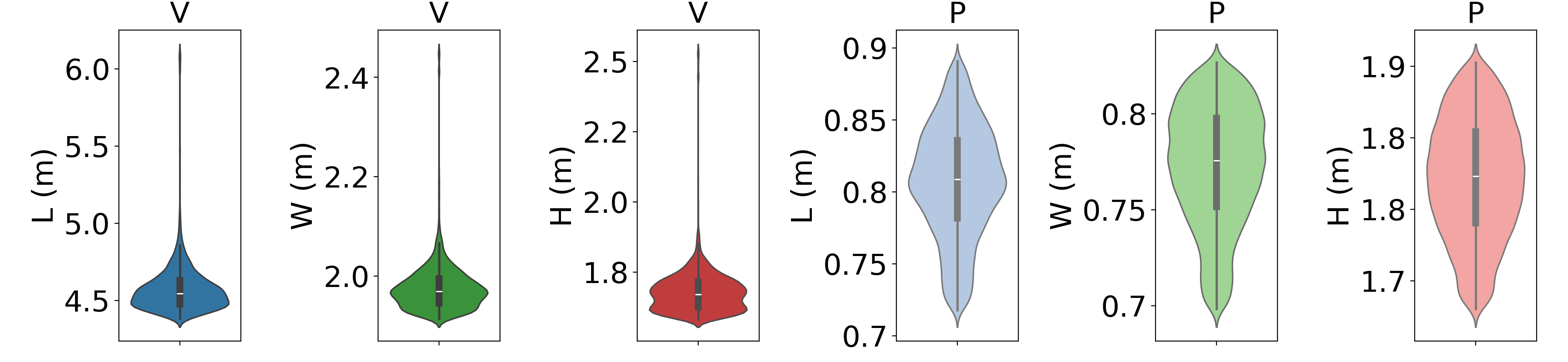}}\end{minipage}
    \\\cmidrule{3-4}
    & & ucity\_small\_loop~~~~ & \begin{minipage}[b]{0.95\columnwidth}\centering\raisebox{-.47\height}{\includegraphics[width=\linewidth]{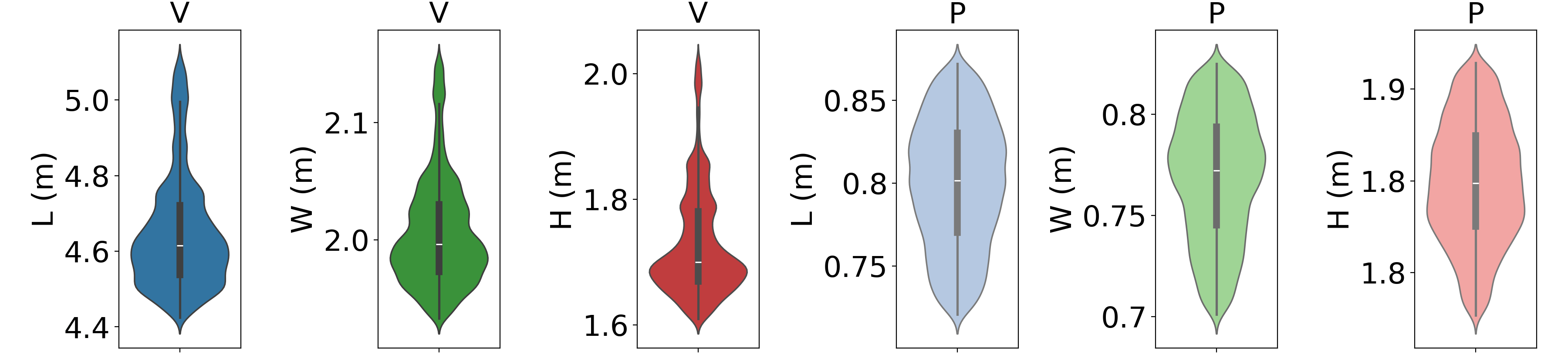}}\end{minipage}
    \\
    \bottomrule
\end{tabular}}
\label{tab:dataset_bbox_geo_vehicle}
\end{table*}
\begin{table*}[t]
\centering
\caption{Summary of \textbf{3D bounding box statistics} (length $L$, width $W$, height $H$) of the \includegraphics[width=0.023\linewidth]{figures/icons/drone.png} \textbf{Drone} data from \textbf{Pi3DET}.}
\vspace{-0.2cm}
\resizebox{\linewidth}{!}{
\begin{tabular}{c|c|l|c}
    \toprule
    \textbf{Platform} & \textbf{Condition} & \textbf{Sequence} & \textbf{3D Box Statistics of Veh. (Left) and Ped. (Right)}
    \\\midrule\midrule
    \multirow{50}{*}{\makecell{\textcolor{pi3det_blue}{~~\textbf{Drone}~~}\\\textcolor{pi3det_blue}{$\mathbf{(7)}$}}} & \multirow{16}{*}{\makecell{Daytime\\$(4)$}} & penno\_parking\_1 & \begin{minipage}[b]{0.95\columnwidth}\centering\raisebox{-.47\height}{\includegraphics[width=\linewidth]{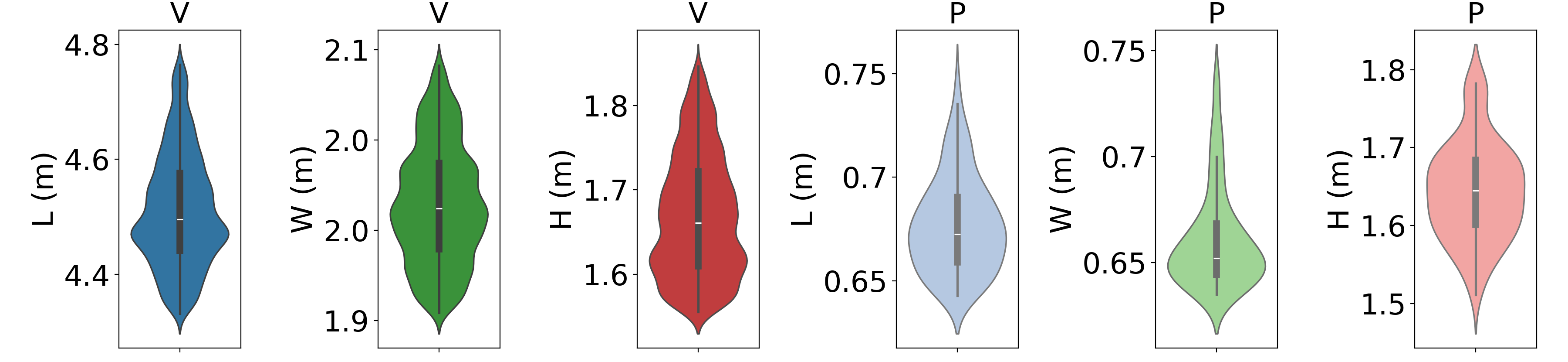}}\end{minipage}
    \\\cmidrule{3-4}
    & & penno\_parking\_2 & \begin{minipage}[b]{0.95\columnwidth}\centering\raisebox{-.47\height}{\includegraphics[width=\linewidth]{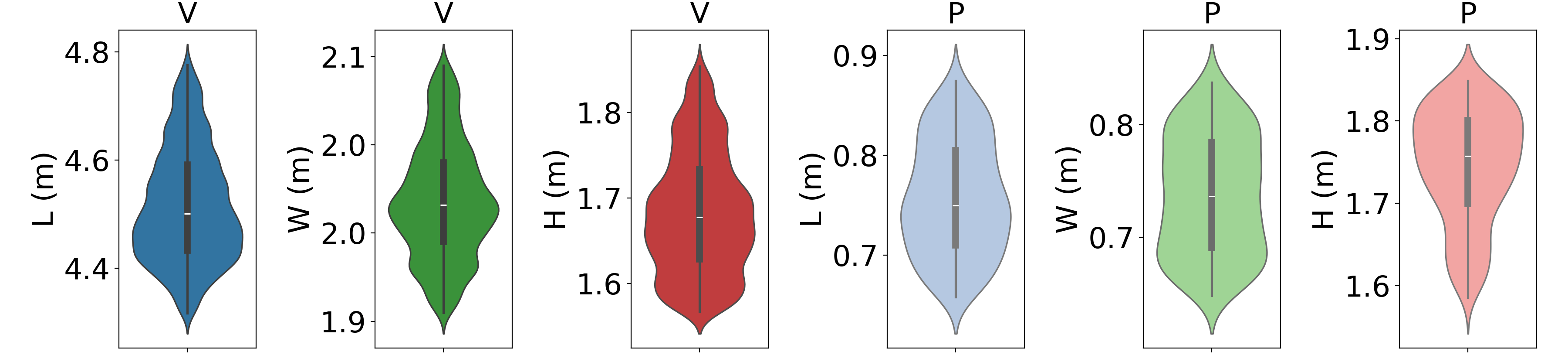}}\end{minipage}
    \\\cmidrule{3-4}
    & & penno\_plaza & \begin{minipage}[b]{0.95\columnwidth}\centering\raisebox{-.47\height}{\includegraphics[width=\linewidth]{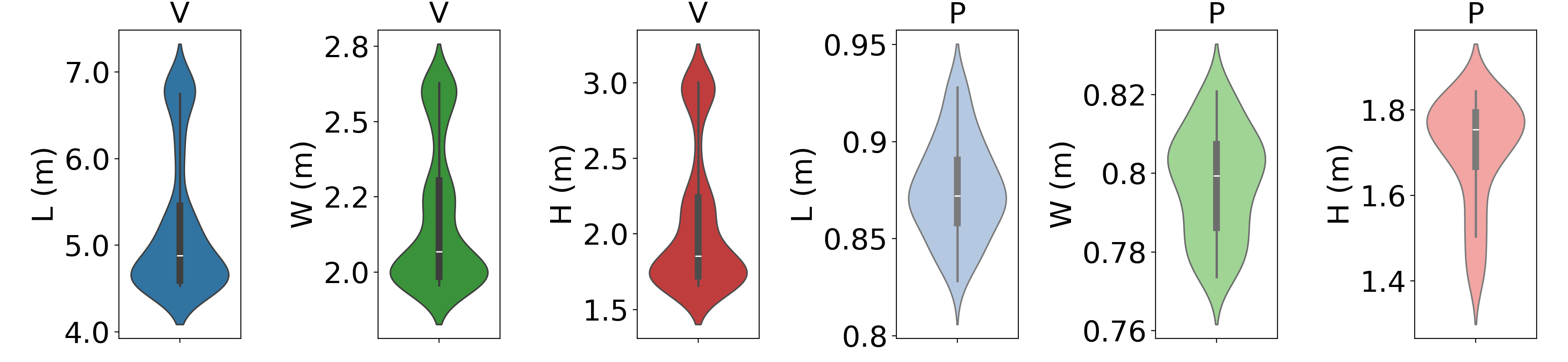}}\end{minipage}
    \\\cmidrule{3-4}
    & & penno\_trees & \begin{minipage}[b]{0.95\columnwidth}\centering\raisebox{-.47\height}{\includegraphics[width=\linewidth]{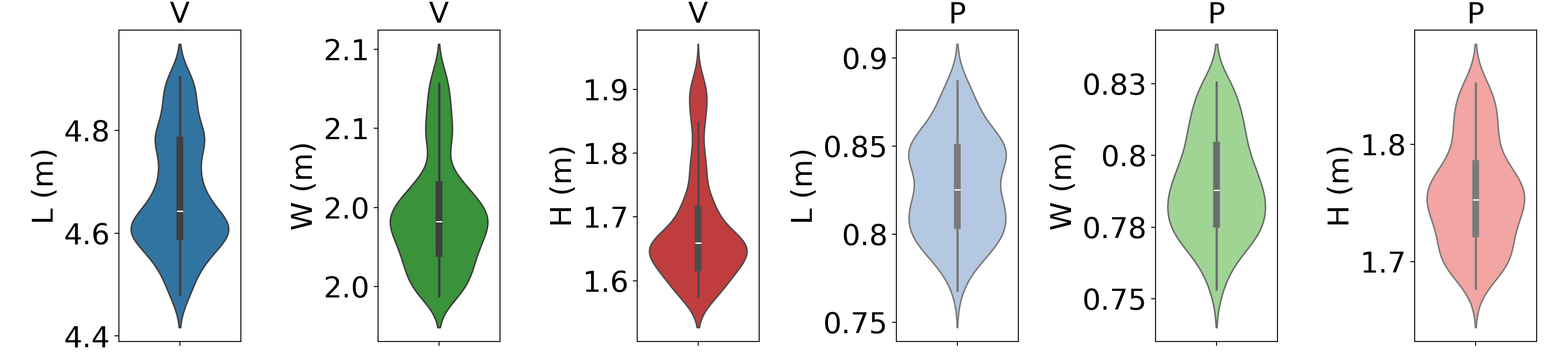}}\end{minipage}
    \\\cmidrule{2-4}
    & \multirow{12}{*}{\makecell{~~Nighttime~~\\$(3)$}} & high\_beams & \begin{minipage}[b]{0.95\columnwidth}\centering\raisebox{-.47\height}{\includegraphics[width=\linewidth]{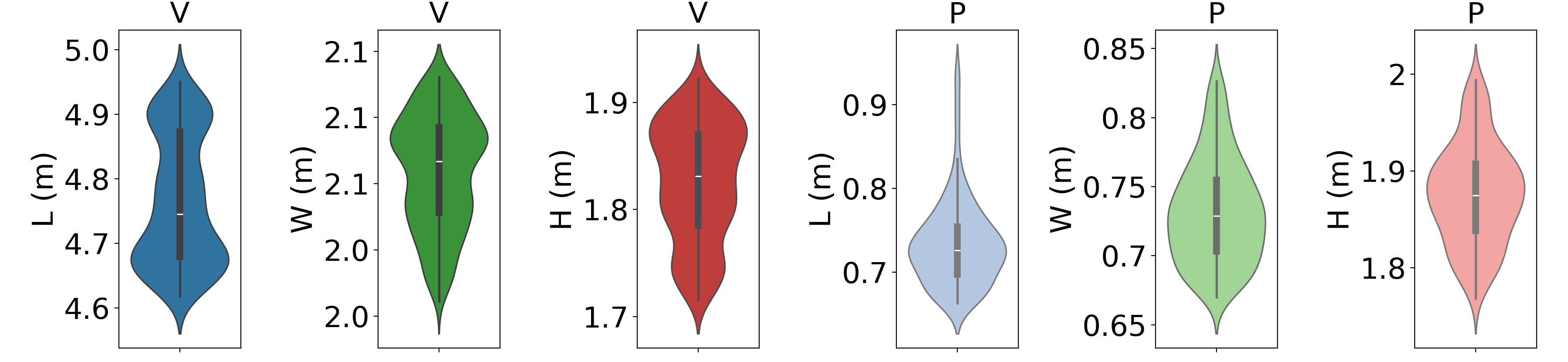}}\end{minipage}
    \\\cmidrule{3-4}
    & & penno\_parking\_1 & \begin{minipage}[b]{0.95\columnwidth}\centering\raisebox{-.47\height}{\includegraphics[width=\linewidth]{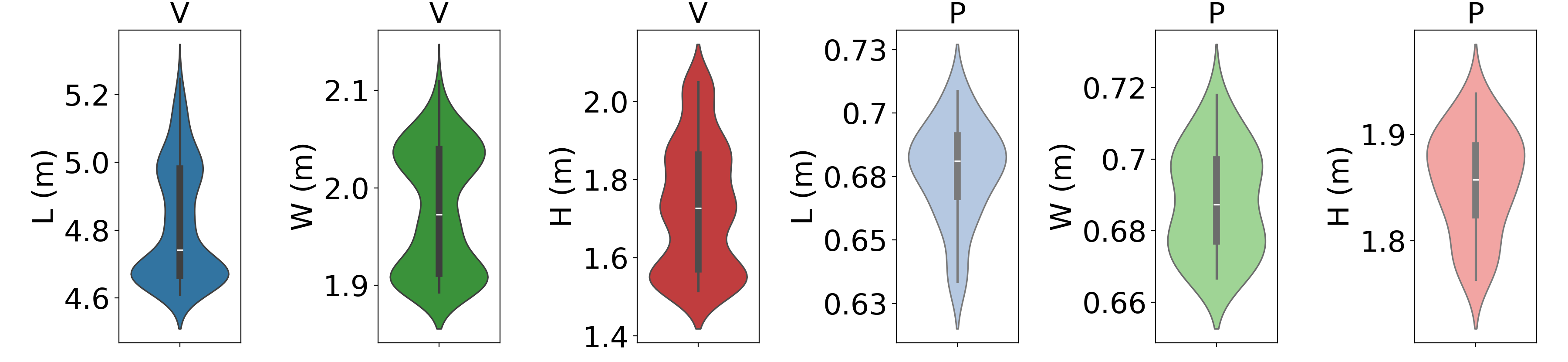}}\end{minipage}
    \\\cmidrule{3-4}
    & & penno\_parking\_2~~~~~ & \begin{minipage}[b]{0.95\columnwidth}\centering\raisebox{-.47\height}{\includegraphics[width=\linewidth]{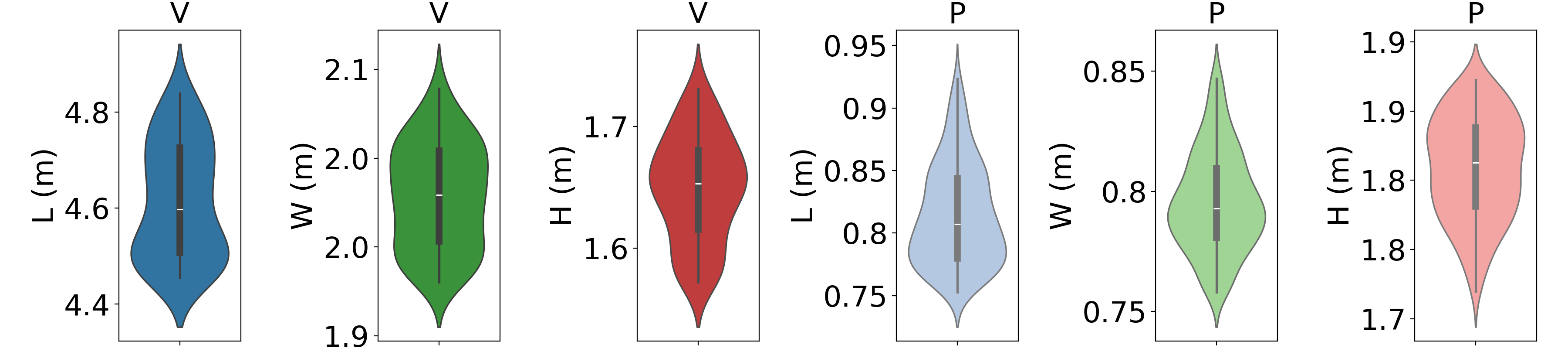}}\end{minipage}
    \\
    \bottomrule
\end{tabular}}
\label{tab:dataset_bbox_geo_drone}
\end{table*}
\begin{table*}[t]
\centering
\caption{Summary of \textbf{3D bounding box statistics} (length $L$, width $W$, height $H$) of the \includegraphics[width=0.023\linewidth]{figures/icons/quadruped.png} \textbf{Quadruped} data from \textbf{Pi3DET}.}
\vspace{-0.2cm}
\resizebox{.8\linewidth}{!}{
\begin{tabular}{c|c|l|c}
    \toprule
    \textbf{Platform} & \textbf{Condition} & \textbf{Sequence} & \textbf{3D Box Statistics of Veh. (Left) and Ped. (Right)}
    \\\midrule\midrule
    \multirow{85}{*}{\makecell{\textcolor{pi3det_green}{\textbf{Quadruped}}\\\textcolor{pi3det_green}{$\mathbf{(10)}$}}} & \multirow{34}{*}{\makecell{Daytime\\$(8)$}} & art\_plaza\_loop & \begin{minipage}[b]{0.95\columnwidth}\centering\raisebox{-.47\height}{\includegraphics[width=\linewidth]{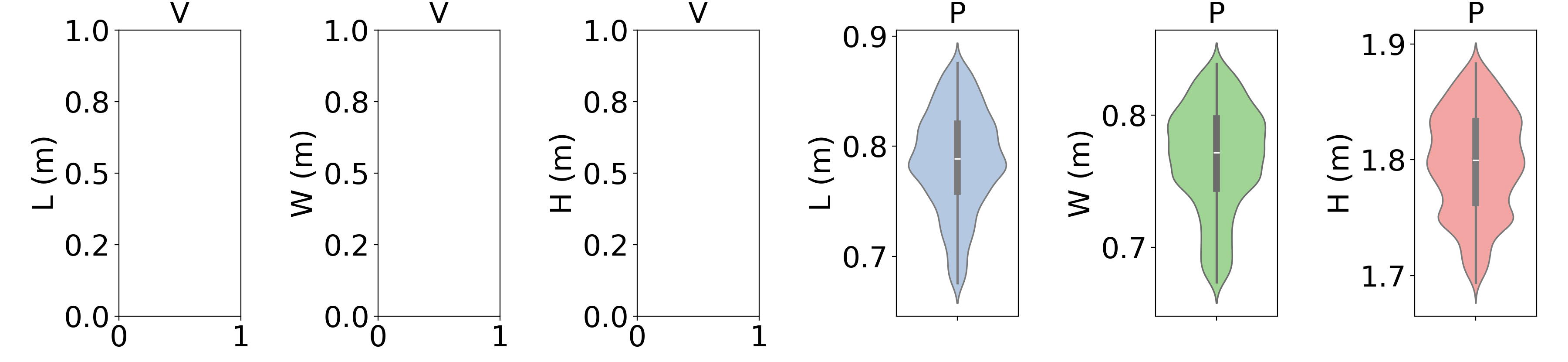}}\end{minipage}
    \\\cmidrule{3-4}
    & & penno\_short\_loop & \begin{minipage}[b]{0.95\columnwidth}\centering\raisebox{-.47\height}{\includegraphics[width=\linewidth]{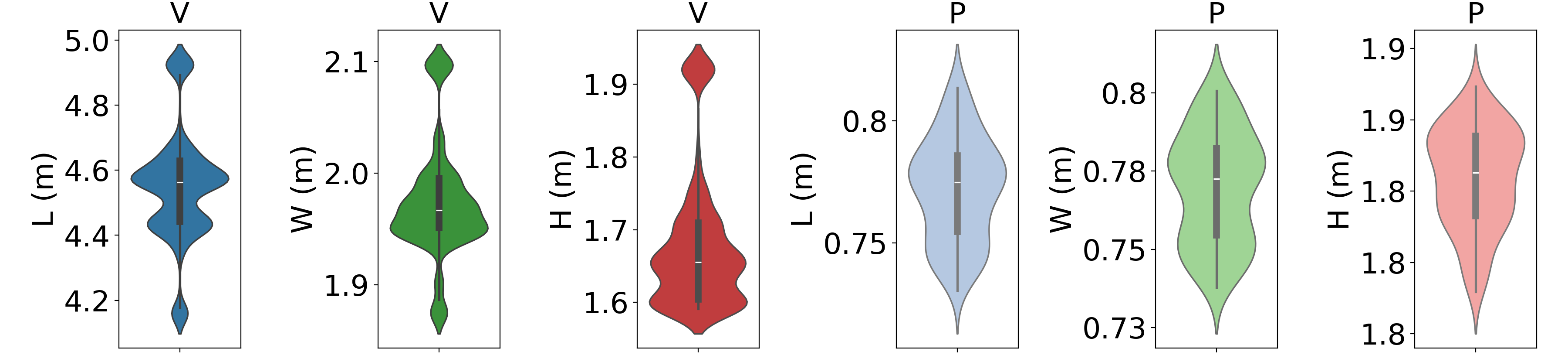}}\end{minipage}
    \\\cmidrule{3-4}
    & & rocky\_steps & \begin{minipage}[b]{0.95\columnwidth}\centering\raisebox{-.47\height}{\includegraphics[width=\linewidth]{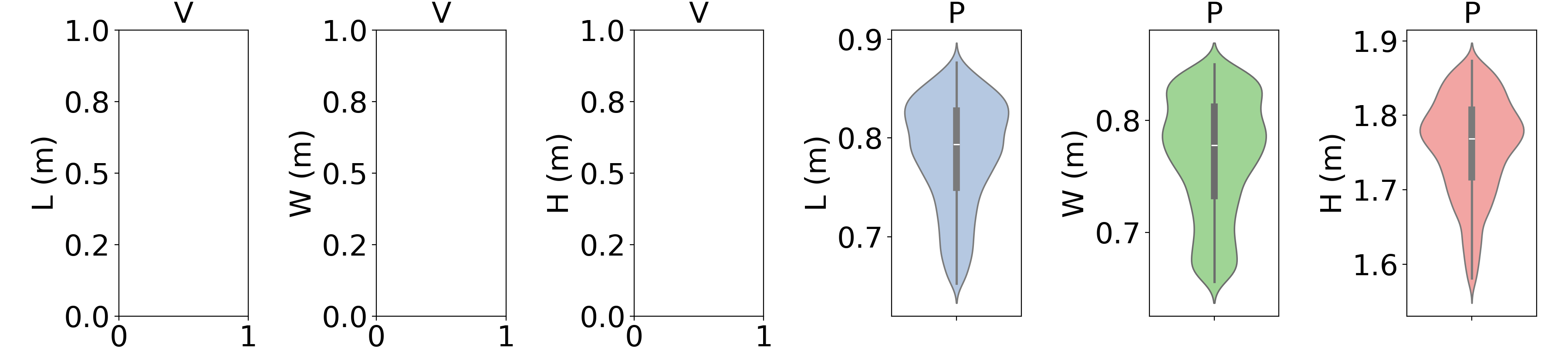}}\end{minipage}
    \\\cmidrule{3-4}
    & & skatepark\_1 & \begin{minipage}[b]{0.95\columnwidth}\centering\raisebox{-.47\height}{\includegraphics[width=\linewidth]{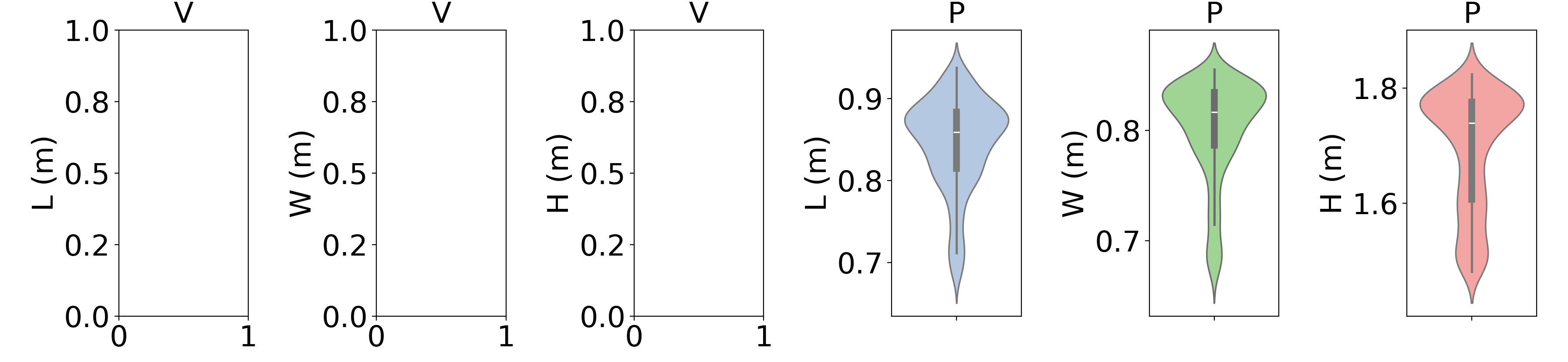}}\end{minipage}
    \\\cmidrule{3-4}
    & & skatepark\_2 & \begin{minipage}[b]{0.95\columnwidth}\centering\raisebox{-.47\height}{\includegraphics[width=\linewidth]{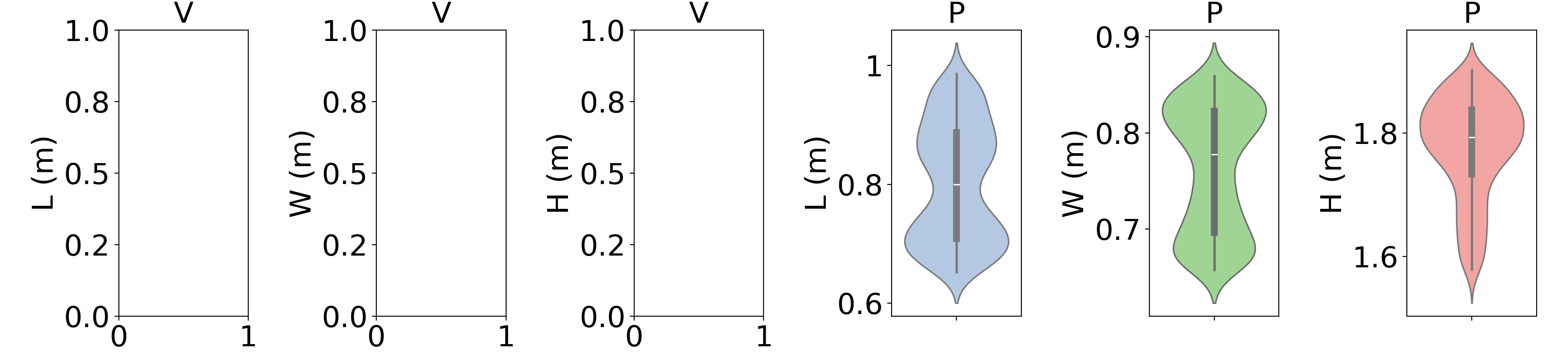}}\end{minipage}
    \\\cmidrule{3-4}
    & & srt\_green\_loop & \begin{minipage}[b]{0.95\columnwidth}\centering\raisebox{-.47\height}{\includegraphics[width=\linewidth]{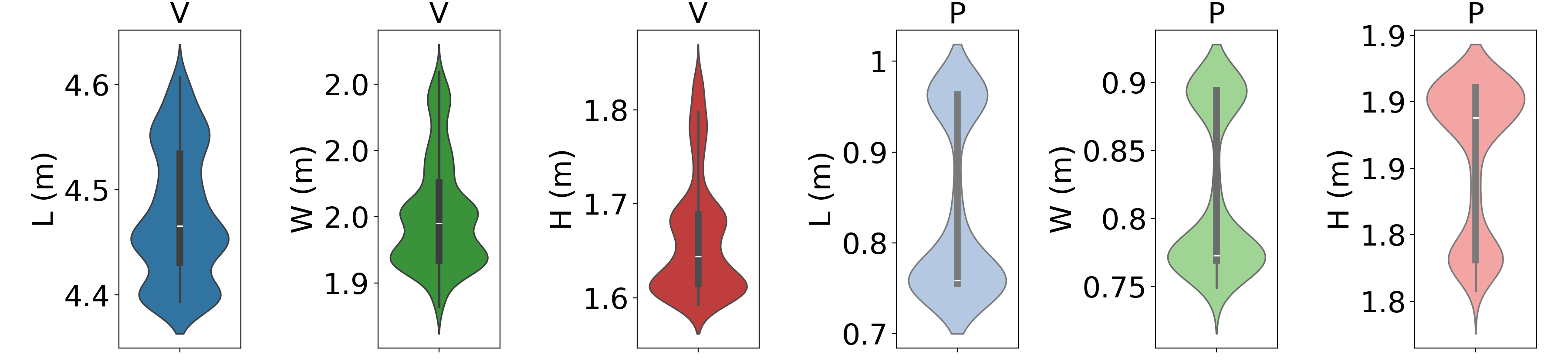}}\end{minipage}
    \\\cmidrule{3-4}
    & & srt\_under\_bridge\_1 & \begin{minipage}[b]{0.95\columnwidth}\centering\raisebox{-.47\height}{\includegraphics[width=\linewidth]{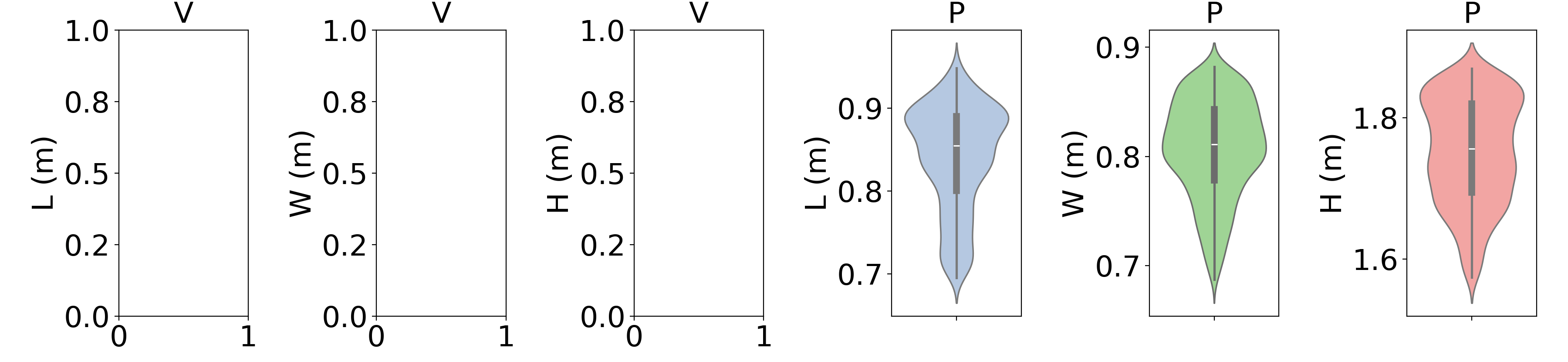}}\end{minipage}
    \\\cmidrule{3-4}
    & & srt\_under\_bridge\_2~~~~~ & \begin{minipage}[b]{0.95\columnwidth}\centering\raisebox{-.47\height}{\includegraphics[width=\linewidth]{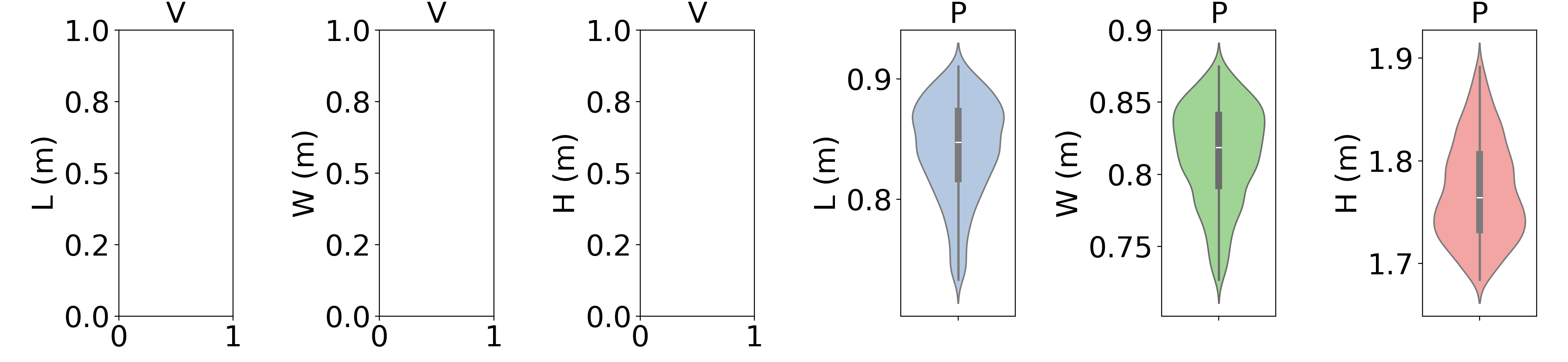}}\end{minipage}
    \\\cmidrule{2-4}
    & \multirow{6}{*}{\makecell{~~Nighttime~~\\$(2)$}} & penno\_plaza\_lights & \begin{minipage}[b]{0.95\columnwidth}\centering\raisebox{-.47\height}{\includegraphics[width=\linewidth]{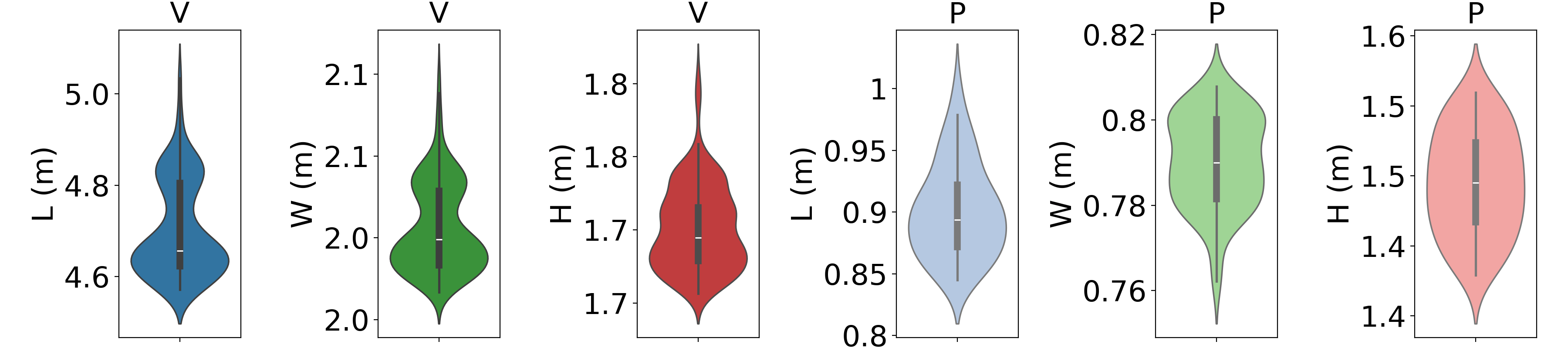}}\end{minipage}
    \\\cmidrule{3-4}
    & & penno\_short\_loop & \begin{minipage}[b]{0.95\columnwidth}\centering\raisebox{-.47\height}{\includegraphics[width=\linewidth]{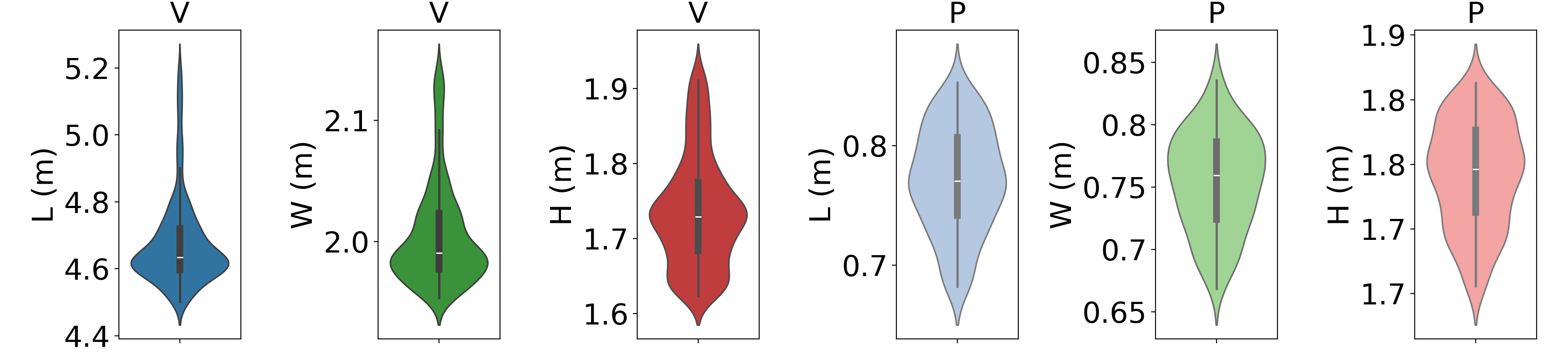}}\end{minipage}
    \\
    \bottomrule
\end{tabular}}
\label{tab:dataset_bbox_geo_quadruped}
\end{table*}

\begin{table*}[t]
\centering
\caption{Summary of \textbf{3D object statistics} (objects per frame and points per box) of the \includegraphics[width=0.0225\linewidth]{figures/icons/vehicle.png} \textbf{Vehicle} data from \textbf{Pi3DET}.}
\vspace{-0.2cm}
\resizebox{\linewidth}{!}{
\begin{tabular}{c|c|l|c}
    \toprule
    \textbf{Platform} & \textbf{Condition} & \textbf{Sequence} & \textbf{Objects Per Frame (Left) and Points Per Box (Right)}
    \\\midrule\midrule
    \multirow{66}{*}{\makecell{\textcolor{pi3det_red}{~~\textbf{Vehicle}~~}\\\textcolor{pi3det_red}{$\mathbf{(8)}$}}} & \multirow{16}{*}{\makecell{Daytime\\$(4)$}} & city\_hall & \begin{minipage}[b]{1.05\columnwidth}\centering\raisebox{-.47\height}{\includegraphics[width=\linewidth]{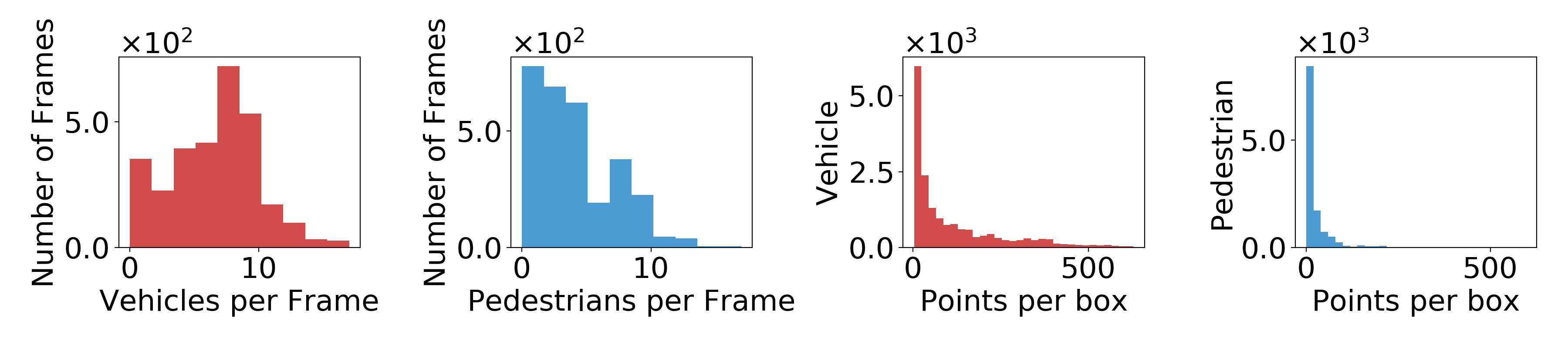}}\end{minipage}
    \\\cmidrule{3-4}
    & & penno\_big\_loop & \begin{minipage}[b]{1.05\columnwidth}\centering\raisebox{-.47\height}{\includegraphics[width=\linewidth]{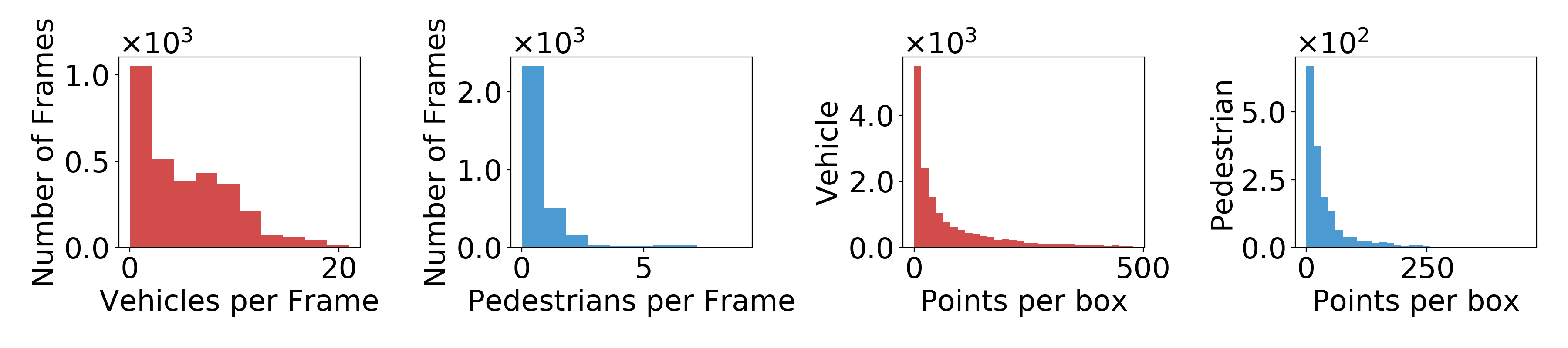}}\end{minipage}
    \\\cmidrule{3-4}
    & & rittenhouse & \begin{minipage}[b]{1.05\columnwidth}\centering\raisebox{-.47\height}{\includegraphics[width=\linewidth]{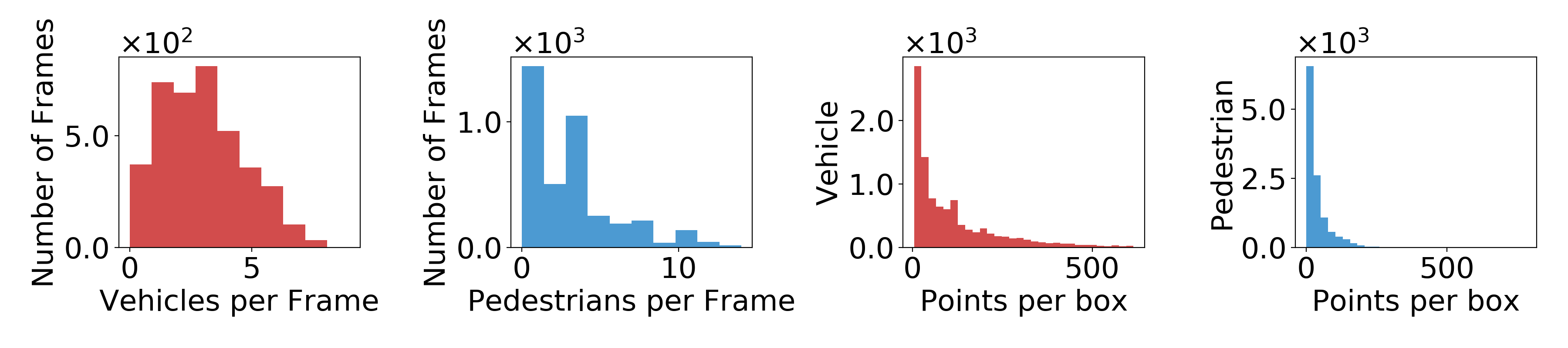}}\end{minipage}
    \\\cmidrule{3-4}
    & & ucity\_small\_loop & \begin{minipage}[b]{1.05\columnwidth}\centering\raisebox{-.47\height}{\includegraphics[width=\linewidth]{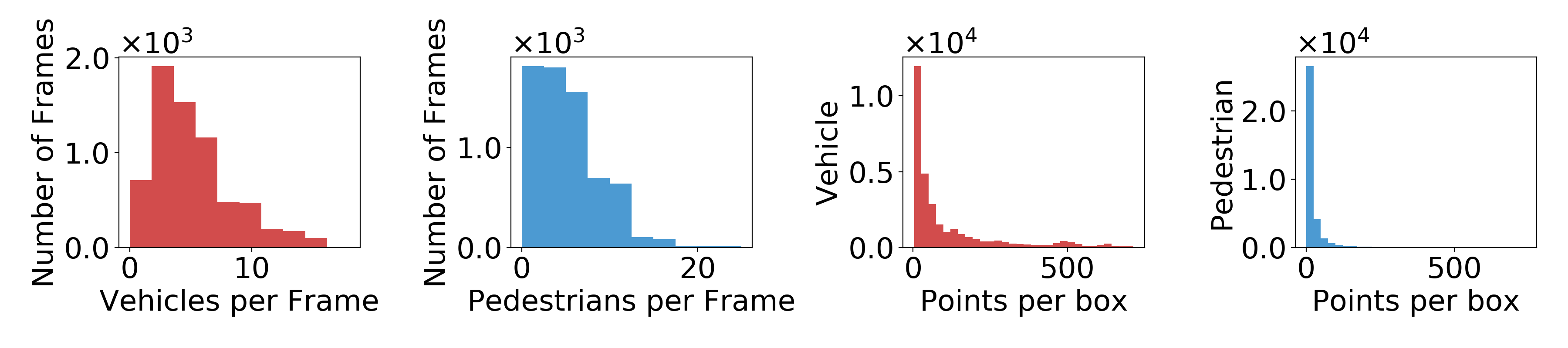}}\end{minipage}
    \\\cmidrule{2-4}
    & \multirow{16}{*}{\makecell{~~Nighttime~~\\$(4)$}} & city\_hall & \begin{minipage}[b]{1.05\columnwidth}\centering\raisebox{-.47\height}{\includegraphics[width=\linewidth]{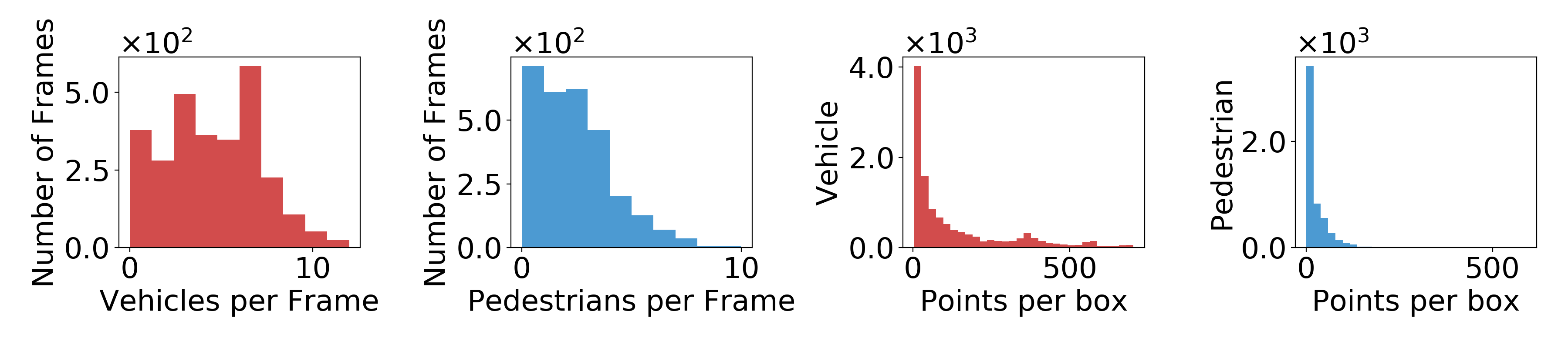}}\end{minipage}
    \\\cmidrule{3-4}
    & & penno\_big\_loop & \begin{minipage}[b]{1.05\columnwidth}\centering\raisebox{-.47\height}{\includegraphics[width=\linewidth]{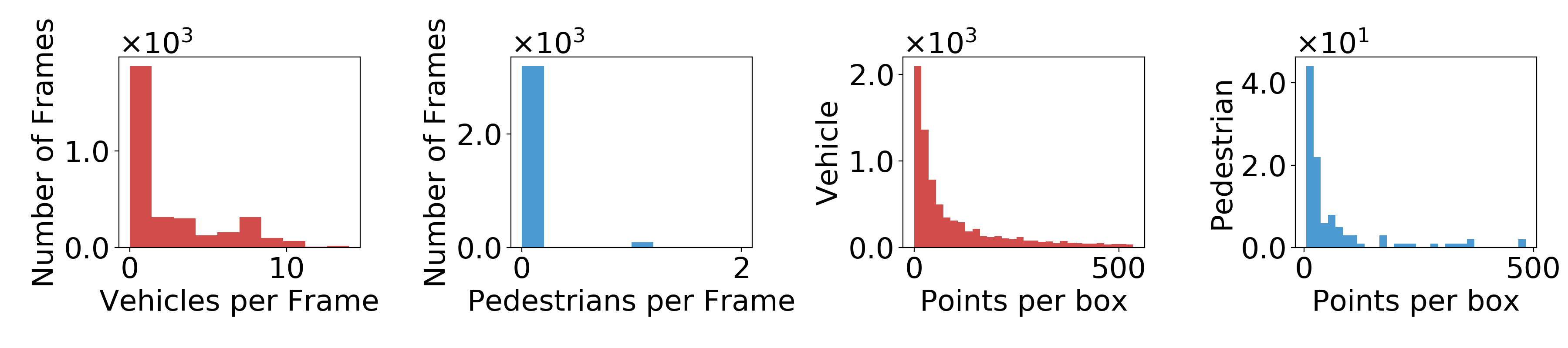}}\end{minipage}
    \\\cmidrule{3-4}
    & & rittenhouse & \begin{minipage}[b]{1.05\columnwidth}\centering\raisebox{-.47\height}{\includegraphics[width=\linewidth]{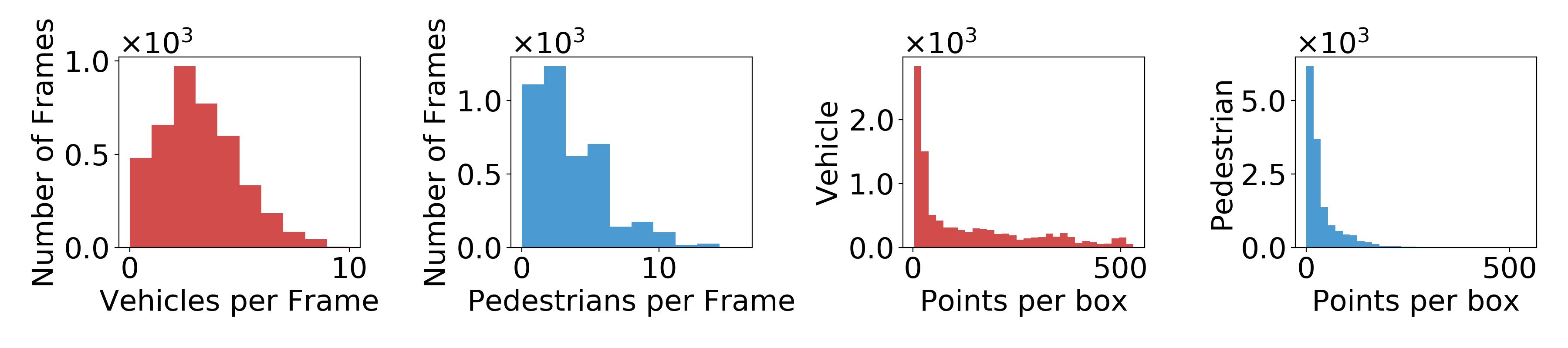}}\end{minipage}
    \\\cmidrule{3-4}
    & & ucity\_small\_loop~~~~ & \begin{minipage}[b]{1.05\columnwidth}\centering\raisebox{-.47\height}{\includegraphics[width=\linewidth]{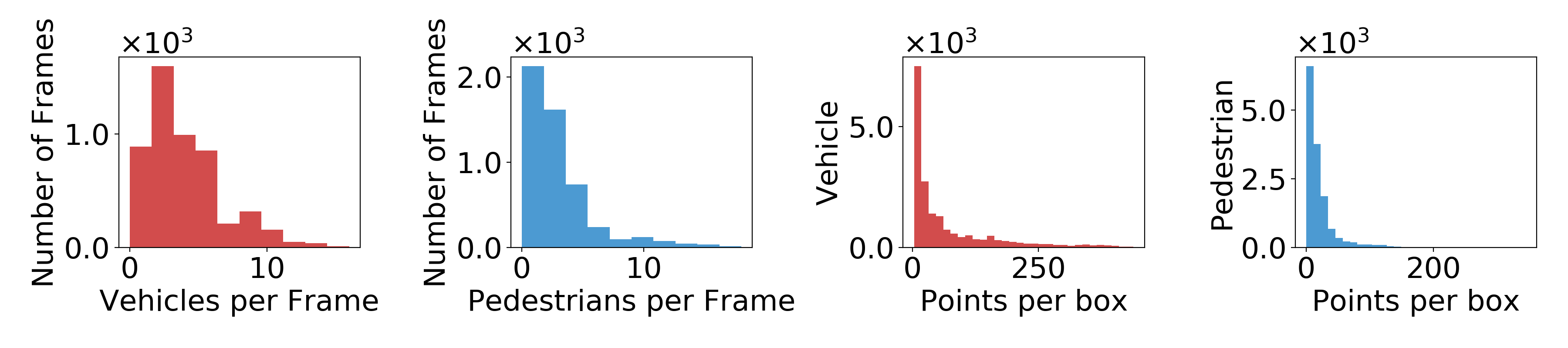}}\end{minipage}
    \\
    \bottomrule
\end{tabular}}
\label{tab:dataset_bbox_vehicle}
\end{table*}

\begin{table*}[t]
\centering
\caption{Summary of \textbf{3D object statistics} (objects per frame and points per box) of the \includegraphics[width=0.023\linewidth]{figures/icons/drone.png} \textbf{Drone} data from \textbf{Pi3DET}.}
\vspace{-0.2cm}
\resizebox{\linewidth}{!}{
\begin{tabular}{c|c|l|c}
    \toprule
    \textbf{Platform} & \textbf{Condition} & \textbf{Sequence} & \textbf{Point Cloud Distributions (X, Y, Z, Intensity)}
    \\\midrule\midrule
    \multirow{55}{*}{\makecell{\textcolor{pi3det_blue}{~~\textbf{Drone}~~}\\\textcolor{pi3det_blue}{$\mathbf{(7)}$}}} & \multirow{16}{*}{\makecell{Daytime\\$(4)$}} & penno\_parking\_1 & \begin{minipage}[b]{1.05\columnwidth}\centering\raisebox{-.47\height}{\includegraphics[width=\linewidth]{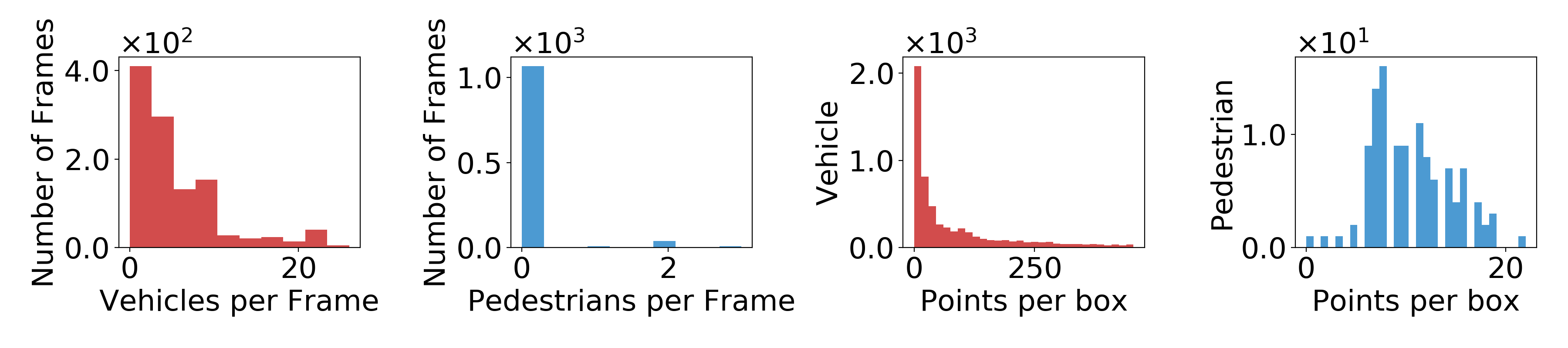}}\end{minipage}
    \\\cmidrule{3-4}
    & & penno\_parking\_2 & \begin{minipage}[b]{1.05\columnwidth}\centering\raisebox{-.47\height}{\includegraphics[width=\linewidth]{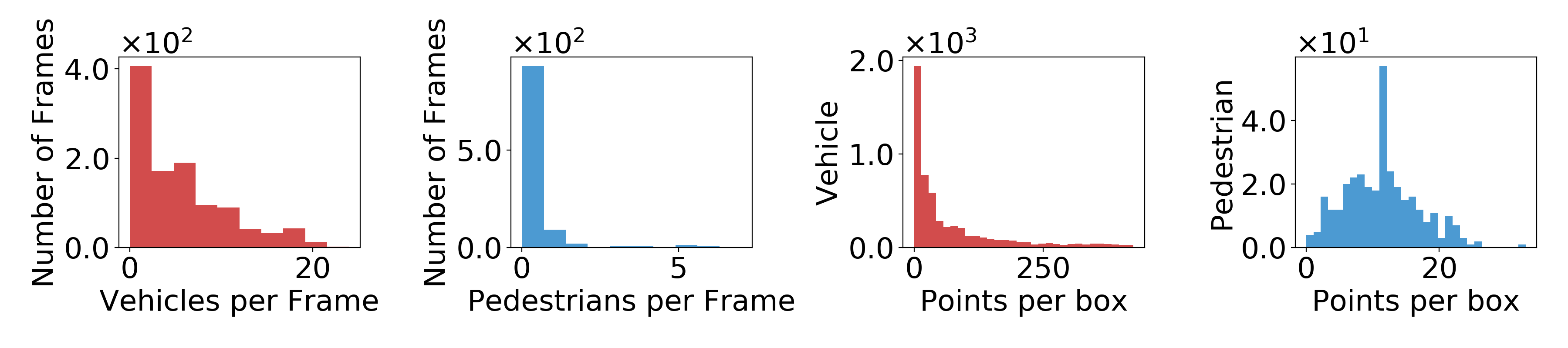}}\end{minipage}
    \\\cmidrule{3-4}
    & & penno\_plaza & \begin{minipage}[b]{1.05\columnwidth}\centering\raisebox{-.47\height}{\includegraphics[width=\linewidth]{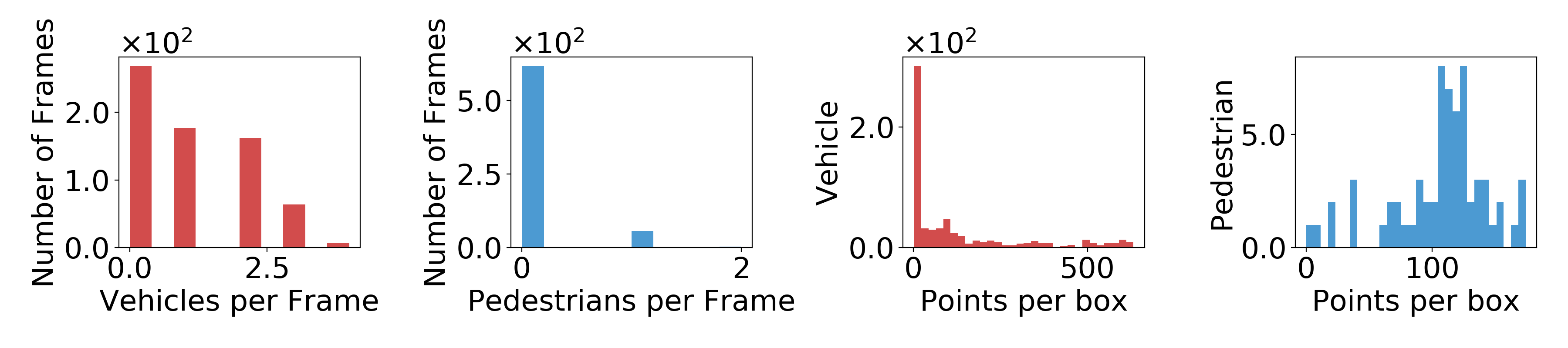}}\end{minipage}
    \\\cmidrule{3-4}
    & & penno\_trees & \begin{minipage}[b]{1.05\columnwidth}\centering\raisebox{-.47\height}{\includegraphics[width=\linewidth]{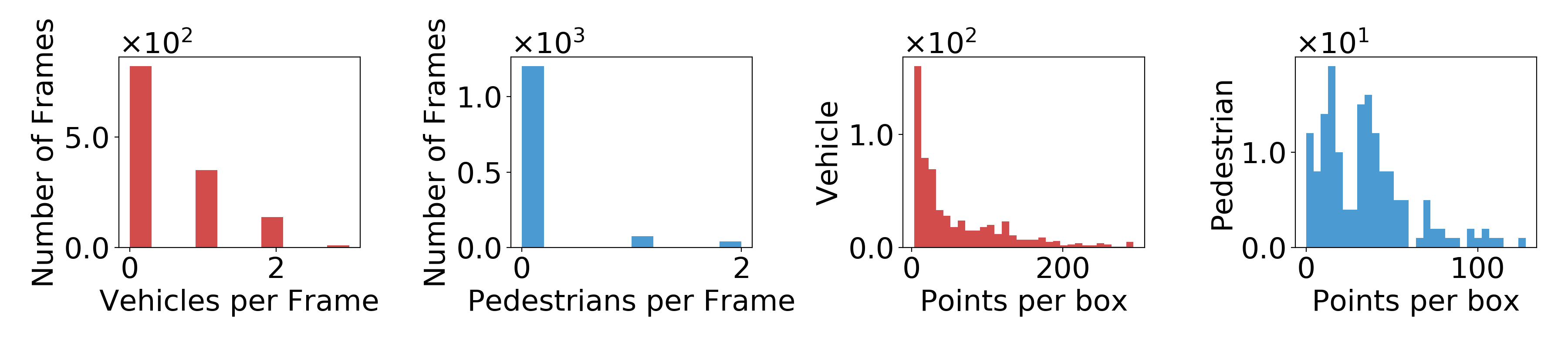}}\end{minipage}
    \\\cmidrule{2-4}
    & \multirow{12}{*}{\makecell{~~Nighttime~~\\$(3)$}} & high\_beams & \begin{minipage}[b]{1.05\columnwidth}\centering\raisebox{-.47\height}{\includegraphics[width=\linewidth]{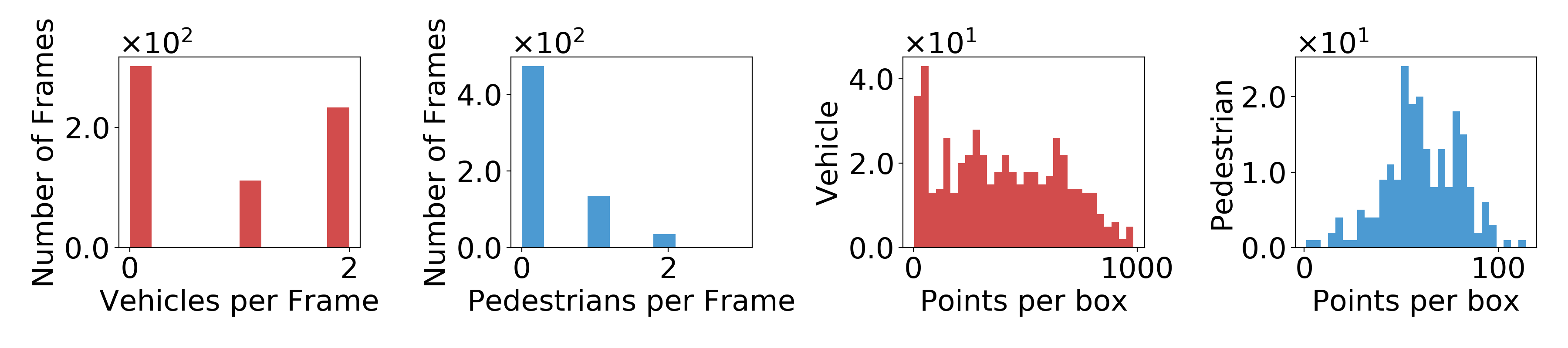}}\end{minipage}
    \\\cmidrule{3-4}
    & & penno\_parking\_1 & \begin{minipage}[b]{1.05\columnwidth}\centering\raisebox{-.47\height}{\includegraphics[width=\linewidth]{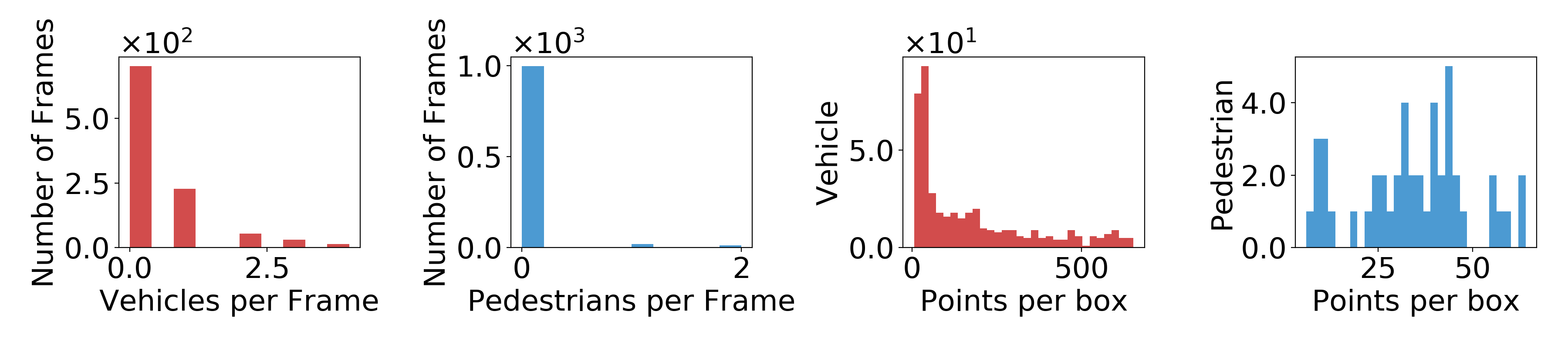}}\end{minipage}
    \\\cmidrule{3-4}
    & & penno\_parking\_2~~~~~ & \begin{minipage}[b]{1.05\columnwidth}\centering\raisebox{-.47\height}{\includegraphics[width=\linewidth]{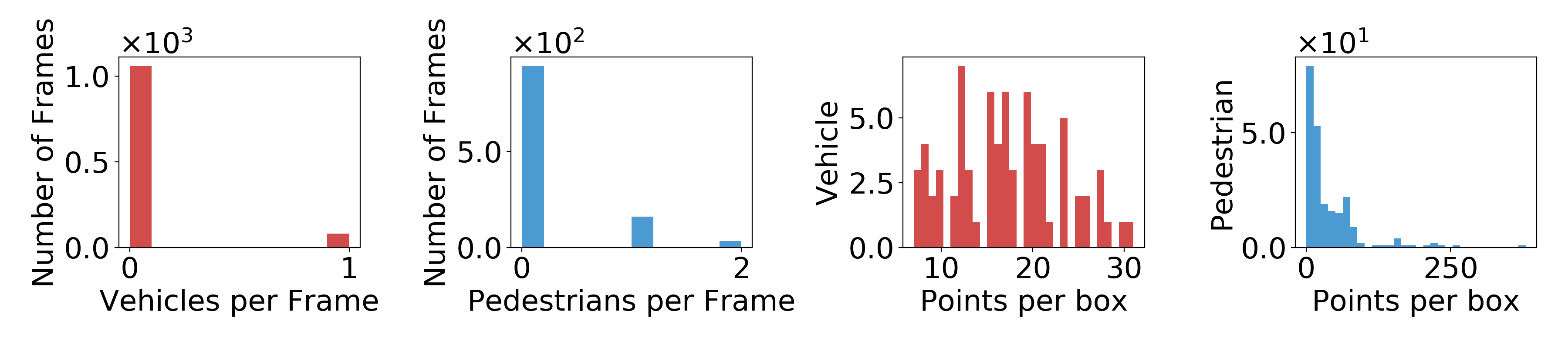}}\end{minipage}
    \\
    \bottomrule
\end{tabular}}
\label{tab:dataset_bbox_drone}
\end{table*}
\begin{table*}[t]
\centering
\caption{Summary of \textbf{3D object statistics} (objects per frame and points per box) of the \includegraphics[width=0.023\linewidth]{figures/icons/quadruped.png} \textbf{Quadruped} data from \textbf{Pi3DET}.}
\vspace{-0.2cm}
\resizebox{.8\linewidth}{!}{
\begin{tabular}{c|c|l|c}
    \toprule
    \textbf{Platform} & \textbf{Condition} & \textbf{Sequence} & \textbf{Point Cloud Distributions (X, Y, Z, Intensity)}
    \\\midrule\midrule
    \multirow{85}{*}{\makecell{\textcolor{pi3det_green}{\textbf{Quadruped}}\\\textcolor{pi3det_green}{$\mathbf{(10)}$}}} & \multirow{36}{*}{\makecell{Daytime\\$(8)$}} & art\_plaza\_loop & \begin{minipage}[b]{1.05\columnwidth}\centering\raisebox{-.47\height}{\includegraphics[width=\linewidth]{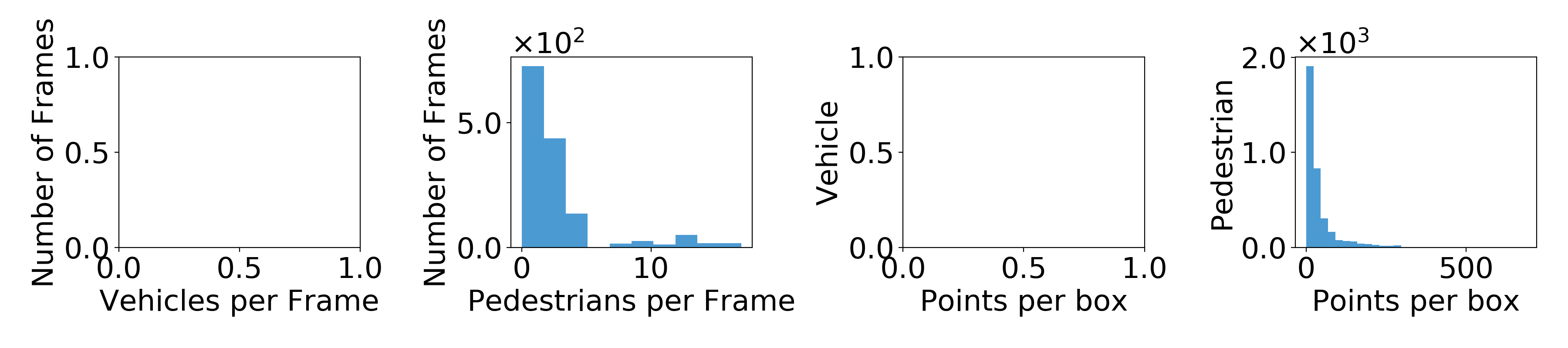}}\end{minipage}
    \\\cmidrule{3-4}
    & & penno\_short\_loop & \begin{minipage}[b]{1.05\columnwidth}\centering\raisebox{-.47\height}{\includegraphics[width=\linewidth]{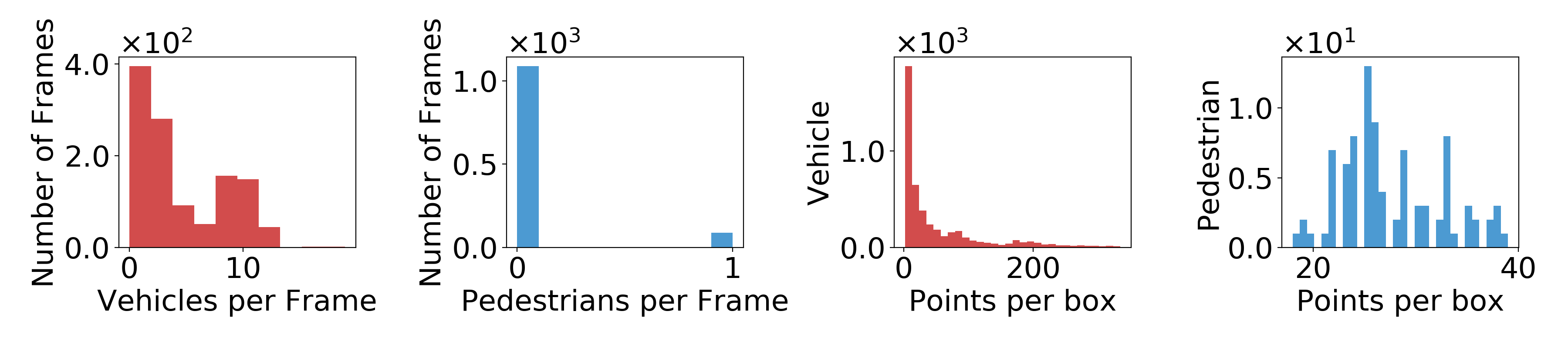}}\end{minipage}
    \\\cmidrule{3-4}
    & & rocky\_steps & \begin{minipage}[b]{1.05\columnwidth}\centering\raisebox{-.47\height}{\includegraphics[width=\linewidth]{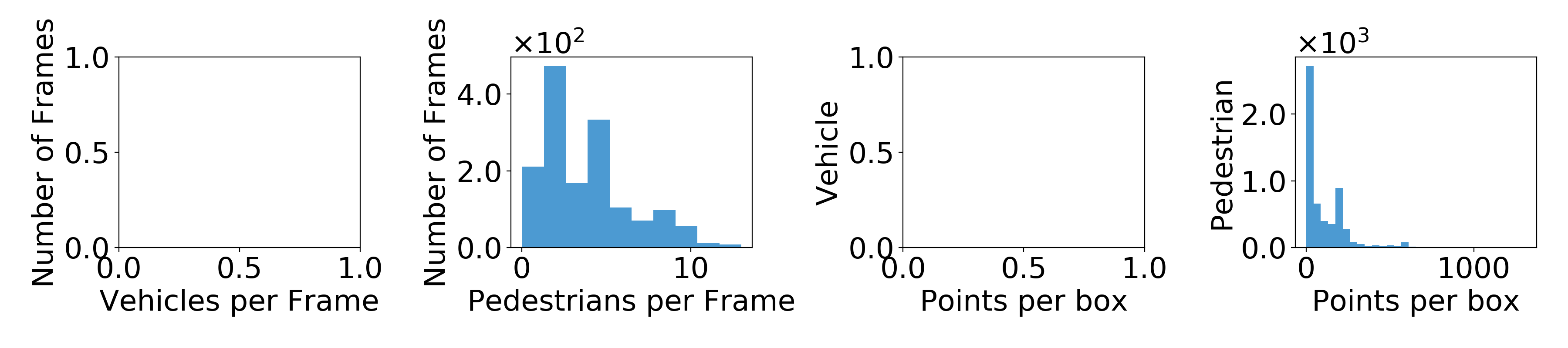}}\end{minipage}
    \\\cmidrule{3-4}
    & & skatepark\_1 & \begin{minipage}[b]{1.05\columnwidth}\centering\raisebox{-.47\height}{\includegraphics[width=\linewidth]{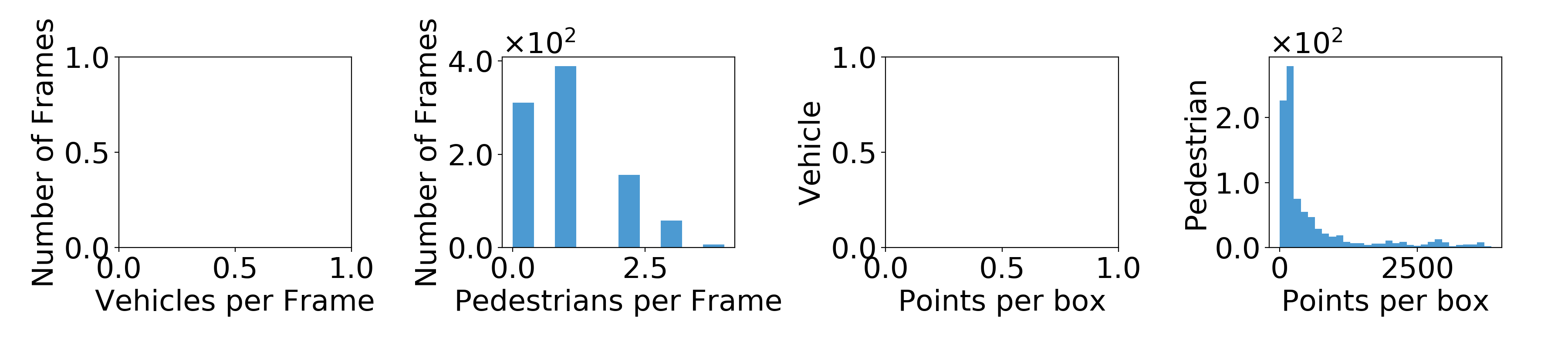}}\end{minipage}
    \\\cmidrule{3-4}
    & & skatepark\_2 & \begin{minipage}[b]{1.05\columnwidth}\centering\raisebox{-.47\height}{\includegraphics[width=\linewidth]{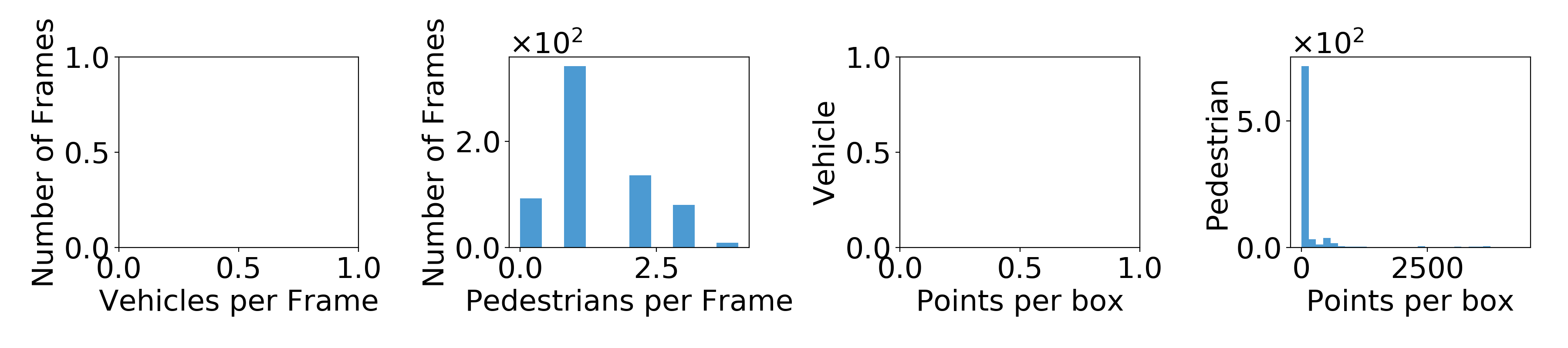}}\end{minipage}
    \\\cmidrule{3-4}
    & & srt\_green\_loop & \begin{minipage}[b]{1.05\columnwidth}\centering\raisebox{-.47\height}{\includegraphics[width=\linewidth]{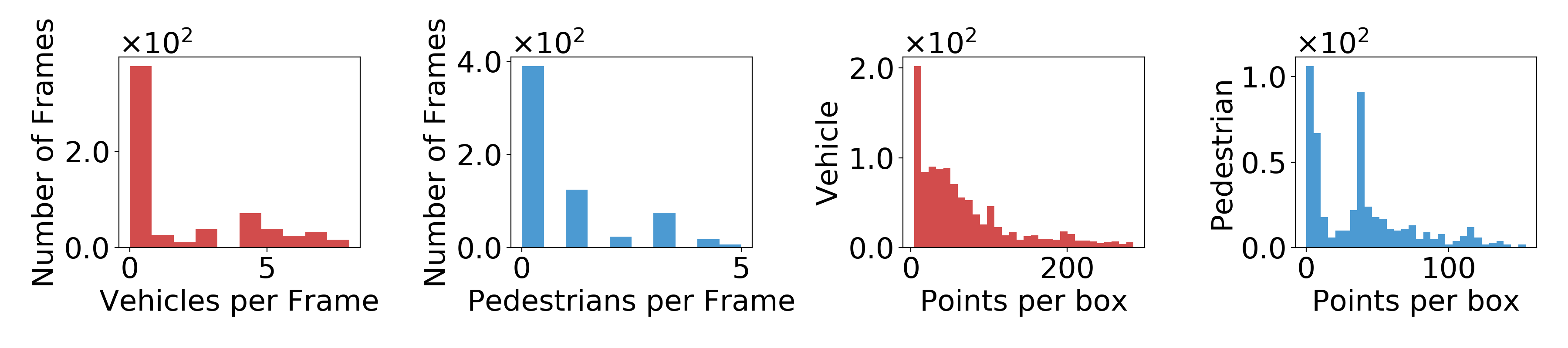}}\end{minipage}
    \\\cmidrule{3-4}
    & & srt\_under\_bridge\_1 & \begin{minipage}[b]{1.05\columnwidth}\centering\raisebox{-.47\height}{\includegraphics[width=\linewidth]{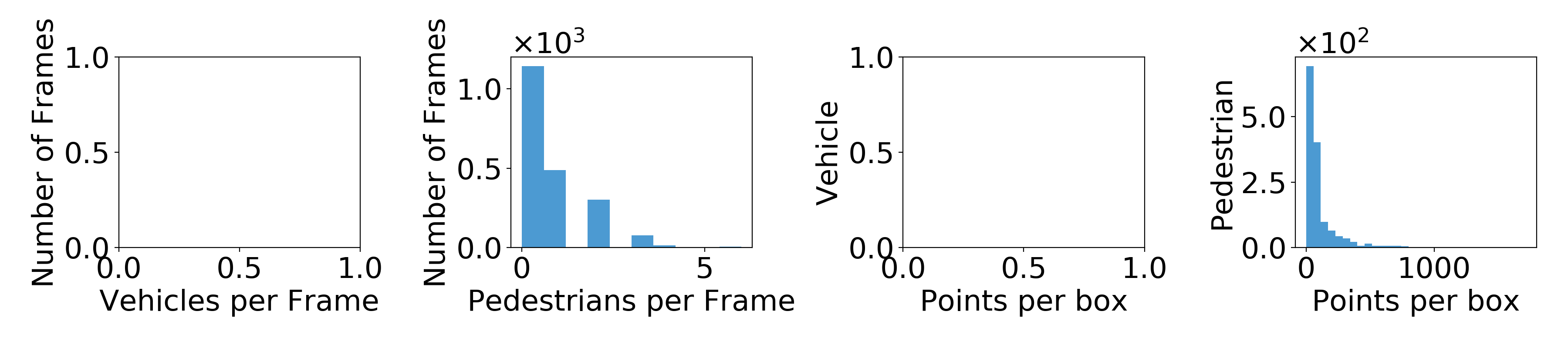}}\end{minipage}
    \\\cmidrule{3-4}
    & & srt\_under\_bridge\_2~~~~~ & \begin{minipage}[b]{1.05\columnwidth}\centering\raisebox{-.47\height}{\includegraphics[width=\linewidth]{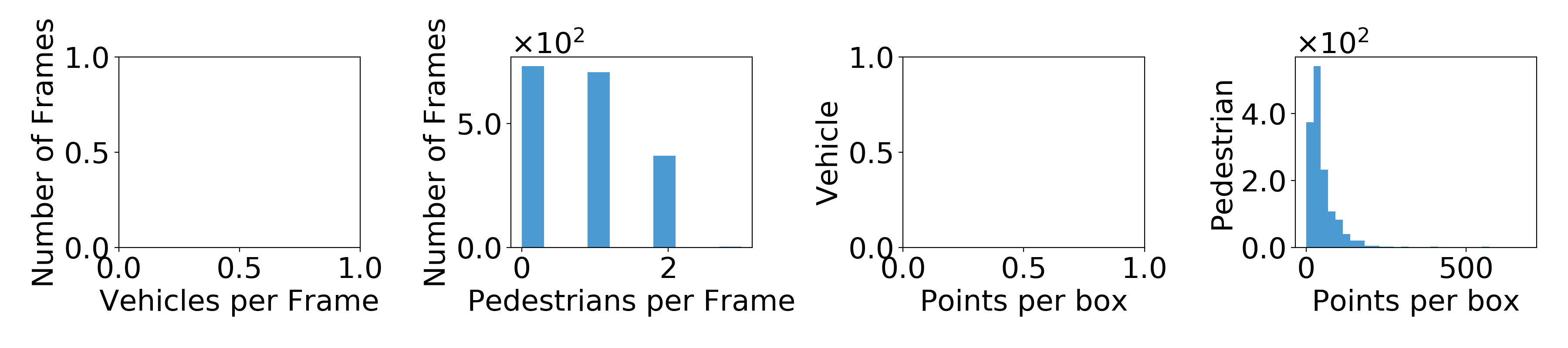}}\end{minipage}
    \\\cmidrule{2-4}
    & \multirow{6}{*}{\makecell{~~Nighttime~~\\$(2)$}} & penno\_plaza\_lights & \begin{minipage}[b]{1.05\columnwidth}\centering\raisebox{-.47\height}{\includegraphics[width=\linewidth]{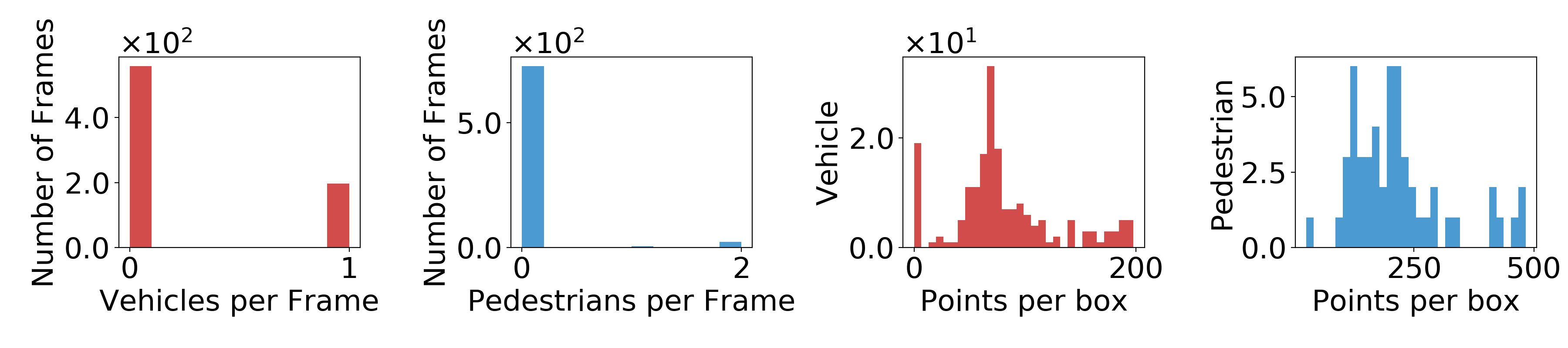}}\end{minipage}
    \\\cmidrule{3-4}
    & & penno\_short\_loop & \begin{minipage}[b]{1.05\columnwidth}\centering\raisebox{-.47\height}{\includegraphics[width=\linewidth]{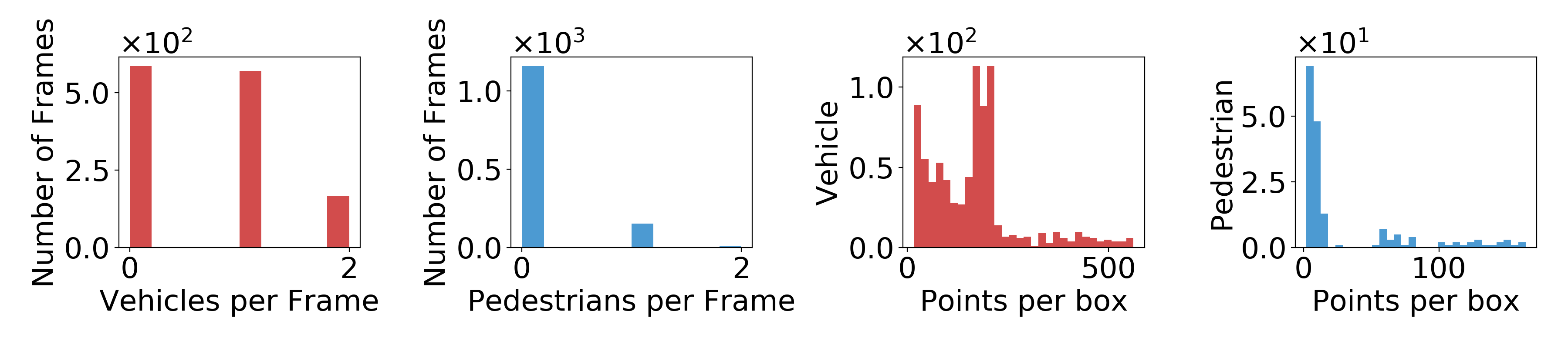}}\end{minipage}
    \\
    \bottomrule
\end{tabular}}
\label{tab:dataset_bbox_quadruped}
\end{table*}

\clearpage

\clearpage

\clearpage

\clearpage

\clearpage

\clearpage

\clearpage\clearpage
\bibliographystyle{plainnat}
\bibliography{main}

\end{document}